\documentclass[runningheads]{llncs}

\usepackage{taseabbrv}
\usepackage{tase}

\usepackage{graphicx}
\usepackage{booktabs}

\usepackage[accsupp]{axessibility}  %
\usepackage[pagebackref,breaklinks,colorlinks,citecolor=taseblue]{hyperref}
\usepackage{orcidlink}

\usepackage{multirow}
\usepackage{array}
\usepackage{subcaption}
\usepackage{rotating}
\usepackage{tikz}
\usepackage{cuted}

\usepackage{pifont}
\newcommand{\cmark}{\ding{51}}
\newcommand{\xmark}{\ding{55}}

\makeatletter
\let\titleold\title
\renewcommand{\title}[1]{\titleold{#1}\gdef\thetitle{#1}}
\makeatother

\newcommand{\maketitlesupplementary}{%
  \begingroup
    \centering
    {\Large\bfseries \thetitle\par}%
    \vspace{0.5em}%
    Supplementary Material\par
    \vspace{1.0em}%
  \endgroup
}

\usetikzlibrary{positioning}
\usetikzlibrary{arrows.meta}
\usetikzlibrary{shapes.multipart}
\usetikzlibrary{shapes.geometric}
\usetikzlibrary{calc}
\usetikzlibrary{fit}

\tikzstyle{block}=[rectangle, draw, rounded corners, minimum width=1cm, minimum height=1cm, align=center, fill=gray!20]
\tikzstyle{cblock}=[rectangle, draw, rounded corners, minimum width=1cm, minimum height=1cm, align=center, fill=orange!20]
\tikzstyle{add}=[circle, draw, inner sep=2pt, minimum size=4mm]
\tikzstyle{lbl}=[align=center]
\tikzset{
    flow/.style={->, thick, shorten >=2pt, shorten <=2pt},
    flowU/.style={->, ultra thick, shorten >=2pt, shorten <=2pt},
    flowNoTip/.style={-, thick, shorten >=2pt, shorten <=2pt},        %
    flowUNoTip/.style={-, ultra thick, shorten >=2pt, shorten <=2pt}  %

}
\tikzset{
    pics/imagestack/.style args={#1}{
            code={
                    \def\stackwidth{1.5cm}
                    \def\stackdx{2mm}
                    \def\stackdy{2mm}
                    \def\xshift{-2mm}
                    \def\yshift{0mm}
                    \def\stackrotate{0}
                    \def\stackfiles{}
                    \def\stackframe{black!20}
                    \def\stackradius{2pt}
                    \def\stacklabel{test}
                    \def\stacklabelsep{0pt}
                    #1

                    \def\last{0}

                    \begin{scope}[shift={(\xshift,\yshift)}]

                        \foreach[count=\idx] \f in \stackfiles {
                            \pgfmathsetlengthmacro{\sx}{(\idx-1)*\stackdx}
                            \pgfmathsetlengthmacro{\sy}{(\idx-1)*\stackdy}
                            \begin{scope}[shift={(\sx,\sy)}, rotate=\stackrotate]
                                \node[inner sep=0pt] (simg\idx) {\includegraphics[width=\stackwidth]{\f}};
                                \draw[\stackframe, rounded corners=\stackradius, line width=0.4pt]
                                ([xshift=-1pt,yshift=-1pt]simg\idx.south west) rectangle
                                ([xshift= 1pt,yshift=  1pt]simg\idx.north east);
                            \end{scope}
                            \xdef\last{\idx} %
                        }

                        \ifnum\last>0
                            \node[fit=(simg1)(simg\last), inner sep=1pt] (stackBBox) {};
                            \node[font=\scriptsize, below=\stacklabelsep of stackBBox.south] (stackLabelNode) {\stacklabel};
                            \node[fit=(stackBBox)(stackLabelNode), inner sep=0pt] (stackAll) {};
                        \else
                            \node[font=\scriptsize, inner sep=2pt] (stackLabelNode) {\stacklabel};
                            \node[fit=(stackLabelNode), inner sep=0pt] (stackAll) {};
                        \fi

                    \end{scope}

                }
        }
}

\newcommand{\imgwithdot}[1]{%
\begin{tikzpicture}
  \node[inner sep=0] (img) {\includegraphics[width=\linewidth]{#1}};
  \begin{scope}[shift={(img.south west)},
                x={(img.south east)}, y={(img.north west)}]
    \fill[red] (\dotx,\doty) circle (\dotsize);
  \end{scope}
\end{tikzpicture}%
}

\usepackage[font=small]{caption}

\def\secref#1{Sec.~\ref{#1}}
\def\figref#1{Fig.~\ref{#1}}
\def\tabref#1{Tab.~\ref{#1}}
\def\eqref#1{Eq.~(\ref{#1})}

\begin{document}

\title{TASE: Truncation-Aware Semantic Embeddings for 3D Scene Understanding and Editing}

\author{Tim-Felix Faasch\inst{1}\orcidlink{0009-0004-7410-5232} \and
Jochen Kall\inst{1} \and
Lucas Nunes\inst{2}\orcidlink{0000-0002-1752-2740} \and 
Jens Behley\inst{3}\orcidlink{0000-0001-6483-0319} \and 
Cyrill Stachniss\inst{3}\orcidlink{0000-0003-1173-6972} \
}

\authorrunning{Faasch, Kall, Nunes, Behley, Stachniss}
\titlerunning{Truncation-Aware Semantic Embeddings for 3D Scene Editing}

\institute{
Bosch Research, Hildesheim, Germany\\
\email{\{tim-felix.faasch, jochen.kall\}@de.bosch.com}
\and
Rheinisch-Westfälische Technische Hochschule Aachen, Aachen, Germany\\
\email{nunes@vision.rwth-aachen.de}
\and
University of Bonn, Bonn, Germany\\
\email{\{jens.behley, cyrill.stachniss\}@uni-bonn.de}\\
}

\maketitle

\begin{figure}
    \vspace{-0.3cm}
    \centering
    \begin{tikzpicture}
        \newlength\ImgH         \setlength{\ImgH}{4.65cm}          %
        \newlength\HSpaceLen    \setlength{\HSpaceLen}{0.1cm}   %
        \newlength\VSpaceLen    \setlength{\VSpaceLen}{0.05cm}   %
        \newlength\TriLen       \setlength{\TriLen}{0.3cm}      %
        \newlength\SceneWidth   \setlength{\SceneWidth}{0.4\linewidth}
        \newlength\PromptWidth  \setlength{\PromptWidth}{0.2\linewidth}

        \newlength\EditWidth
        \setlength{\EditWidth}{%
            \dimexpr(\linewidth - \SceneWidth - \PromptWidth - \TriLen - 3\HSpaceLen)/2\relax}

        \newlength\EditTileH
        \setlength{\EditTileH}{\dimexpr(\ImgH - 2\VSpaceLen)/3\relax}

        \newdimen\ArrowHead   \ArrowHead=0.25cm
        \newdimen\ArrowHalfH  \ArrowHalfH=0.5\EditTileH
        \newdimen\RectW \newdimen\Lx \newdimen\Rx \newdimen\Ly \newdimen\Ry

        \newcommand{\GapArrow}[3]{%
            \path (#1); \pgfgetlastxy{\Lx}{\Ly}%
            \path (#2); \pgfgetlastxy{\Rx}{\Ry}%
            \RectW=\Rx \advance\RectW by -\Lx \advance\RectW by -\ArrowHead
            \coordinate (RectR) at ($(#2)+(-\ArrowHead,0)$);
            \coordinate (RectL) at (#1|-#2);
            \filldraw[fill=blue!30, draw=none]
            ($(RectL)+(0,-\ArrowHalfH)$) -- ($(RectR)+(0,-\ArrowHalfH)$) -- (#2)
            -- ($(RectR)+(0,\ArrowHalfH)$) -- ($(RectL)+(0,\ArrowHalfH)$) -- cycle;
            \node[align=center, font=\scriptsize\itshape, inner sep=2pt,
                text width=\the\RectW, anchor=center]
            at ($(RectL)!0.5!(RectR)$) {#3};
        }
        \begin{scope}%
            \node[inner sep=0] (Base) {%
                \includegraphics[width=\SceneWidth, height=\ImgH, keepaspectratio=false, trim=35 0 185 0, clip]{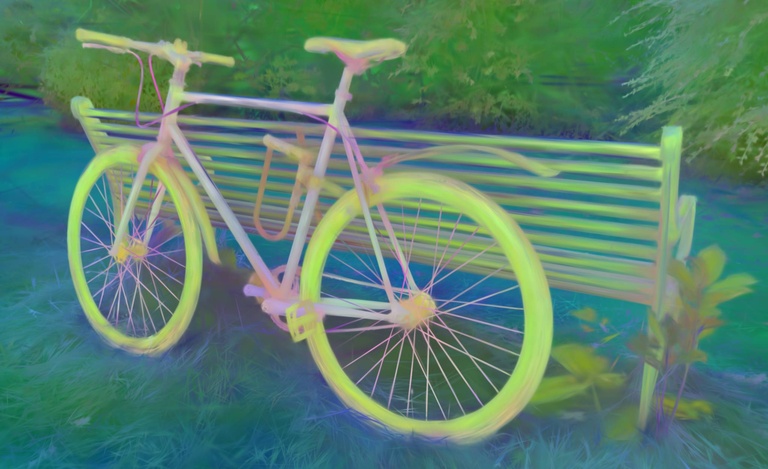}};
            \node[inner sep=0] (cut_ul) at ($(Base.north west)!0.33!(Base.north east)$) {};
            \node[inner sep=0] (cut_ur) at ($(Base.north west)!0.66!(Base.north east)$) {};
            \node[inner sep=0] (cut_ll) at ($(Base.south west)!0.33!(Base.south east)$) {};
            \node[inner sep=0] (cut_lr) at ($(Base.south west)!0.66!(Base.south east)$) {};
            \begin{scope}
                \clip ($(Base.north west)!0.5!(Base.north east)$) -- (Base.north east) -- (Base.south east)
                -- ($(Base.south west)!0.5!(Base.south east)$) -- cycle;
                \node[inner sep=0] at (Base.center) {\includegraphics[width=\SceneWidth, height=\ImgH, keepaspectratio=false, trim=35 0 185 0, clip]{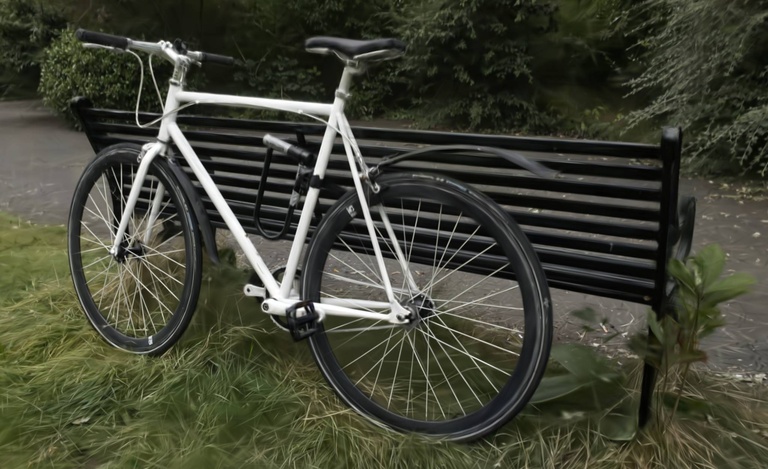}};
            \end{scope}
            \begin{scope}
                \clip (cut_ul) -- ($(Base.north west)!0.66!(Base.north east)$) -- ($(Base.south west)!0.66!(Base.south east)$)
                -- (cut_ll) -- cycle;
                \node[inner sep=0] at (Base.center) {\includegraphics[width=\SceneWidth, height=\ImgH, keepaspectratio=false, trim=35 0 185 0, clip]{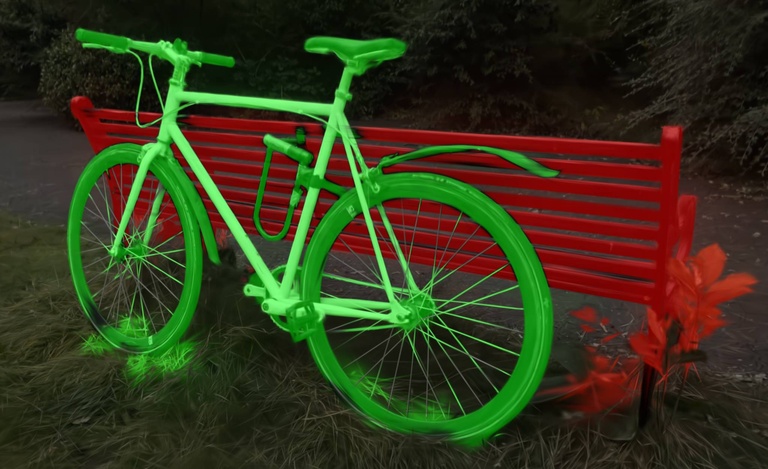}};
            \end{scope}
            \node[inner sep=0, align=center] (lbl_1) at ($(Base.south west)!0.166!(Base.south east) + (0, -0.4)$) {\scriptsize Semantic \\[-2pt] \scriptsize Features};
            \node[inner sep=0, align = center] (lbl_2) at ($(Base.south west)!0.5!(Base.south east) +(0, -0.4)$) {\scriptsize Segmentation \\[-2pt] \scriptsize Mask};
            \node[inner sep=0, align=center] (lbl_3) at ($(Base.south west)!0.83!(Base.south east) + (0, -0.4)$) {\scriptsize RGB \\[-2pt] \scriptsize Appearance};
        \end{scope}

        \coordinate (RightStart) at ($(Base.east)+(\TriLen+\PromptWidth+2*\HSpaceLen,0)$);

        \node[inner sep=0] (edit_11) at ($(RightStart)+(0.5*\EditWidth,0)$)
        {\includegraphics[height=\EditTileH, keepaspectratio, trim=125 0 50 0, clip]{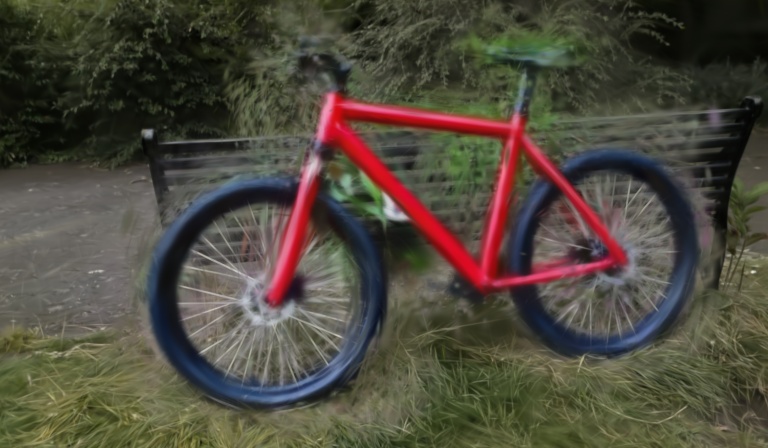}};
        \node[inner sep=0, right=\HSpaceLen of edit_11]
        (edit_12) {\includegraphics[height=\EditTileH, keepaspectratio, trim=175 0 0 0, clip]{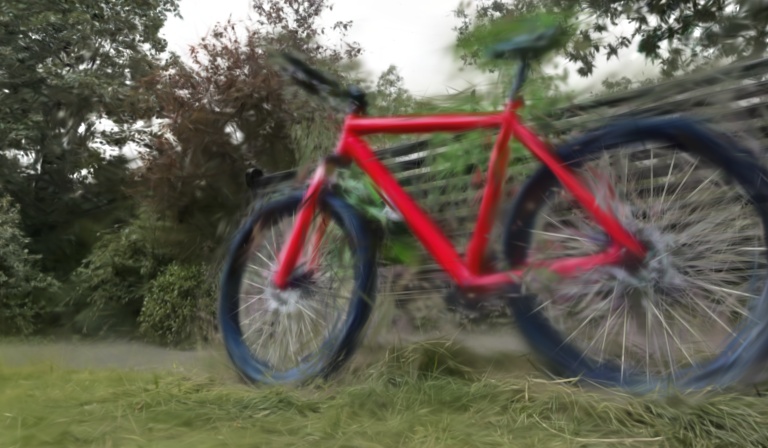}};

        \node[inner sep=0, above=\VSpaceLen of edit_11]
        (edit_21) {\includegraphics[height=\EditTileH, keepaspectratio, trim=125 0 50 0, clip]{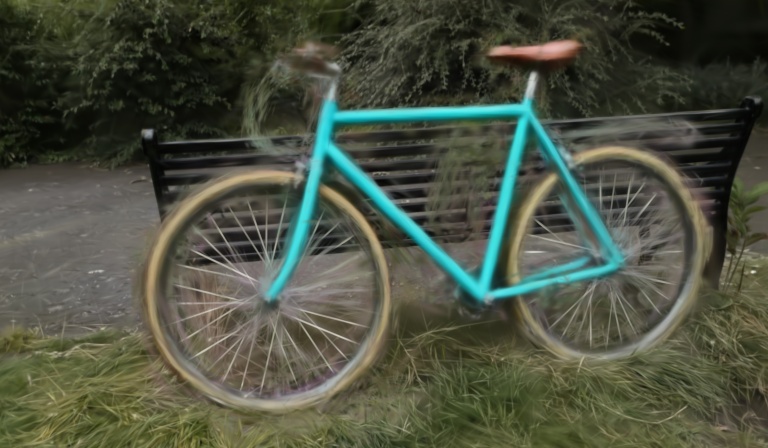}};
        \node[inner sep=0, right=\HSpaceLen of edit_21]
        (edit_22) {\includegraphics[height=\EditTileH, keepaspectratio, trim=175 0 0 0, clip]{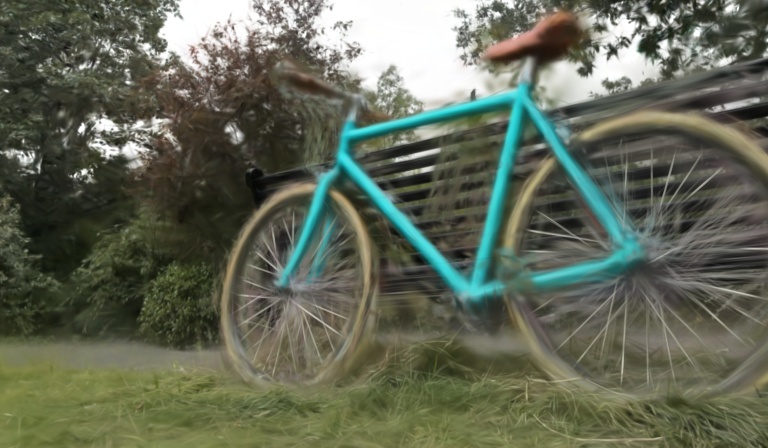}};

        \node[inner sep=0, below=\VSpaceLen of edit_11]
        (edit_31) {\includegraphics[height=\EditTileH, keepaspectratio, trim=125 0 50 0, clip]{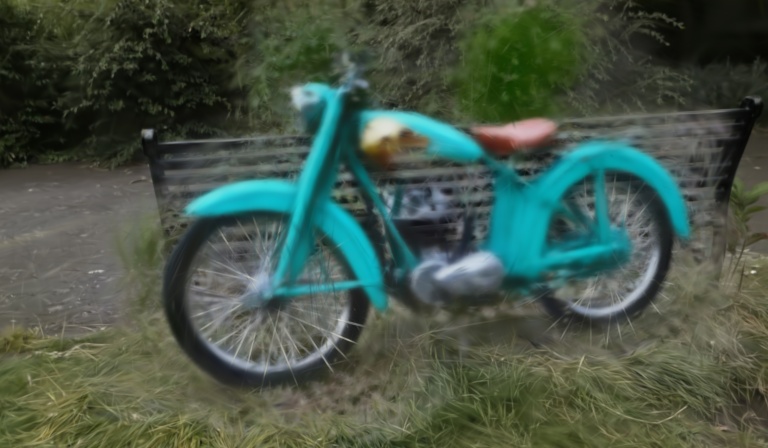}};
        \node[inner sep=0, right=\HSpaceLen of edit_31]
        (edit_32) {\includegraphics[height=\EditTileH, keepaspectratio, trim=175 0 0 0, clip]{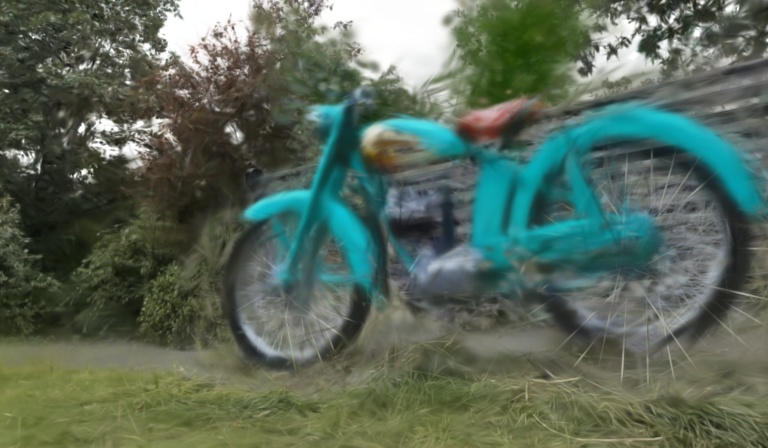}};

        \node (pos_edits) at ($(edit_11)!0.5!(edit_12)$) {};
        \node[inner sep=0] (lbl_edits) at (pos_edits |- lbl_1) {\scriptsize Edited Scenes};

        \def\TriHalf{\EditTileH/2}

        \coordinate (Lgap1) at ($(Base.east)+(\HSpaceLen,0)$) |- (edit_11.center);
        \coordinate (Rtip1) at ($(edit_11.west)+(-\HSpaceLen,0)$);
        \GapArrow{Lgap1}{Rtip1}{\tiny{a light red mountain bike MTB with a chrome suspension fork, disc brakes and fat tires}}

        \coordinate (Lgap2) at ($(Base.east)+(\HSpaceLen,0)$) |- (edit_21.center);
        \coordinate (Rtip2) at ($(edit_21.west)+(-\HSpaceLen,0)$);
        \GapArrow{Lgap2}{Rtip2}{\tiny{a light blue classic road bike with a brown and chrome handlebar, a brown saddle and beige tires}}

        \coordinate (Lgap3) at ($(Base.east)+(\HSpaceLen,0)$) |- (edit_31.center);
        \coordinate (Rtip3) at ($(edit_31.west)+(-\HSpaceLen,0)$);
        \GapArrow{Lgap3}{Rtip3}{\tiny{a classic blue motorcycle with a low brown saddle, chrome details and a round headlight}}

        \node (pos_prompt) at ($(Lgap1)!0.5!(Rtip1)$) {};
        \node[inner sep=0] (lbl_prompt) at (pos_prompt |- lbl_1) {\scriptsize Text Prompt};

    \end{tikzpicture}
    \captionof{figure}{We integrate truncation-aware semantic embeddings (TASE) to 3DGS, enabling semantic segmentation and text-guided edits. The figure shows the original scene, segmentation, and features (left), and the resulting localized edits (right), where only the specified object (bicycle) is modified.}
    \label{fig:motivation}
    \vspace{-1.0cm}
\end{figure}
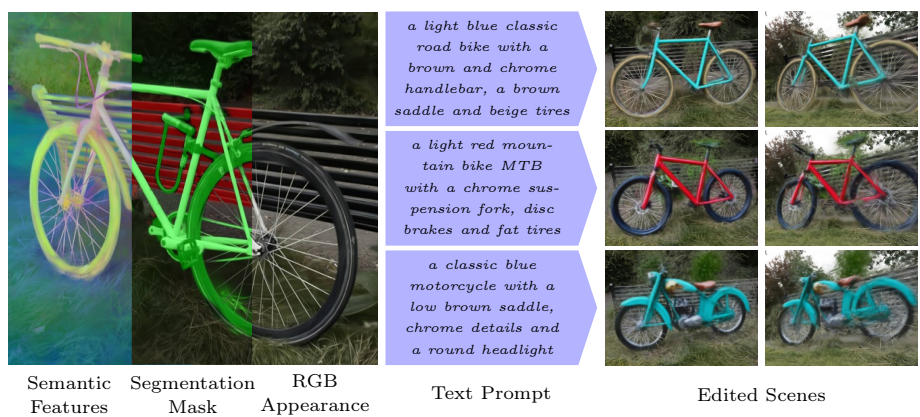

\begin{abstract}
High-fidelity semantic 3D scene representations are crucial for numerous applications, including robotics, autonomous driving, and simulation. Beyond this, the ability to edit such representations enables developers to adapt these applications more easily to specific target scenarios. Current approaches provide limited support for controllable editing. We introduce TASE, a method that projects pretrained 2D semantic features into a truncation-aware embedding space to enable flexible 3D scene editing. Our method explicitly optimizes a feature space in which progressively reducing feature channels yields increasingly abstract semantic representations, while retaining more channels preserves fine-grained detail. Additionally, we improve multi-view consistency of the features using a scale- and translation-equivariance loss. The resulting truncation-aware embedding space enables text-driven edits to 3D scenes, providing explicit control over how strongly edits adhere to the original scene content and allowing more substantial modifications than prior methods. Moreover, we propose a finetuning stage for the editing diffusion model to mitigate artifacts caused by geometric changes. Experimental results demonstrate competitive performance in 3D scene editing, substantially outperforming prior methods on edits involving large geometric modifications.
\end{abstract}

\section{Introduction}
\label{sec:intro}

Accurate and efficient 3D scene representations, like 3D Gaussian Splatting (3DGS), are increasingly important for robotics, autonomous driving, and simulation~\cite{Yan2024, Zheng2024, Zhou2024a, Tonderski2024, Jiang2024, Xie2024c}. Many of these applications require the ability to efficiently modify scenes, e.g., to alter objects or to simulate weather and lighting changes~\cite{Ye2024, NVIDIA2025}. However, current 3D representations provide limited support for structured, controllable editing. We argue that a semantic scene representation is key to efficiently enabling such edits. Recent advances in large-scale pretrained 2D feature extractors~\cite{Oquab2023, Simeoni2025} and diffusion-based image generation models~\cite{Labs2024, Podell2023} suggest that semantically meaningful embeddings can serve as a powerful interface for controllable scene manipulation.

Incorporating semantic information into 3DGS~\cite{Kerbl2023, Wu2024a} has emerged as an active area of research. The integration of semantics facilitates downstream tasks such as segmentation of elements in the scene~\cite{Zhi2021, Kundu2022} and editing of the scene appearance and geometry~\cite{Qiu2024, Ye2024}. Some existing works rely on discrete class or instance logits, which are inherently limited by their coarse, fixed vocabulary \cite{Ye2024, Zhou2024d,Zhou2024}. Open-vocabulary vision encoders such as CLIP~\cite{Radford2021} and DINO~\cite{Oquab2023, Simeoni2025} provide rich semantic embeddings beyond discrete class logits. Recent works integrate such features into neural scene representations and 3DGS~\cite{Guo2024a, Wu2024c, Shi2024, Qin2024, Kim2025, Qiu2024, JunSeong2025}. However, directly lifting 2D embeddings to 3D remains challenging: high feature dimensionality increases memory and compute costs \cite{Qin2024, JunSeong2025, Wu2024c}, absolute positional biases hinder cross-view consistency \cite{Yang2024, Yang2024b, Yue2024}, and existing approaches lack a principled mechanism to control semantic abstraction, i.e., the level of semantic specificity ranging from coarse class-level structure to fine-grained instance and texture detail \cite{Wang2024d}.

Controllable semantic abstraction is beneficial for scene editing since it allows to define, how closely changes should be tied to the original scene content. Current 3D editing approaches still struggle with edits requiring substantial geometric change, because their edits remain too closely tied to the original scene \cite{Chen2025d,YCZCCZFWXYYWZCLYHLGL2023, Wu2024b}. Compression based approaches such as quantization \cite{Shi2024, Guo2024a, JunSeong2025,Wu2024c} or autoencoding \cite{Qin2024} address memory, but do not directly allow for controllable abstraction. By contrast, channel-ordered representations, such as principle component analysis~(PCA), appear to be effective for controlling abstraction in 2D image generation \cite{Wang2024d}, yet remain under-explored for 3D scene representations.

The main contribution of this paper is a novel method that projects pretrained 2D semantic features to truncation-aware semantic embeddings (TASE) that are free from 2D positional bias. By imposing a channel-ordered structure, TASE enable controllable semantic abstraction: retaining more channels preserves fine-grained detail, while truncation of channels yields increasingly abstract representations. Removing 2D positional cues from the embedding improves cross-view consistency when features are fused in 3D. TASE support text-driven 3D scene editing via a ControlNet~\cite{Zhang2023}, and 3D semantic segmentation. We additionally propose using a finetuning strategy for editing diffusion models to mitigate artifacts introduced by large geometry changes in the 3D scene. Our approach requires no 3D or multi-view data during training. See Fig.~\ref{fig:motivation} for examples.

\section{Related Work}
\label{sec:related}

\textbf{3D Scene Representations:} Explicit 3D data structures that can be used in rasterization pipelines, like meshes and point clouds, have been the standard representations for a long time \cite{AkenineMoeller2018}. Neural Radiance Fields (NeRF)~\cite{Mildenhall2020} have been introduced as an implicit, learning-based representation that can be optimized to match posed multi-view images. This led to their widespread use in 3D reconstruction~\cite{Barron2022}, simulation~\cite{Tonderski2024}, generation~\cite{Sargent2023, Qian2023} and editing~\cite{Haque2023}. 3D Gaussian Splatting (3DGS)~\cite{Kerbl2023} has emerged as a faster alternative to NeRFs. It models the radiance field using 3D multivariate Gaussians and renders images via splatting, enabling significantly faster optimization and interactive rendering. Its explicit representation also facilitates editing operations such as composition and geometric transformations (e.g., scaling, translation, rotation). Consequently, 3DGS has been widely adopted for scene generation~\cite{Tang2023, Hao2025, Chen2024b} and editing~\cite{YCZCCZFWXYYWZCLYHLGL2023, Wang2024e, Chen2025d, Wu2024b, Mei2024a}. Similarly, we employ 3DGS as our 3D scene representation, extending it to incorporate semantic information.

\textbf{3D Scene Editing:} 3D scene editing is the task of changing the appearance and the geometry of an existing 3D scene to reflect a given control signal like a text prompt. Existing approaches usually leverage the guidance from an image generation diffusion model to change the parameters of a scene representation, either using score distillation sampling~\cite{Poole2022, Zhuang2023, Xiong2025} or pseudo views~\cite{Haque2023, YCZCCZFWXYYWZCLYHLGL2023, Wang2024e, Chen2025d, Wu2024b, Wang2024g, Wen2025a, Lee_2025_CVPR}. Most of these methods use current RGB \cite{Haque2023, YCZCCZFWXYYWZCLYHLGL2023, Wang2024e, Chen2025d, Wang2024g, Wen2025a, Lee_2025_CVPR} or depth \cite{Wu2024b} renderings as input to the diffusion model. Even though effective, this limits the amount of geometric change that can be introduced by these methods. Distinctly, our method uses semantic features to condition the editing diffusion model instead, allowing for complex edits including large changes in geometry.

\textbf{Open Vocabulary Semantic Features:} Weakly- and self-supervised learning for open vocabulary feature extraction has gained a lot of interest for downstream tasks such as classification, detection, or segmentation, especially where training data is sparse~\cite{Jiang2023}. CLIP~\cite{Radford2021} popularized the use of a contrastive loss to create a joint embedding space for text and images. The approach has been extended to dense image features by using masked self distillation~\cite{Dong2023}. Another line of work, including DINO, uses pretraining on visual data only to produce both dense and global deep features~\cite{Zhou2021, Oquab2023, Simeoni2025}. 

A number of recent works have proposed to include pretrained image features into scene reconstruction to aid with scene understanding, allowing to segment and specifically target objects within the scene~\cite{Yang2023,JunSeong2025,Qin2024,Wu2024c,Guo2024a}. Due to the high dimensionality of the semantic features, most of these methods use compression based approaches such as quantization \cite{Shi2024,Guo2024a,JunSeong2025,Wu2024c} or a per-scene autoencoder (AE) \cite{Qin2024}. This reduces the required memory, but does not allow for semantic abstraction. Some works investigate channel-ordered latent representations, in which earlier channels capture more important information than later ones \cite{Rippel2014, Kusupati2022}. Nested dropout~\cite{Rippel2014} encourages such structure by stochastically truncating later channels of the latent vector during training. Matryoshka representation learning~\cite{Kusupati2022} extends this idea by jointly optimizing for a fixed set of truncation levels, improving stability and robustness across different embedding sizes. 

Despite their effectiveness, 2D positional cues produced by pretrained 2D feature extractors hinder their direct applicability in 3D, especially in a per-scene optimization setting \cite{Yang2023, Yang2024b, Yue2024}. Existing solutions either suppress these cues during scene optimization~\cite{Yang2023} or apply fine-tuning of the backbone~\cite{Yang2024b, Yue2024}.

Following previous work, our method uses pretrained image features as the basis for the semantic embeddings. We use an AE trained with a Matryoshka representation learning to simultaneously reduce the memory required by the features, and to enable controllable semantic abstraction. We additionally employ a scale- and translation-equivariance loss to remove 2D positional bias in the latent space.

\textbf{Controllable Image Generation:} Latent diffusion models have recently emerged as the primary technique for high-fidelity image and video generation~\cite{SohlDickstein2015, Rombach2022, Blattmann2023, Labs2024}. They are trained to progressively remove Gaussian noise from images through a reverse diffusion process. To improve efficiency, latent diffusion models perform denoising in the latent space of a variational autoencoder (VAE). During generation, the process can be conditioned on text prompts, enabling text-to-image synthesis. Some methods additionally allow to condition the generation on spatial control modalities such as edge maps, depth maps, or segmentation maps~\cite{Mou2024, Zhang2023}. ControlNet~\cite{Zhang2023} does this by creating a copy of the diffusion model, that is trained to process the additional control modality, while the weights of the original model are kept frozen. Similarly, we use a ControlNet style network conditioned on our truncation-aware embedding space to generate edited views of the 3D Scene.

We propose leveraging positional-bias-free, truncation-aware semantic embeddings (TASE) for 3D scene editing. TASE are multi-view consistent and enable controllable levels of semantic abstraction. Our pipeline supports complex text-driven edits of 3D scenes, allowing for substantial modifications of the scene appearance and geometry.

\section{Preliminaries on 3D Gaussian Splatting}

In 3DGS~\cite{Kerbl2023}, the radiance field of a scene is represented as a set of $N$ multi-variate gaussians~$\mathcal{G} = \{\mathcal{G}_1, \dots, \mathcal{G}_N\}$. Each Gaussian~$\mathcal{G}_i$ is parametrized by a mean value~$\boldsymbol{\mu}_i\in\mathbb{R}^3$ and a covariance matrix~$\mathbf{\Sigma_i}\in\mathbb{R}^{3\times3}$. The value of the~$i^\text{th}$ Gaussian~$\mathcal{G}_i$ at the position~$\mathbf{x}\in\mathbb{R}^3$ is given as:
\begin{equation}
    \mathcal{G}_i(\mathbf{x}) = \exp\left(-\frac{1}{2}(\mathbf{x} - \boldsymbol{\mu}_i)^\top\mathbf{\Sigma}_i^{-1}(\mathbf{x}- \boldsymbol{\mu}_i)\right).
    \label{eq:gaussian}
\end{equation}

During rendering, the gaussians are projected to image space, leading to a 2-dimensional covariance~$\mathbf{\Sigma}_i'\in\mathbb{R}^{2\times2}$ given the viewing transformation matrix~$\mathbf{W}\in\mathbb{R}^{3\times3}$ and it's Jacobian~$\mathbf{J}\in\mathbb{R}^{2\times3}$~\cite{Zwicker2001}:
\begin{equation}
    \mathbf{\Sigma}_i' = \mathbf{J} \mathbf{W} \mathbf{\Sigma}_i \mathbf{W}^\top \mathbf{J}^\top.
    \label{eq:projection}
\end{equation}

The covariance matrix~$\mathbf{\Sigma}_i$, being orthogonal, can be decomposed into a rotation~$\mathbf{R}_i\in\mathbb{R}^{3\times3}$ and a scaling matrix~$\mathbf{S}_i\in\mathbb{R}^{3\times3}$ with:
\begin{equation}
    \mathbf{\Sigma}_i = \mathbf{R}_i \mathbf{S}_i \mathbf{S}_i^\top \mathbf{R}_i^\top.
    \label{eq:cov_decomp}
\end{equation}

In practice, the covariance is parametrized through a rotation quaternion~$\mathbf{q}_i\in\mathbb{R}^{4}$ and a scale vector~$\mathbf{s}_i\in\mathbb{R}^3$ with $\mathbf{s}_i = \text{diag}(\mathbf{S}_i)$ which ensures orthogonality and reduces the number of parameters. The contribution of a single Gaussian~$\mathcal{G}_i$ towards the overall radiance of the scene is defined by it's opacity~$\alpha_i$ and a set of spherical harmonic (SH) coefficients~$\mathbf{\hat{c}}_{0, i}, ..., \mathbf{\hat{c}}_{M, i}$, which are multiplied with the value of the corresponding SH basis function~$H_k(\mathbf{d})$ of the order~$k$ to arrive at a color~$\mathbf{c}_i$ depending on the view direction~$\mathbf{d}\in\mathbb{R}^3$. The final pixel color~$\mathbf{c}_{\text{pixel}}$ is composed by z-ordered alpha blending the contribution~$\mathbf{c}_i$ of all Gaussians $\mathcal{G}_i$ in $\mathcal{G}$:
\begin{equation}
    \mathbf{c}_{\text{pixel}} = \sum_{i=1}^{N} T_i \alpha_i \mathbf{c}_i,
\end{equation}
with:
\begin{equation}
    T_i = \prod_{j=1}^{i-1} (1 - \alpha_j) \text{, }
    \mathbf{c}_i = \sum_{k=0}^{M} \mathbf{\hat{c}}_{k, i}\, H_k(\mathbf{d}),
    \label{eq:alpha_blending}
\end{equation}
where~$M$ is the maximum SH order used for view-dependent color modeling. Thus each Gaussian~$\mathcal{G}_i$ is parametrized by $\Theta_i = (\boldsymbol{\mu}_i, \mathbf{q}_i, \mathbf{s}_i, \alpha_i, \mathbf{\hat{c}}_{0...M, i})$.

\section{Our Approach to 3D Scene Editing}
\label{sec:method}

\begin{figure}[t]
    \centering
    \resizebox{\linewidth}{!}{
        \begin{tikzpicture}[
                font=\footnotesize,>=stealth,rounded corners,
                every label/.style={font=\scriptsize, inner sep=1pt},
                node distance=1.6cm and 10mm
            ]

            \def\stackx{1cm}
            \def\stacky{0.85cm}

            \node[ellipse, minimum width=4.75cm, minimum height=3.75cm, inner sep = 0pt] (scene) {};
            \node[inner sep=0pt, align=center]  at ($(scene) + (0, 0.3)$) {\includegraphics[height=4cm]{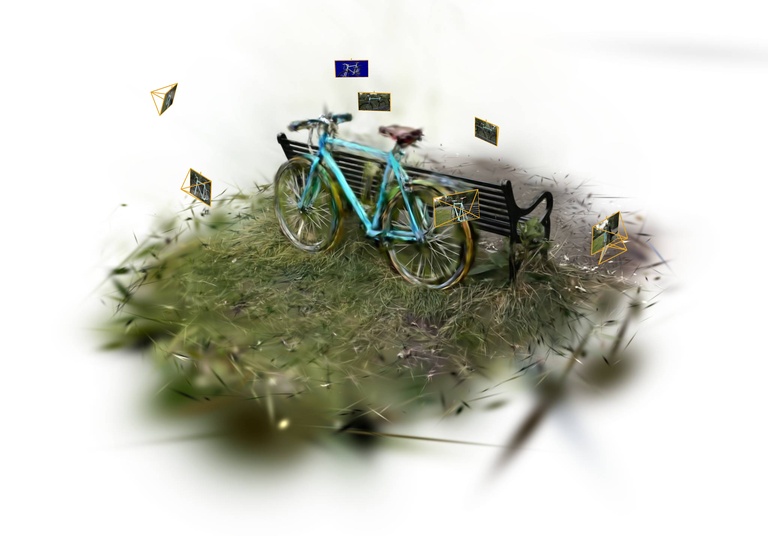}};
            \node[above=0.1cm of scene.south] {$\mathcal{G}$};

            \begin{scope}[shift={(7.5,3.3)}]
                \def\blockX{2.2}
                \def\blockY{0.9}

                \node[cblock] (b1) at (0,0) {$\mathcal{M}^\text{DiT}_1$};
                \node[] (bx) at (0.9*\blockX,0) {\ldots};
                \node[cblock] (bn) at (1.5*\blockX,0) {$\mathcal{M}^\text{DiT}_n$};

                \node[cblock] (c1) at (0,-2*\blockY) {$\mathcal{M}^\text{Ctrl}_1$};
                \node[] (cx) at ({$(c1.north east)!0.6!(c1.south east)$} -| bx) {\ldots};
                \node[cblock] (cn) at (1.5*\blockX,-2*\blockY) {$\mathcal{M}^\text{Ctrl}_n$};

                \node[add] (a0) at (-1.25*\blockX,0) {\tiny$+$};
                \node[add, minimum size=0.6cm] (blend) at (-0.6*\blockX,0) {\small$\mathbf{m}$};
                \node[add] (a1) at (0.5*\blockX,0) {\tiny$+$};
                \node[add] (a3) at (2*\blockX,0) {\tiny$+$};

                \node[block, left=0.5cm of a0] (vae_e) {$\mathcal{M}^\text{VAE}_\text{enc}$};
                \node[block, right=0.7cm of a3] (vae_d) {$\mathcal{M}^\text{VAE}_\text{dec}$};

                \draw[flow] (vae_e) -- (a0);
                \draw[flow] (a0) -- (blend);
                \draw[flow] (a3) -- (vae_d.west);
                \draw[flow] (blend) -- (b1.west);
                \draw[flow] (blend.south) |- ($(c1.north west)!0.3!(c1.south west)$);

                \draw[flow] (b1.east) -- (a1);
                \draw[flow] (a1) -- (bx) -- (bn.west);
                \draw[flow] (bn.east) -- (a3);

                \draw[flow] ($(c1.north east)!0.6!(c1.south east)$) -- (cx)-- ($(cn.north west)!0.6!(cn.south west)$);
                \draw[flow] ($(c1.north east)!0.3!(c1.south east)$) -| (a1);
                \draw[flow] (cn) -| (a3);
            \end{scope}

            \node[above=0.4cm of a0, inner sep= 0pt] (noise) {$\mathcal{N}(\mathbf{0}, \mathbf{I})$};
            \node[inner xsep = \stackx, inner ysep= \stacky, left=1cm of vae_e] (rgbA) {};
            \node[inner xsep = \stackx, inner ysep= \stacky] (ctrlA) at ($(c1) + (-3.5, -0.3)$) {};
            \node[inner xsep = \stackx, inner ysep= \stacky] (segA) at ($(rgbA) + (-2.5, 0.3)$) {};
            \node[block, minimum height = 0.5cm, fill = blue!30] (prompt) at (vae_e |- {$(b1)!0.5!(c1)$}) {\scriptsize Prompt};
            \node [inner sep=0pt] (pos3) at ($(segA.north east)!0.07!(segA.south east)$){};

            \path (rgbA) pic {imagestack={
                            \def\stackfiles{{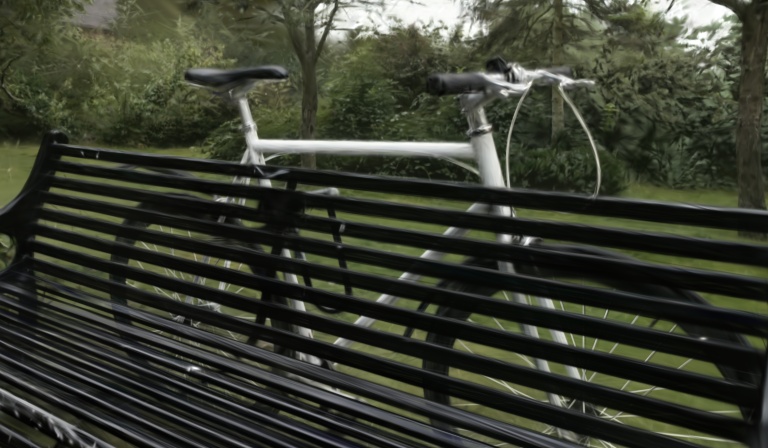},{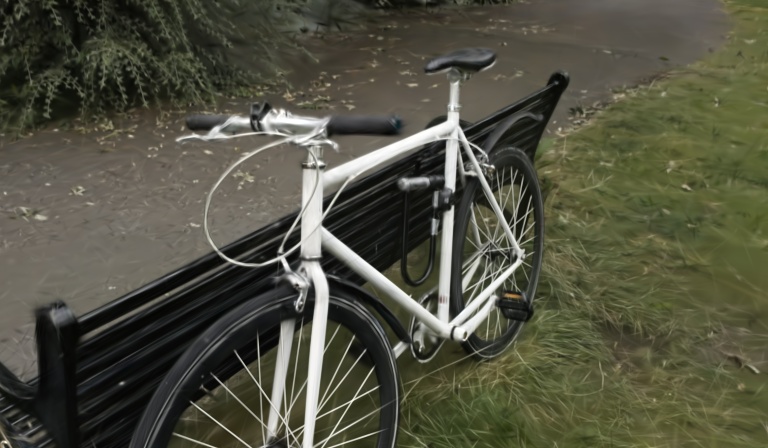},{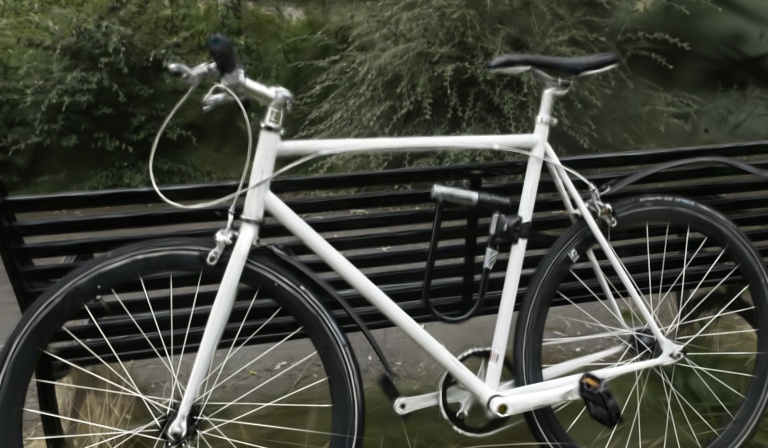}};
                            \def\stacklabel{$\mathbf{I}^\text{render}$}
                        }};

            \path (ctrlA) pic {imagestack={
                            \def\stackfiles{{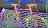},{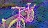},{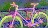}};
                            \def\stacklabel{$\mathcal{F}_h^\text{render}$}
                        }};

            \path (segA) pic {imagestack={
                            \def\stackfiles{{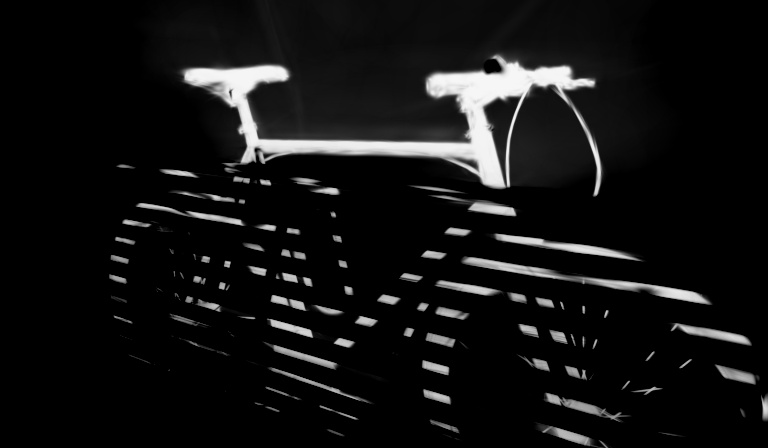},{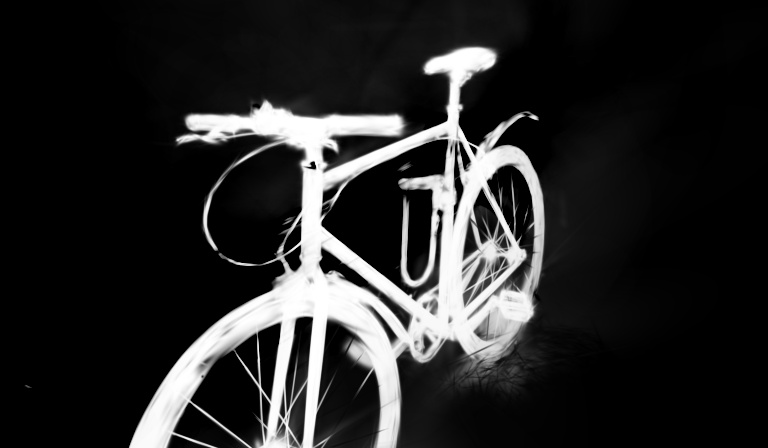},{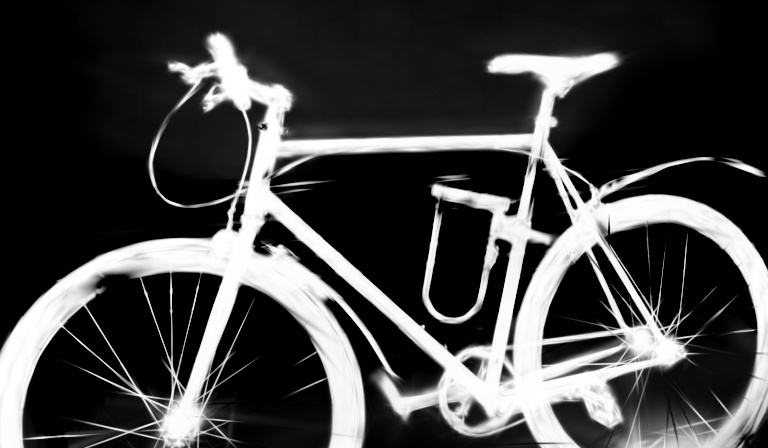}};
                            \def\stacklabel{$\mathbf{M}^\text{render}$}
                        }};

            \node[cblock, below=1.3cm of c1] (adapter) {$\mathcal{M}^\text{TASE}_\text{enc}$};
            \node[block, right=of adapter] (encoder) {$\mathcal{M}^\text{TASE}_\text{base}$};

            \node[inner xsep = 1.5*\stackx, inner ysep= 1.5*\stacky] (pvRGB) at ($(vae_d) + (0, -3.9)$) {};
            \node[inner xsep = \stackx, inner ysep= \stacky, left=1cm of adapter] (pvFeat) {};

            \path (pvRGB) pic {imagestack={
                            \def\stackfiles{{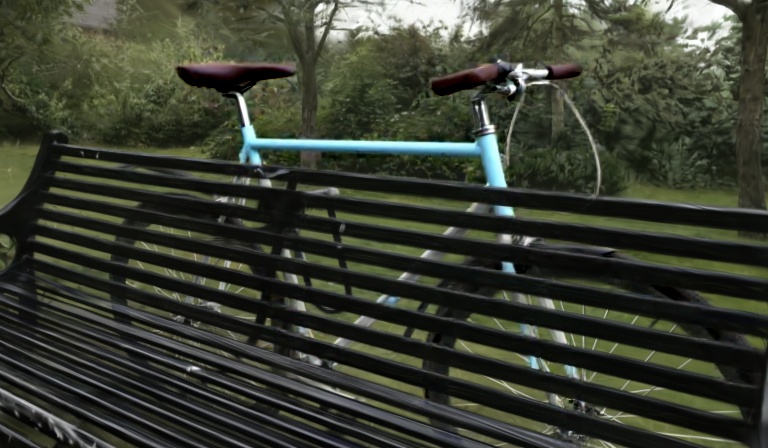},{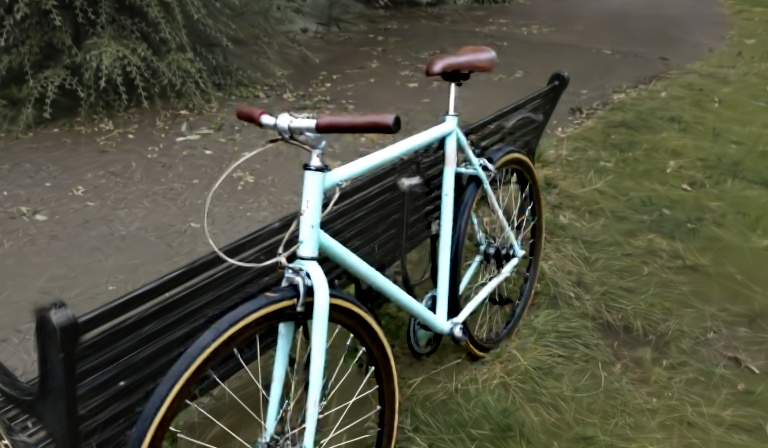},{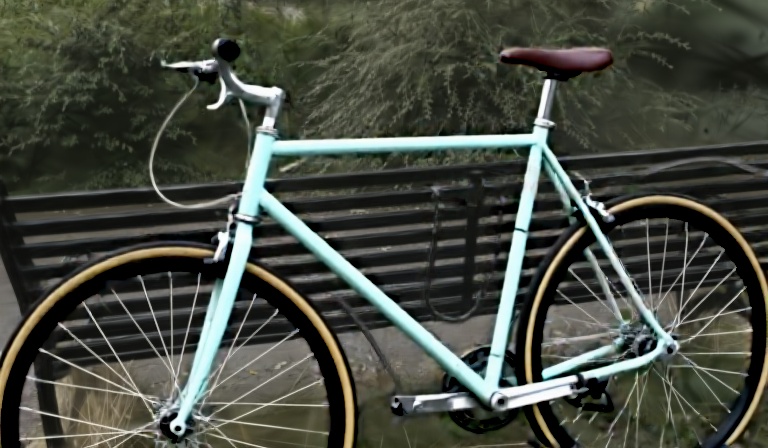}}
                            \def\stackwidth{2.5cm};
                            \def\stacklabel{$\mathbf{I}^\text{novel}$}
                        }};
            \path (pvFeat) pic {imagestack={
                            \def\stackfiles{{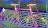},{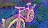},{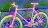}};
                            \def\stacklabel{$\mathcal{F}_h^\text{novel}$}
                        }};

            \node [inner sep=0pt] (pos) at (b1 |- pos3) {};
            \draw[flow] (a3) |- (b1 |- pos3) -| (blend.north east);
            \draw[flow] (noise) -- (a0);

            \node [] (promptx) at (prompt -| cx){\ldots};
            \draw[flow] (prompt) -| (c1);
            \draw[flow] (prompt) --(promptx) -| (cn);

            \draw[flow] (prompt) -| (b1);
            \draw[flow] (prompt) --(promptx) -| (bn);

            \draw[flowU, blue!50] (scene.north west -| segA) -- (segA);
            \draw[flowU, blue!50]  (scene.north -| rgbA) -- (rgbA);
            \draw[flowU, blue!50] (scene.north east |- ctrlA) -- (ctrlA);

            \draw[flow] (rgbA.east) -- (vae_e.west);
            \node [inner sep=0pt] (pos4) at ($(c1.north west)!0.6!(c1.south west)$){};
            \draw[flow] (ctrlA.east |- pos4) -- (pos4);
            \draw[flow] (pos3) -| (blend.north west);

            \draw[flow] (vae_d) -- (pvRGB.north);

            \draw[flow] (pvRGB.west |- encoder.east) -- (encoder.east);
            \draw[flow] (encoder.west) -- (adapter.east);
            \draw[flow] (adapter.west) -- (pvFeat.east);

            \node [inner sep=0pt] (pos2) at ($(pvRGB.north west)!0.2!(pvRGB.south west)$) {};
            \draw[flowU, color=red!50] (pos2) -- (scene.east |- pos2);
            \draw[flowU, color=red!50] (pvFeat.west) -- (scene.south east |- pvFeat.west);

            \node[draw, rounded corners, dashed, fit=(c1)(pos)(blend)(a3), inner sep=4pt, label=below:{Denoising Loop}] (loop_D) {};
            \node[draw, rounded corners, dashed, fit=(pvFeat)(pvRGB)(noise)(segA)(loop_D), inner sep=4pt, label=below:{Editing Loop}] (loop_E) {};

            \begin{scope}
                \def\ypos{0.05}
                \node[anchor=west] (leg_opt) at ($(loop_E.south west)!\ypos!(loop_E.south east) + (0, -0.55)$) {\scriptsize 3DGS Optimization};
                \draw[ultra thick, red!50] ($(leg_opt.west)+(-0.35, 0)$)--+(0.3,0);
                \node[anchor=west] (leg_rend) at ($(loop_E.south west)!\ypos+0.18!(loop_E.south east) + (0, -0.55)$) {\scriptsize 3DGS Rendering};
                \draw[ultra thick, blue!50] ($(leg_rend.west)+(-0.35, 0)$)--+(0.3,0);
                \node[anchor=west] (leg_add) at ($(loop_E.south west)!\ypos+0.34!(loop_E.south east) + (0, -0.55)$) {\scriptsize Elementwise Adding};
                \node[add, minimum size=0.35cm] (leg_add_sym) at ($(leg_add.west)+(-0.15, 0)$) {\tiny$+$};
                \node[anchor=west] (leg_blend) at ($(loop_E.south west)!\ypos+0.52!(loop_E.south east) + (0, -0.55)$) {\scriptsize Masked Blending};
                \node[add, minimum size=0.35cm] (leg_blend_sym) at ($(leg_blend.west)+(-0.15, 0)$) {\tiny$\mathbf{m}$};
                \node[anchor=west] (leg_frozen) at ($(loop_E.south west)!\ypos+0.68!(loop_E.south east) + (0, -0.55)$) {\scriptsize Frozen Model};
                \node[block, minimum size=0.35cm] (leg_frozen_sym) at ($(leg_frozen.west)+(-0.15, 0)$) {};
                \node[anchor=west] (leg_frozen) at ($(loop_E.south west)!\ypos+0.82!(loop_E.south east) + (0, -0.55)$) {\scriptsize Trained Model};
                \node[cblock, minimum size=0.35cm] (leg_frozen_sym) at ($(leg_frozen.west)+(-0.15, 0)$) {};
            \end{scope}
        \end{tikzpicture}
    }
    \caption{Overview of the proposed pipeline: Rendered semantic feature maps~$\mathcal{F}_h^\text{render}$ and a text prompt condition the diffusion model~$(\mathcal{M}^\text{DiT}, \mathcal{M}^\text{Ctrl})$ for the generation of novel views~$\mathbf{I}^\text{novel}$. We add some noise~$\mathcal{N}(\mathbf{0}, \mathbf{I})$ to the latent of the current RGB render and use it as input to~$\mathcal{M}^\text{DiT}$ and~$\mathcal{M}^\text{Ctrl}$. For local edits, the rendered segmentation masks~$\mathbf{M}^\text{render}$ are use as inpainting masks. From~$\mathbf{I}^\text{novel}$, semantic features~$\mathcal{F}_h^\text{novel}$ are extracted and optimized into the the 3D Gaussian scene~$\mathcal{G}$, jointly with~$\mathbf{I}^\text{novel}$.}
    \label{fig:main}
    \vspace{-0.4cm}
\end{figure}
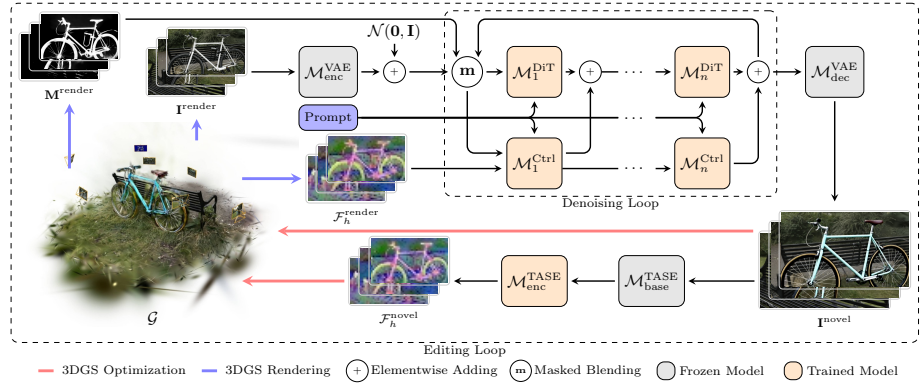

The task of 3D scene editing can be defined as follows: given a pretrained set of 3D Gaussians that represent a scene, and a control signal, such as a text prompt, the goal is to modify the parameters of the Gaussians such that the rendered images of the modified scene reflect the control signal. To this end, we leverage truncation-aware semantic embeddings (TASE) which are detailed in \secref{subsec:TASE}. We integrate TASE into a 3DGS reconstruction of the scene as described in \secref{subsec:lift_3dgs} to facilitate downstream segmentation and editing. We edit the 3D scene using a ControlNet conditioned on TASE, which is described in \secref{subsec:ctrlnet}. We optimize the parameters of the Gaussians to match novel views generated by the ControlNet as detailed in \secref{subsec:3d_editing}. For local object edits we segment the editing target in 3D space using a simple, yet effective, similarity-based algorithm, which is described in \secref{subsec:segmentation}. An overview illustration of our approach is depicted \figref{fig:main}. 

\subsection{Truncation-Aware Semantic Embeddings}
\label{subsec:TASE}

In order to condition the editing diffusion model, we need semantic features that are multi-view consistent and that enable controllable semantic abstraction. To generate our truncation-aware semantic embeddings (TASE), we start with a pretrained image feature extraction backbone~$\mathcal{M}^\text{TASE}_{\text{base}}$, which predicts a patch-wise feature map~$\mathcal{F}_o\in\mathbb{R}^{H \times W \times N_o}$ from an image~$\mathbf{I}$. We then train a symmetrical AE model with an encoder~$\mathcal{M}^\text{TASE}_{\text{enc}}$ and a decoder~$\mathcal{M}^\text{TASE}_{\text{dec}}$, which takes~$\mathcal{F}_o$ as input. The hidden representation~$\mathcal{F}_h=\mathcal{M}^\text{TASE}_{\text{enc}}(\mathcal{F}_o)$ will act as our projected TASE space, where $\mathcal{F}_h\in\mathbb{R}^{H \times W \times N_h}$. $\mathcal{F}_h$ has the same spatial resolution $H \times W$ as $\mathcal{F}_o$, but a lower number of channels~$N_h < N_o$ to create the desired compression. We jointly train $\mathcal{M}^\text{TASE}_{\text{enc}}$ and $\mathcal{M}^\text{TASE}_{\text{dec}}$ while keeping $\mathcal{M}^\text{TASE}_{\text{base}}$ frozen. The reconstruction loss~$\mathcal{L}_r$ between $\mathcal{F}_o$ and the reconstructed features~$\mathcal{F}_r=\mathcal{M}^\text{TASE}_{\text{dec}}(\mathcal{F}_h)$ is defined as:
\begin{equation}
    \mathcal{L}_r(\mathcal{F}_o, \mathcal{F}_r) = \frac{1}{HW} \sum_{x=1}^{W}\sum_{y=1}^{H} \ell_r(\mathbf{f}_{o, x, y}, \mathbf{f}_{r, x, y}),
    \label{eq:recon_loss}
\end{equation}
where the loss $\ell_r$ between individual feature vectors~$\mathbf{f}_o$ and~$\mathbf{f}_r$ at the patch location $x, y$ is defined as:
\begin{equation}
    \ell_r(\mathbf{f}_o, \mathbf{f}_r)
    = \lambda_{\text{cos}}\!\left(1-\frac{\mathbf{f}_o^\top\mathbf{f}_r}{\|\mathbf{f}_o\|\|\mathbf{f}_r\|}\right)
    + \lambda_{\text{MSE}}\|\mathbf{f}_o-\mathbf{f}_r\|_2^2,
\end{equation}
with the hyperparameters~$\lambda_{\text{cos}}\in\mathbb{R}$ and~$\lambda_{\text{MSE}}\in\mathbb{R}$.

To make~$\mathcal{F}_h$ truncation aware, we employ Matryoshka representation learning~\cite{Kusupati2022}, masking suffix channels of the embedding vectors $\mathbf{f}_h\in\mathcal{F}_h$ with zeros. We define as set of truncation levels $\mathcal{T}=\{2^n\}$ for $n=1,\dots,\log_2(N_h)$. For each training iteration we compute the mean loss across all levels $t\in\mathcal{T}$:
\begin{equation}
    \tilde{\mathcal{L}}_r = \frac{1}{\mathcal{T}} \sum_{t\epsilon\mathcal{T}}\mathcal{L}_{r}(\mathcal{F}_o, \mathcal{M}^\text{TASE}_{\text{dec}}(\texttt{mask}(\mathcal{F}_h, t))),
\end{equation}
where $\texttt{mask}(\mathcal{F}_h, t)$ sets the last~$N_h - t$ channels of each embedding vector~$\mathbf{f}_h$ in~$\mathcal{F}_h$ to zero.

Taking inspiration from DVT~\cite{Yang2024b}, we remove 2D positional cues from the embeddings by defining an additional equivariance loss~$\mathcal{L}_{\text{eqv}}$ during training:
\begin{equation}
    \mathcal{L}_\text{eqv} = \mathcal{L}_r(\texttt{crop}_\text{eqv}(\mathcal{F}_h), \mathcal{F}_{h, \text{crop}}),
\end{equation}
with
\begin{equation}
    \mathcal{F}_{h, \text{crop}} = \mathcal{M}^\text{TASE}_\text{enc}(\mathcal{M}^\text{TASE}_\text{base}(\texttt{crop}(\mathbf{I}))),
\end{equation}
where $\texttt{crop}(\mathbf{I})$ randomly crops a section of $\mathbf{I}$ and $\texttt{crop}_\text{eqv}(\mathcal{F}_h)$ crops the corresponding region from $\mathcal{F}_h$. The total loss~$\mathcal{L}_{\text{TASE}}$ for training~$\mathcal{M}^\text{TASE}_\text{enc}$ and~$\mathcal{M}^\text{TASE}_\text{dec}$ is then defined as:
\begin{equation}
    \mathcal{L}_{\text{TASE}} = \tilde{\mathcal{L}}_r + \lambda_{\text{eqv}} \mathcal{L}_{\text{eqv}},
    \label{eq:tase_loss}
\end{equation}
with the hyperparameter~$\lambda_{\text{eqv}}\in\mathbb{R}$.

\subsection{Lifting of the Embeddings into 3DGS}
\label{subsec:lift_3dgs}

To use TASE for 3D scene editing, we need to integrate the semantic features~$\mathcal{F}_h$ into a 3DGS scene. To this end, append a randomly initialized feature vector~$\mathbf{f}_i$ to the Gaussian parameters~$\Theta_i$. We pass each input image through~$\mathcal{M}^\text{TASE}_{\text{base}}$ and~$\mathcal{M}^\text{TASE}_{\text{enc}}$ to obtain~$\mathcal{F}_h^\text{image}$. We then optimize $\mathbf{f}_i$ alongside the other parameters in $\Theta_i$ to minimize the feature loss~$\mathcal{L}_f$ between the rendered feature maps~$\mathcal{F}_h^\text{rendered}$ and~$\mathcal{F}_h^\text{image}$:
\begin{equation}
    \mathcal{L}_f = \mathcal{L}_r(\mathcal{F}_h^\text{image}, \mathcal{F}_h^\text{rendered}).
    \label{eq:feature_loss}
\end{equation}

In addition, we use the photometric loss~$\mathcal{L}_\text{l1}$ and the SSIM loss $\mathcal{L}_\text{SSIM}$~\cite{Wang2004} used in standard 3DGS reconstruction~\cite{Kerbl2023}. Since~$\mathcal{F}_h$ has a lower spatial resolution than the input images, we add a local smoothness loss~$\mathcal{L}_{s}$ on the feature vectors of the k-nearest Gaussians~$\texttt{kNN}(\mathcal{G}_i, \mathcal{G})$ to prevent aliasing effects:
\begin{equation}
    \mathcal{L}_s = \frac{1}{N} \sum_{i=1}^{N} \frac{1}{k} \sum_{j \in \texttt{kNN}(\mathcal{G}_i, \mathcal{G})} \mathcal{L}_r(\mathbf{f}_i, \mathbf{f}_j).
\end{equation}

We additionally introduce a regularization loss $\mathcal{L}_{\text{maha}}$. This loss penalizes the Mahalanobis distance~\cite{mclachlan1999mahalanobis} between each per-Gaussian feature vector $\mathbf{f}_i$ and the feature distribution computed from per-patch features from the training dataset used to train $\mathcal{M}^{\text{TASE}}_{\text{enc}}$. Intuitively, this constrains the Gaussian features to remain close to the feature statistics observed during training and thereby discourages splatting-induced artifacts that could otherwise degrade segmentation or editing performance. The overall loss for the reconstruction is then defined as:
\begin{equation}
    \begin{split}
        \mathcal{L}_{\text{splat}} =
        &\ \lambda_\text{l1} \mathcal{L}_\text{l1} +
        \lambda_{\text{SSIM}} \mathcal{L}_{\text{SSIM}} \\
        &+ \lambda_f \mathcal{L}_f + \lambda_s \mathcal{L}_s + \lambda_{\text{maha}} \mathcal{L}_{\text{maha}},
    \end{split}
    \label{eq:splat_loss}
\end{equation}
with the hyperparameters~$\lambda_\text{l1}, \lambda_{\text{SSIM}}, \lambda_f, \lambda_s, \lambda_{\text{maha}}\in\mathbb{R}$.

\subsection{Segmentation}
\label{subsec:segmentation}

For local edits, only a specific object within the scene should be changed and therefore needs to be segmented from the rest of the scene. To enable this, the user marks the object in one or multiple views. We then extract the median feature vector~$\mathbf{f}^\text{anchor}_j$ of the $j^\text{th}$ marked area in the respective rendered feature maps~$\mathcal{F}_h^\text{rendered}$. These act as anchors to identify corresponding Gaussians in 3D space. We compute the cosine similarity between the feature vector of each Gaussian~$\mathbf{f}_i$ and the anchor vectors~$\mathbf{f}^\text{anchor}_k$ and assign a binary segmentation label~$\delta_i$ to each Gaussian $\mathcal{G}_i$ based on a similarity threshold~$\tau_s$:
\begin{equation}
    \delta_i = \bigvee_{k} \frac{\mathbf{f}_i^\top \mathbf{f}^\text{anchor}_k}{||\mathbf{f}_i|| \, ||\mathbf{f}^\text{anchor}_k||} > \tau_s.
\end{equation}

To obtain a segmentation that fully covers the target object, but no unrelated Gaussians, we incorporate spatial consistency into the process. First, the segmentation labels~$\delta_i$ are iteratively propagated from the selected Gaussians to neighboring Gaussians whose $\mathbf{f}_i$ is sufficiently similar. For this we can use a relaxed threshold~$\tau_p < \tau_s$ without running the risk of creating false positives outside the target object. To remove isolated, potentially misclassified Gaussians, we perform a majority vote among their $k$ nearest neighbors and reassign their label accordingly. Optionally, the user can provide negative anchors to explicitly exclude certain regions from the segmentation that might be semantically similar. Throughout, the semantic abstraction level can be controlled by truncating the channels of $\mathbf{f}_i$ and $\mathbf{f}^\text{anchor}_k$.

\subsection{ControlNet Training}
\label{subsec:ctrlnet}

To perform 3D scene editing, we leverage a ControlNet-style~\cite{Zhang2023} latent diffusion model that is conditioned on the TASE~$\mathcal{F}_h$. The ControlNet is based on a pretrained diffusion image transformer (DiT)~$\mathcal{M}^{\text{DiT}}$ and operates in the latent space of a pretrained VAE~$\mathcal{M}^\text{VAE}$. We construct the control branch $\mathcal{M}^{\text{Ctrl}}$ by duplicating $\mathcal{M}^{\text{DiT}}$ and introducing zero-initialized residual connections that add the output of each transformer block $\mathcal{M}^{\text{Ctrl}}_i$ to the corresponding output of $\mathcal{M}^{\text{DiT}}_i$. For each image in a training dataset, we extract the semantic features~$\mathcal{F}_h$ using~$\mathcal{M}^\text{TASE}_{\text{base}}$ and~$\mathcal{M}^\text{TASE}_{\text{enc}}$ and use them as input to~$\mathcal{M}^{\text{Ctrl}}$.

To enable controllable semantic abstraction during image generation, we mask some of the channels of~$\mathcal{F}_h$ with zeros, uniformly selecting one truncation level $t$ from~$\mathcal{T}$ for each training step. Since $\mathcal{F}_h$ is already closer in spatial dimensions and channel depth to the latent space of $\mathcal{M}^\text{VAE}$ than to RGB images, we omit passing $\mathcal{F}_h$ through~$\mathcal{M}^\text{VAE}_\text{enc}$ and instead resize it to the exact dimensions of the image latents via bilinear interpolation. We train~$\mathcal{M}^\text{Ctrl}$ using a flow matching diffusion loss~\cite{Labs2024}, while keeping~$\mathcal{M}^\text{DiT}, \mathcal{M}^\text{TASE}_\text{base}$, and~$\mathcal{M}^\text{TASE}_\text{enc}$ frozen.

Since significant geometry changes during editing can introduce artifacts in rendered views, we propose a finetuning strategy inspired by \mbox{Difix3D+}~\cite{Wu2025c} to mitigate them. We construct a small dataset of clean and corrupted image triplets, including corresponding feature maps~$\mathcal{F}_h$ exhibiting the same artifacts. Using the AnySplat feed-forward splatting model~\cite{Jiang2025}, we generate 3DGS scenes from single-view images and obtain the feature vectors~$\mathbf{f}_i$ by projecting~$\mathcal{F}_h$ onto the Gaussians. Subsequently, we randomly perturb~$\boldsymbol{\mu}_i, \mathbf{s}_i$, and~$\mathbf{q}_i$ to create corrupted renders of the scene. To emulate the local editing scenario, we apply a stronger perturbation to a spherical region around a randomly selected splat. As in Difix3D+, we treat the corrupted image as a noisy image at an intermediate denoising timestep ($t_\text{corrupted} = 200$), pass it through~$\mathcal{M}^\text{VAE}_\text{enc}$ and add additional noise~$\mathcal{N}(\mathbf{0}, \mathbf{I})~$ corresponding to the timestep~$t_\text{added}$ to retrieve the noisy image latent. The assumed timestep~$t_\text{assumed}$ of the noisy latent then computes as:
\begin{equation}
    t_\text{assumed} = t_\text{corrupted} + \frac{t_\text{total} - t_\text{corrupted}}{t_\text{total}} t_\text{added},
    \label{eq:timestep}
\end{equation}
with~$t_\text{total}$ being the total number of diffusion steps used during training. We finetune both~$\mathcal{M}^\text{DiT}$ and~$\mathcal{M}^\text{Ctrl}$, again using a flow matching diffusion loss~\cite{Labs2024}.

\subsection{3D Scene Editing}
\label{subsec:3d_editing}

3D scene editing can be done globally, modifying all the Gaussians~$\mathcal{G}$ to change, for example, the weather or the lighting of the scene. For this, we start by sampling a set of camera poses from the training poses that were used to optimize the original scene. We then render feature maps~$\mathcal{F}_h^\text{render}\in\mathbb{R}^{H \times W \times N_h}$ from these poses and generate a set of novel views $\mathbf{I}^\text{novel}$ using the ControlNet $(\mathcal{M}^\text{DiT}, \mathcal{M}^\text{Ctrl})$. We optimize $\mathcal{G}$ to reflect the content of $\mathbf{I}^\text{novel}$, using the loss~$\mathcal{L}_\text{splat}$ defined in \eqref{eq:splat_loss}. Subsequently, we sample a new set of poses and repeat the process. The number of iterations of this editing loop is an editing hyperparameter that depends on the desired amount of change in the scene.

For the first iterations of the editing loop, we truncate $\mathcal{F}_h^\text{render}$ by masking $N_h-t$ of the $N_h$ channels with zeros, to allow the diffusion model to align more closely with the text prompt, rather than with the current scene content. Over the course of the edit, we increase the number of retained channels $t$. During later iterations, we also integrate the current RGB renders~$\mathbf{I}^\text{render}$ into the image generation in the same way as during the finetuning stage of the ControlNet training described in \secref{subsec:ctrlnet}. We create schedules for $t$, for the amount of noise~$\mathcal{N}(\mathbf{0}, \mathbf{I})$ added to~$\mathbf{I}^\text{render}$ and for the learning rates of the parameters that define the geometry of the scene~$(\boldsymbol{\mu}, \mathbf{s}, \mathbf{q})$. These schedules are editing hyperparameters that also depend on the desired amount of change in the scene.

The 3D scene can also be edited locally, changing the appearance and geometry only of a specified object in the scene. To this end, we first select a subset of Gaussians~$\mathcal{G}_S\subseteq\mathcal{G}$ with the approach described in \secref{subsec:segmentation}. We then sample novel object centric camera poses focused on a bounding ellipsoid that we construct around~$\mathcal{G}_S$. In addition to~$\mathcal{F}_h$ we also render a segmentation mask~$\mathbf{M}^\text{render}$ that we first dilate and then use as an inpainting mask for the image generation. For local edits, we also add regularization terms as well as weight decay for the opacity~$\alpha_i$ and the base color~$\mathbf{\hat{c}}_{0, i}$ to allow for the removal of existing geometry, mitigating over-densification and over-saturation.

\section{Experimental Evaluation}
\label{sec:exp}
Our experiments evaluate the effectiveness of truncation-aware semantic embeddings (TASE) for 3D scene editing. In \secref{sec:edit_exp}, we show that our semantic embeddings enable controllable 3D scene editing, including substantial geometric modifications. In \secref{sec:trade-off}, we demonstrate how channel truncation can be used to control semantic abstraction during 2D image generation conditioned on TASE. Finally, in \secref{sec:ablations}, we show the individual effects of: our specific method to create a truncation-aware embedding, of our equivariance loss aimed at removing 2D positional bias, and of the Difix3D+~\cite{Wu2025c} finetuning strategy.

\subsection{Implementation Details}

We use a pretrained DINOv3 model~\cite{Simeoni2025} as the backbone~$\mathcal{M}^\text{TASE}_\text{base}$, which has been shown to contain rich semantic information. For the encoder~$\mathcal{M}^\text{TASE}_\text{enc}$ and the decoder~$\mathcal{M}^\text{TASE}_\text{dec}$, we use a single transformer block. The hidden channel dimensionality~$N_h$ of the autoencoder is set to 64 and we use~$\mathcal{T} = [2^1, 2^2, ..., 2^6]$ as the truncation levels for the Matryoshka representation learning training. The diffusion model uses FLUX.1[dev]~\cite{Labs2024} as~$\mathcal{M}^\text{DiT}$. We train both~$\mathcal{M}^\text{TASE}_\text{enc}$ and the control branch~$\mathcal{M}^\text{Ctrl}$ on \mbox{ImageNet-1k}~\cite{Deng2009} for a single epoch, using a learning rate of~$10^{-4}$ for~$\mathcal{M}^\text{TASE}_\text{enc}$ and~$\mathcal{M}^\text{TASE}_\text{dec}$. For the ControlNet we use a cosine learning rate schedule with a maximum learning rate of~$10^{-5}$. We train $\mathcal{M}^\text{DiT}$ and $\mathcal{M}^\text{Ctrl}$ randomly using an empty prompt or a prompt constructed from the class labels associated with the image as text guidance. For the finetuning stage, we create 50,000 clean/corrupted image pairs from \mbox{ImageNet-1k}'s train set and finetune for 4 epochs with a reduced maximum learning rate of~$10^{-6}$ and~$10^{-7}$ for~$\mathcal{M}^\text{Ctrl}$ and the~$\mathcal{M}^\text{DiT}$ respectively. More detail on the implementation, datasets and the used hyperparameters is provided in the supplement.  

\subsection{3D Scene Editing Capabilities}
\label{sec:edit_exp}

\begin{figure}[t]
    \centering
    \def\imgw{0.1875\textwidth}
    \def\imgtrim{0.044\textwidth}
    \def\imgh{0.1\textwidth}
    \begin{tabular}{{>{\centering\arraybackslash}m{\imgw}>{\centering\arraybackslash}m{0.17\textwidth}>{\centering\arraybackslash}m{\imgw}>{\centering\arraybackslash}m{\imgw}>{\centering\arraybackslash}m{\imgw}}}
        \toprule
        Original                                                                                                                                                                                            & Prompt                                                                                                                                                  & Ours                                                                                                                                                                                                                  & DGE~\cite{Chen2025d}                                                                                                                                                                                                & GE~\cite{YCZCCZFWXYYWZCLYHLGL2023}                                                                                                                                                                                                        \\
        \midrule

        \includegraphics[width=\imgw, trim=75 0 0 0, clip]{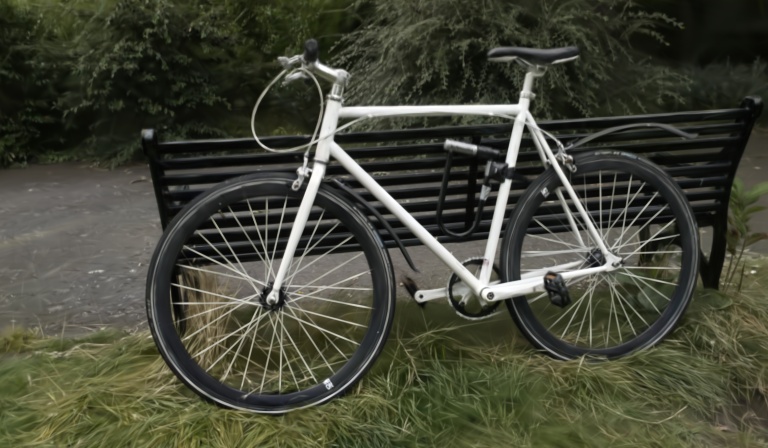} \includegraphics[width=\imgw, trim=75 0 0 0, clip]{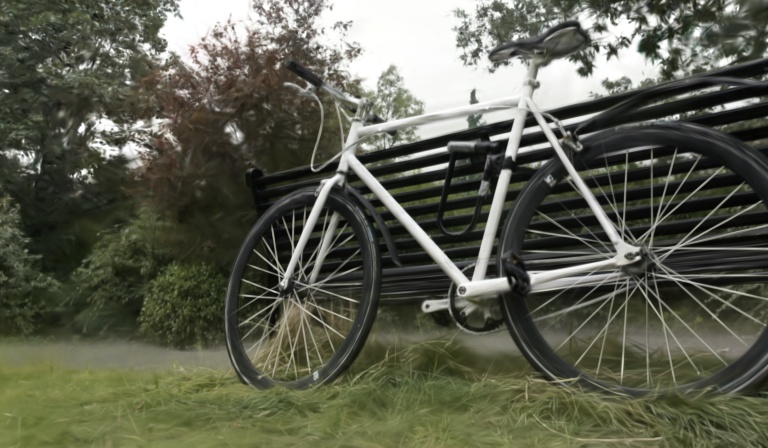}                                                   & \linespread{0.75}\selectfont \textit{\scriptsize a bright orange enduro dirt bike with a low black saddle, knobby tires, a big chrome front suspension fork, disk brakes and big orange fenders} & \includegraphics[width=\imgw, trim=75 0 0 0, clip]{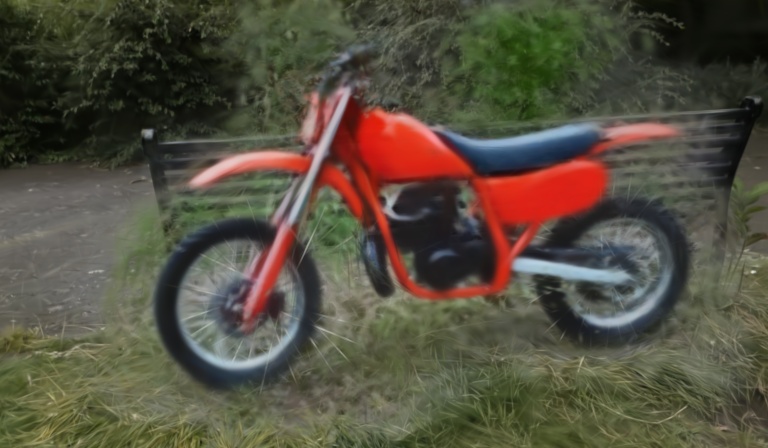}
        \includegraphics[width=\imgw, trim=75 0 0 0, clip]{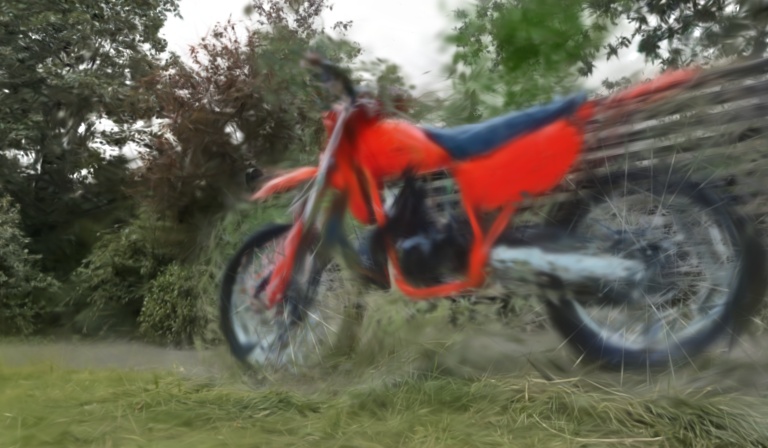}                                                                                                                     & \includegraphics[width=\imgw, trim=75 0 0 0, clip]{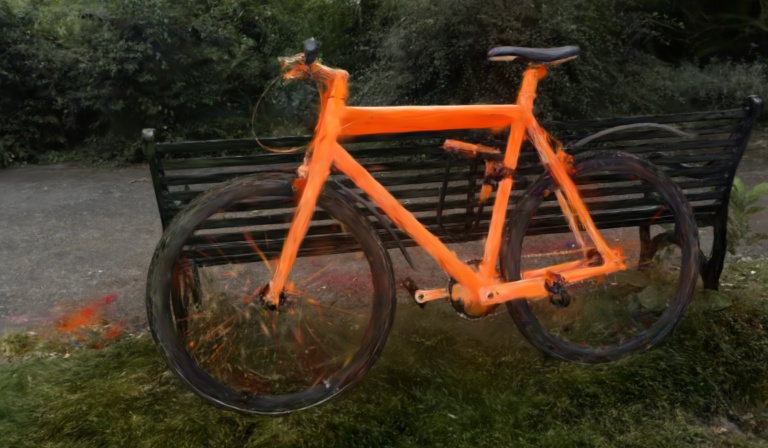}
        \includegraphics[width=\imgw, trim=75 0 0 0, clip]{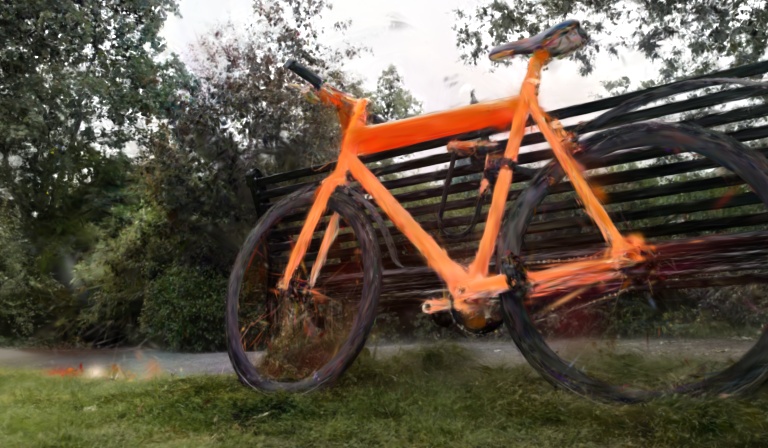}                                                                                                                      & \includegraphics[width=\imgw, trim=75 0 0 0, clip]{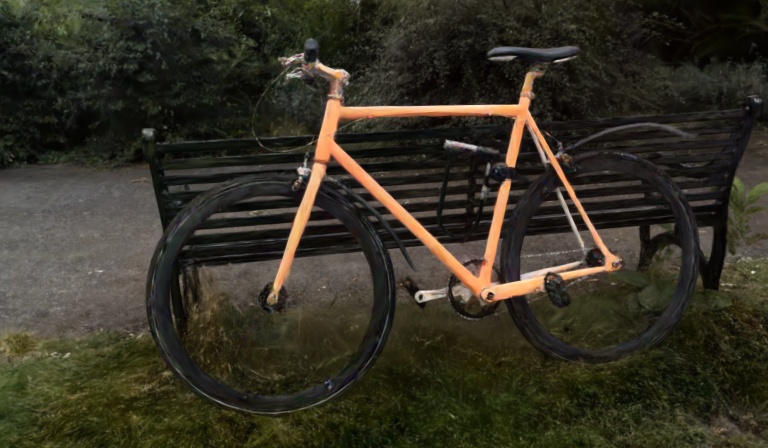}
        \includegraphics[width=\imgw, trim=75 0 0 0, clip]{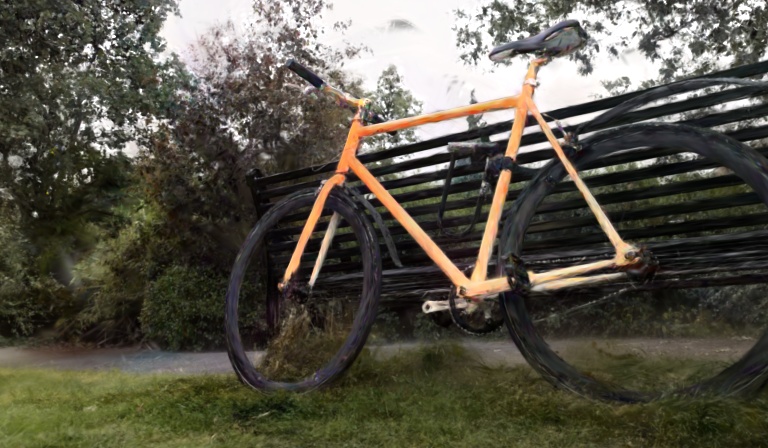}                                                                                                                                                                                                                                                                                                                                                                                                                                                                                                                                                                                                                                                                                                                                                                                                                                                                                                                                                               \\
        \midrule
        \includegraphics[width=\imgw, trim=25 0 50 0, clip]{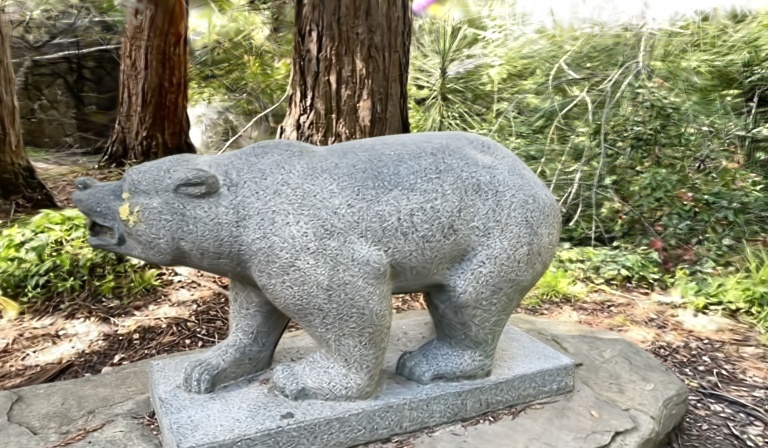} \includegraphics[width=\imgw, trim=50 0 25 0, clip]{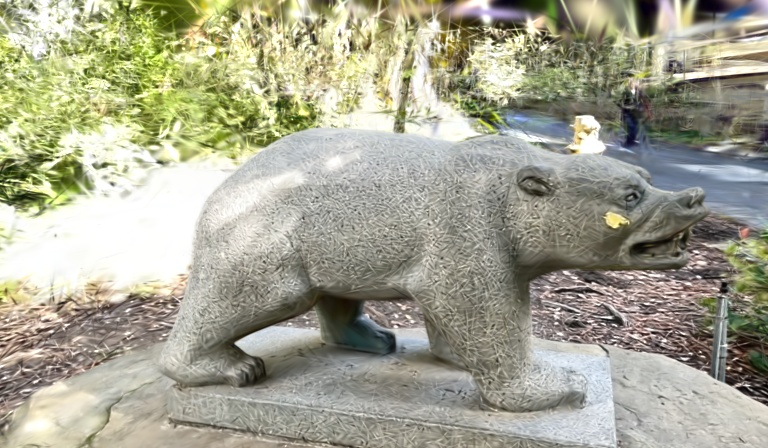}                                                    & \linespread{0.75}\selectfont \textit{\scriptsize a rottweiler dog with a shiny black coat and brown markings}                                                                                     & \includegraphics[width=\imgw, trim=25 0 50 0, clip]{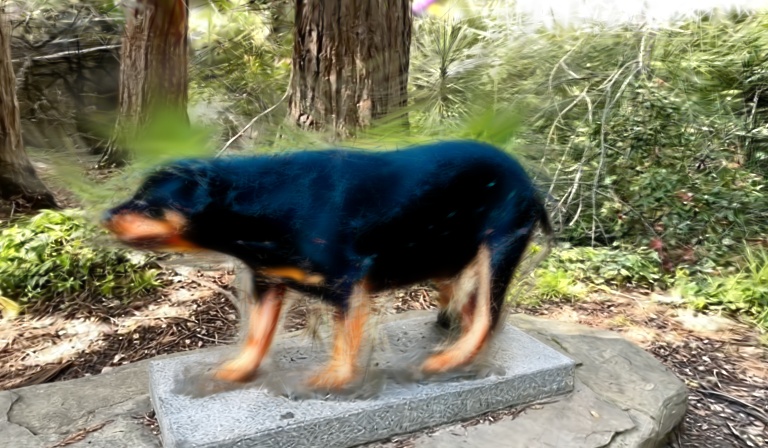}
        \includegraphics[width=\imgw, trim=50 0 25 0, clip]{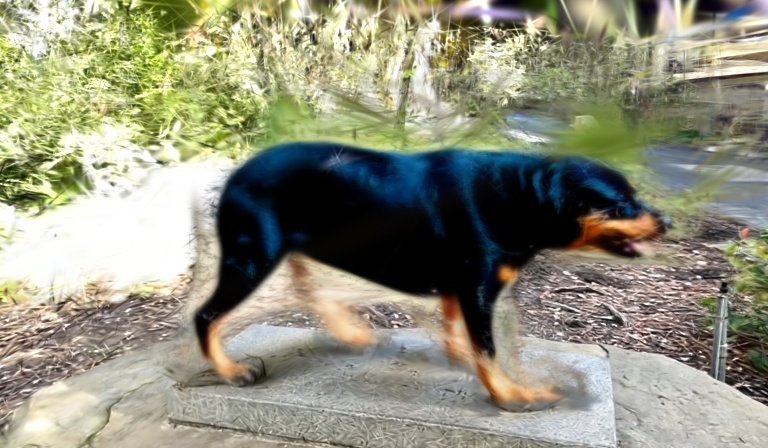}                                                                                                                  & \includegraphics[width=\imgw, trim=25 0 50 0, clip]{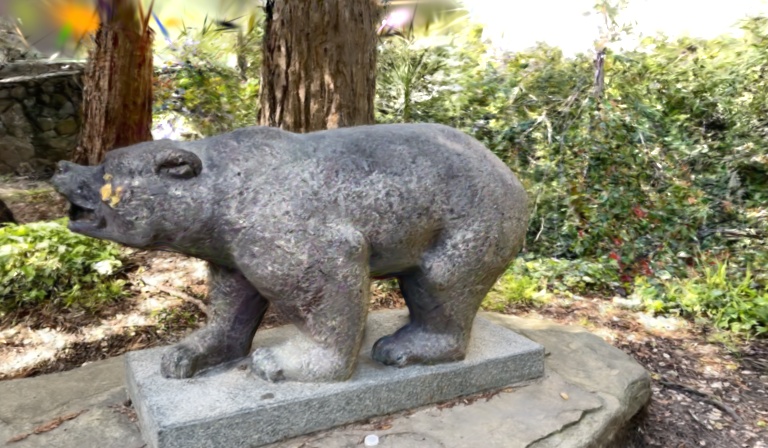}
        \includegraphics[width=\imgw, trim=50 0 25 0, clip]{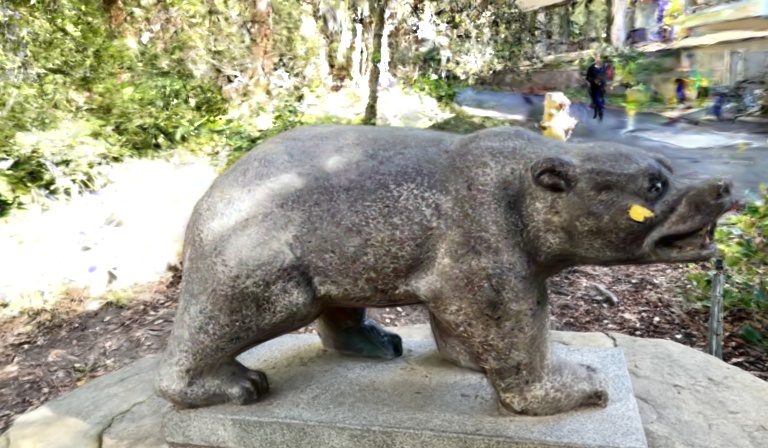}                                                                                                                   & \includegraphics[width=\imgw, trim=25 0 50 0, clip]{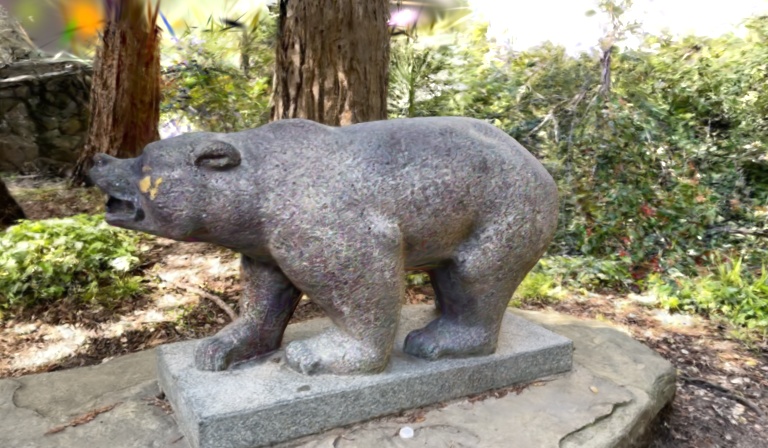}
        \includegraphics[width=\imgw, trim=50 0 25 0, clip]{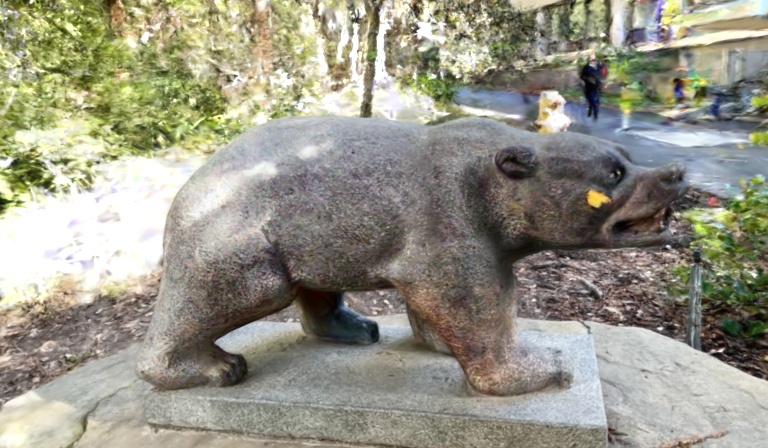}                                                                                                                                                                                                                                                                                                                                                                                                                                                                                                                                                                                                                                                                                                                                                                                                                                                                                                                                                            \\
        \midrule
        \includegraphics[width=\imgw, trim=25 0 50 0, clip]{pics/baseline_comparisons/bear/orig_0.jpg} \includegraphics[width=\imgw, trim=50 0 25 0, clip]{pics/baseline_comparisons/bear/orig_1.jpg}                                                    & \linespread{0.75}\selectfont \textit{\scriptsize a tiger with bold orange fur and striking black stripes, fierce eyes, and a powerful build}                                                     & \includegraphics[width=\imgw, trim=25 0 50 0, clip]{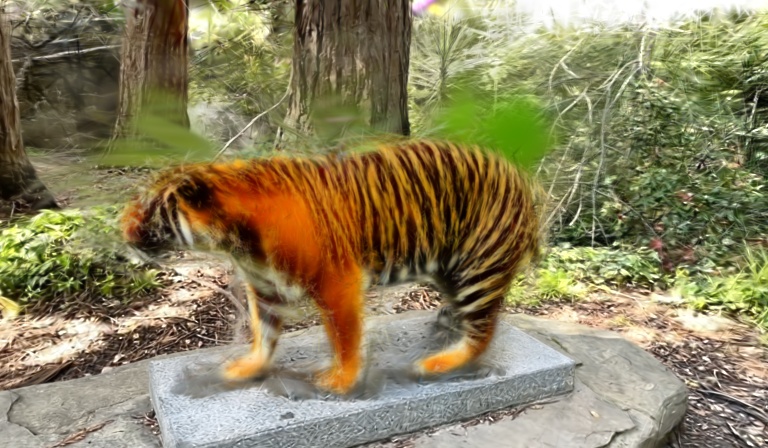}
        \includegraphics[width=\imgw, trim=50 0 25 0, clip]{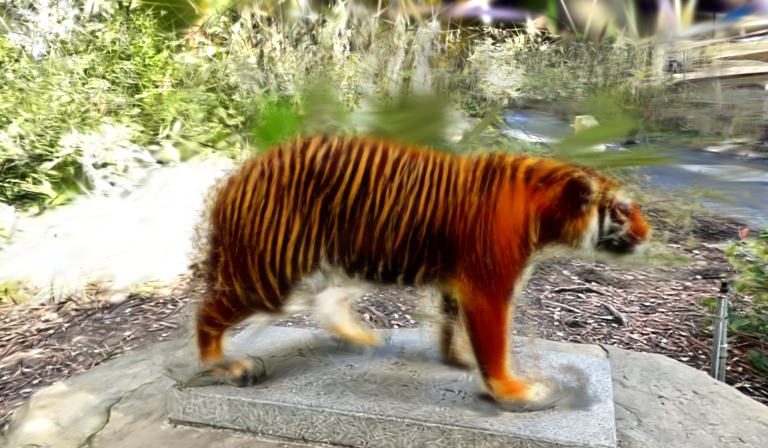}                                                                                                                      & \includegraphics[width=\imgw, trim=25 0 50 0, clip]{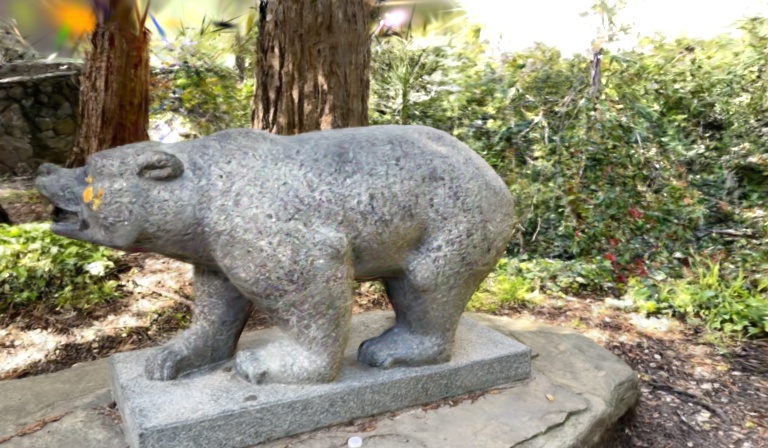}
        \includegraphics[width=\imgw, trim=50 0 25 0, clip]{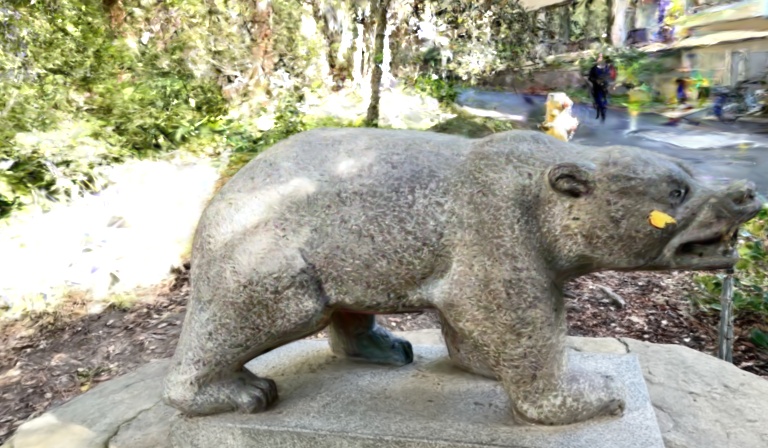}                                                                                                                       & \includegraphics[width=\imgw, trim=25 0 50 0, clip]{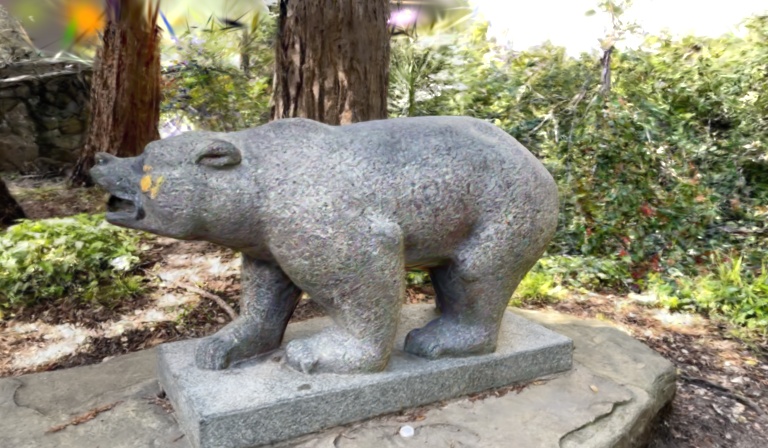}
        \includegraphics[width=\imgw, trim=50 0 25 0, clip]{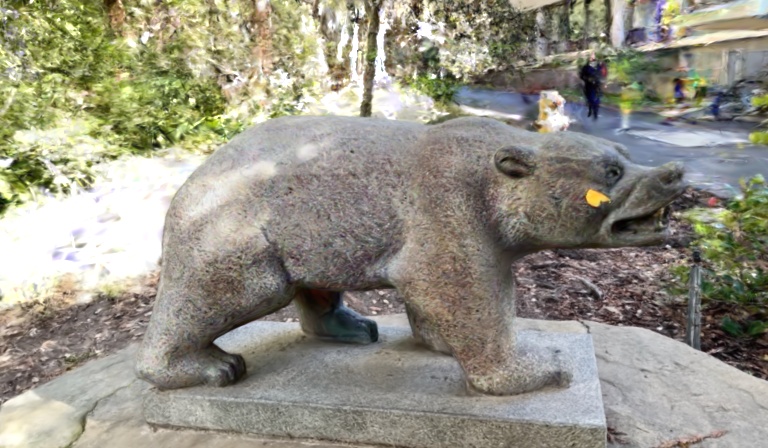}                                                                                                                                                                                                                                                                                                                                                                                                                                                                                                                                                                                                                                                                                                                                                                                                                                                                                                                                                                \\
        \midrule
        \includegraphics[height=\imgh, trim=198 0 198 0, clip]{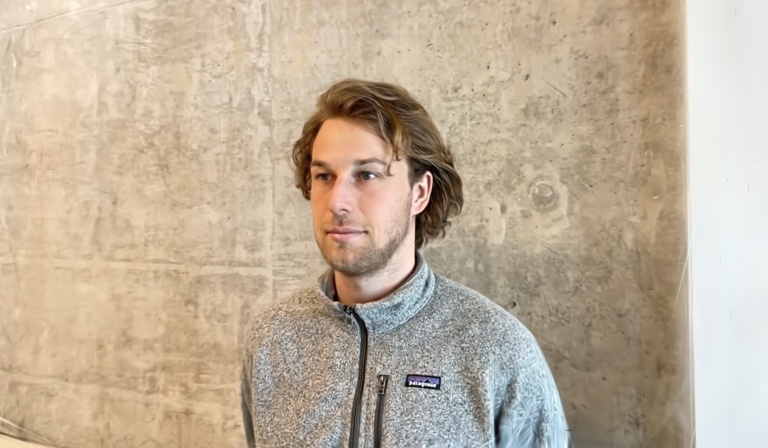} \includegraphics[height=\imgh, trim=198 0 198 0, clip]{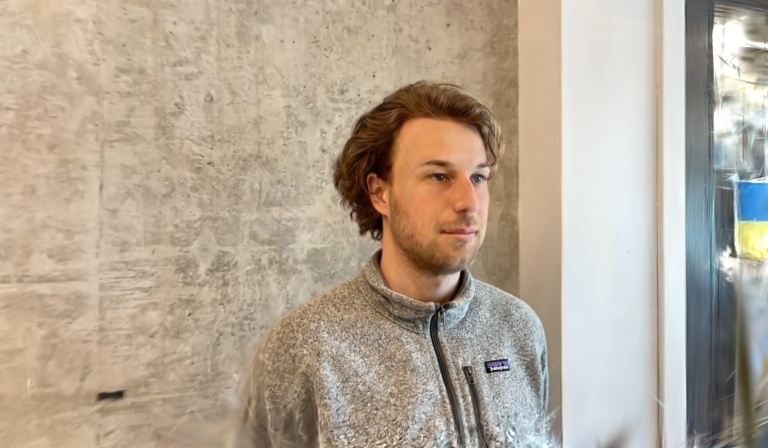}  & \linespread{0.75}\selectfont \textit{\scriptsize albert einstein with wild gray hair and mustache}                                                                                               & \includegraphics[height=\imgh, trim=198 0 198 0, clip]{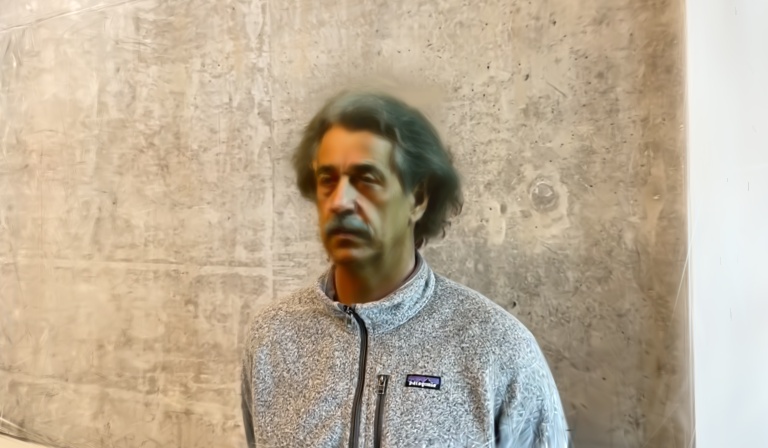} \includegraphics[height=\imgh, trim=198 0 198 0, clip]{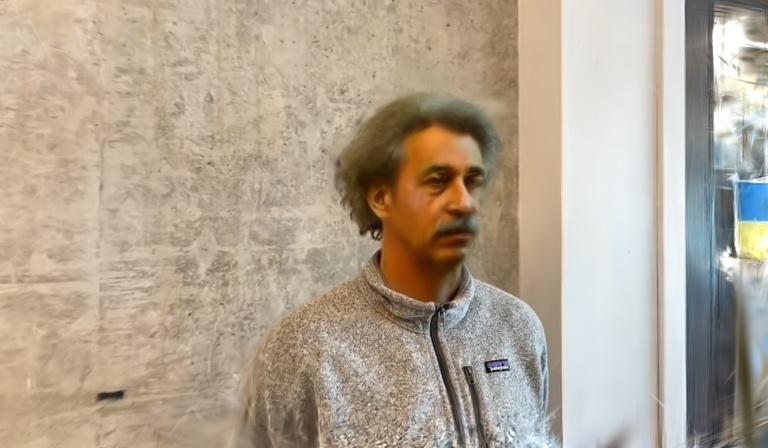} & \includegraphics[height=\imgh, trim=198 0 198 0, clip]{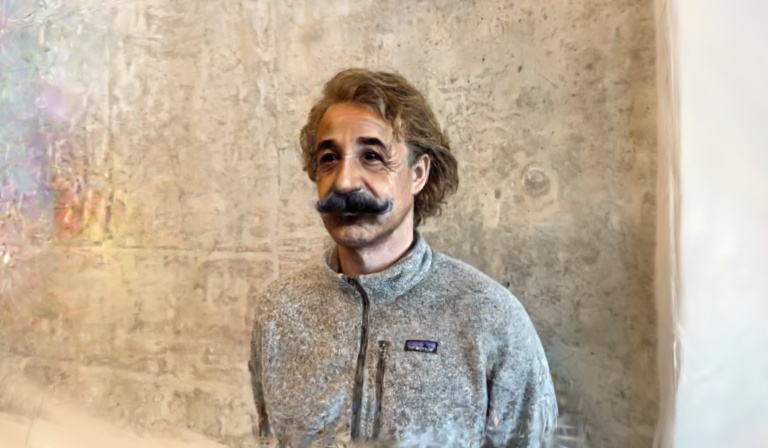} \includegraphics[height=\imgh, trim=198 0 198 0, clip]{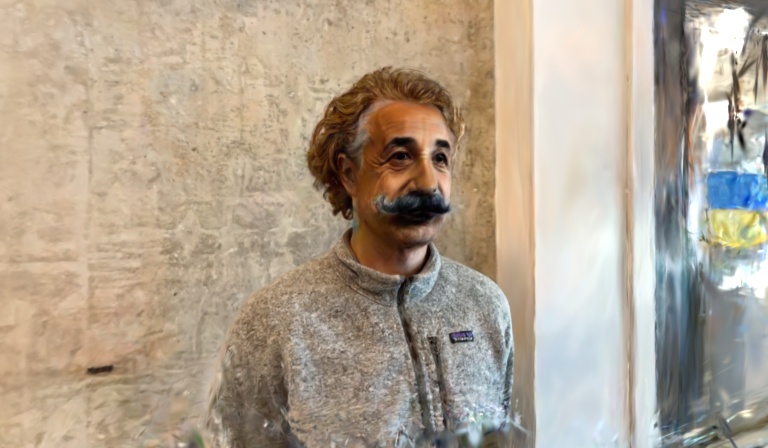} & \includegraphics[height=\imgh, trim=198 0 198 0, clip]{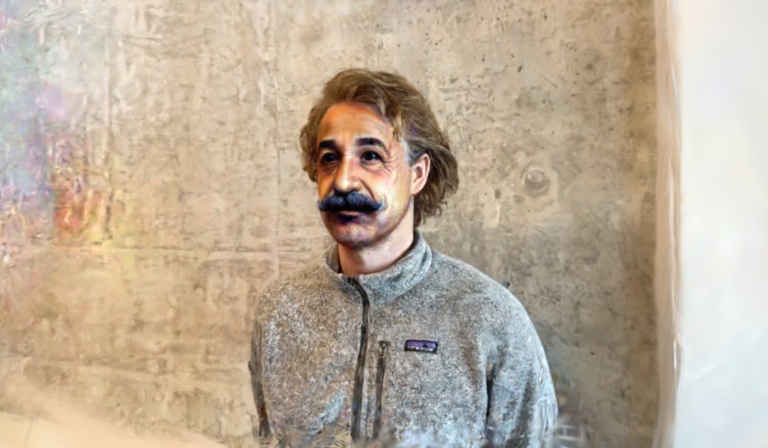} \includegraphics[height=\imgh, trim=198 0 198 0, clip]{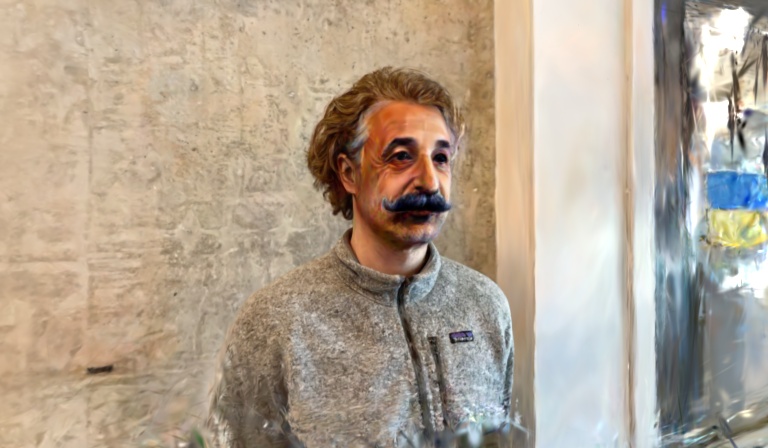} \\

        \bottomrule
    \end{tabular}
    \caption{Qualitative comparison of our editing results against the state-of-the-art 3D scene editing methods Direct Gaussian Editing~(DGE)~\cite{Chen2025d} and GaussianEditor~(GE)~\cite{YCZCCZFWXYYWZCLYHLGL2023} for local 3D scene editing.}
    \label{fig:editing}
    \vspace{-0.4cm}
\end{figure}

To evaluate our 3D scene editing capabilities, we perform diverse edits on multiple datsets. We compare against Direct~Gaussian~Editing~(DGE)~\cite{Chen2025d} and GaussianEditor~(GE)~\cite{YCZCCZFWXYYWZCLYHLGL2023} as state-of-the-art baselines for 3DGS scene editing. While some methods use a pretrained 3D generation model to replace objects within the scene~\cite{Wang2024e, Xiao2025}, this is not equivalent to manipulating the original object geometry and only changing it, where it is necessary. We therefore compare our results only to methods that change the contents of the original scene.

\figref{fig:editing} provides qualitative examples, which show that our method is able to perform local edits with high visual fidelity and consistency across views. As seen, when large modifications to the geometry are required, e.g., changing the bicycle to a motorcycle, the baseline methods change the color of the object towards the target color to varying degrees, but fail to change the geometry sufficiently. When changing the bear into a different kind of animal, requiring a drastic change in appearance and geometry, the baselines fail to reflect the requested changes. In contrast, our approach arrives at a high quality result. This highlights the effectiveness of using TASE as a control signal, enabling substantial edits to 3D scenes.

\begin{figure}[t]
    \centering
    \def\imgw{0.134\textwidth}
    \begin{tabular}{cccc}
        \toprule
                                                                                                                                                          &                      & Prompt                               \\
        \cmidrule(lr){2-4}
        Original                                                                                                                                          & \textit{In the snow} & \textit{At sunset} & \textit{In the rain} \\
        \midrule
        \includegraphics[width=\imgw]{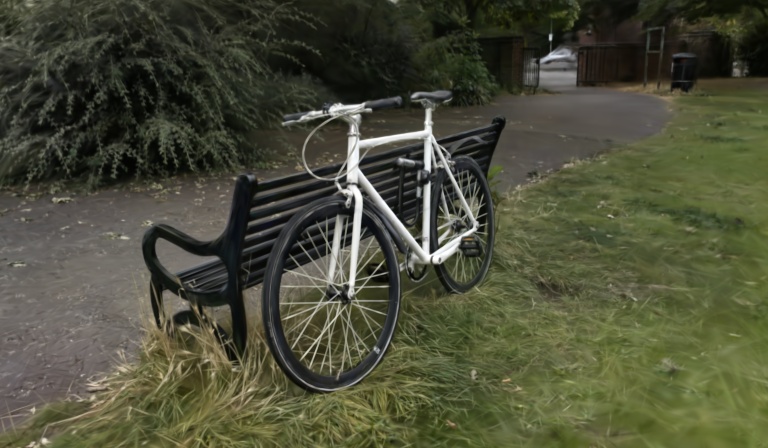}                                                                                  &
        \includegraphics[width=\imgw]{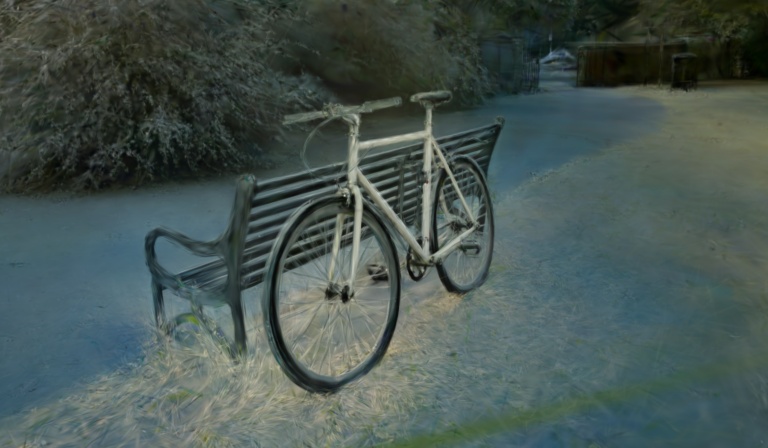} \includegraphics[width=\imgw]{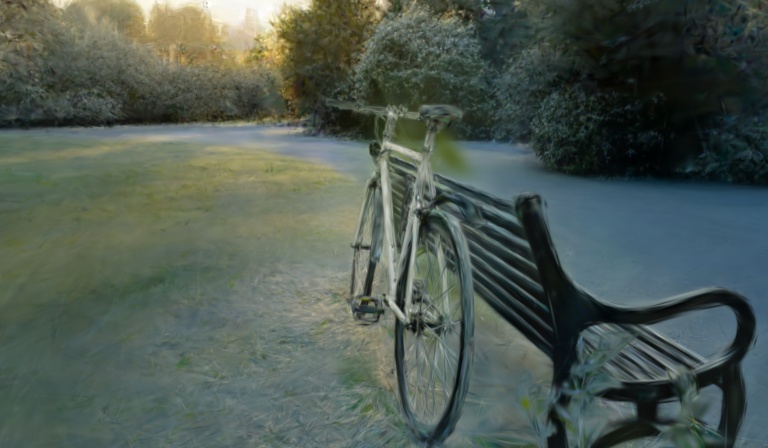}       &
        \includegraphics[width=\imgw]{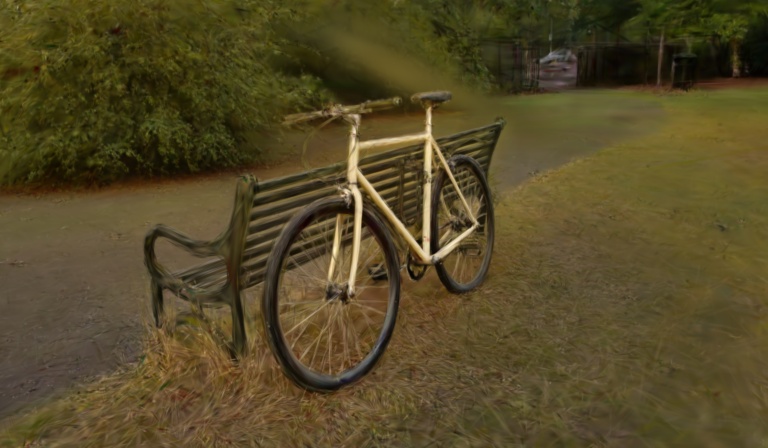} \includegraphics[width=\imgw]{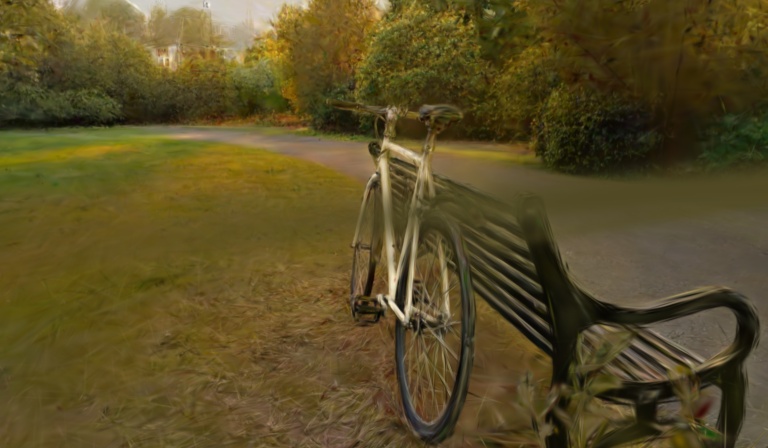}     &
        \includegraphics[width=\imgw]{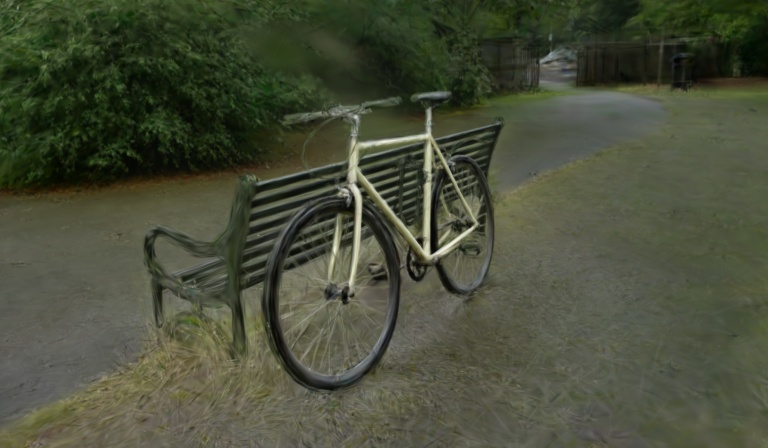} \includegraphics[width=\imgw]{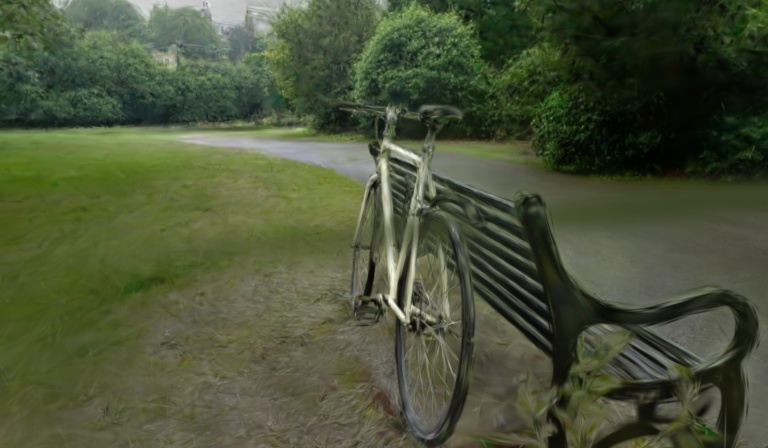}                                                                     \\
        \includegraphics[width=\imgw]{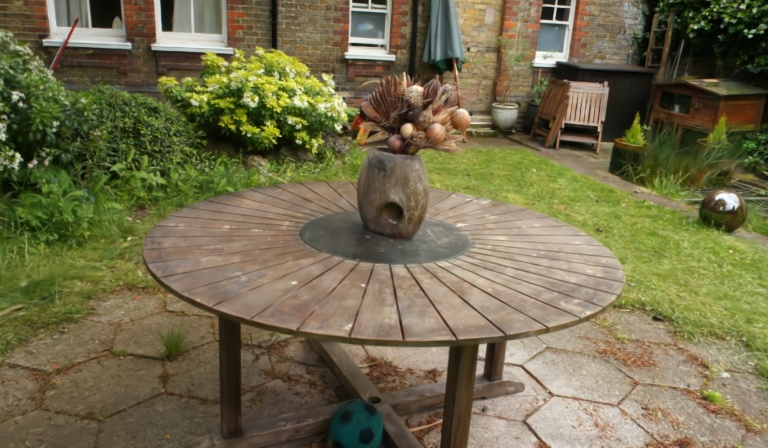}                                                                                &
        \includegraphics[width=\imgw]{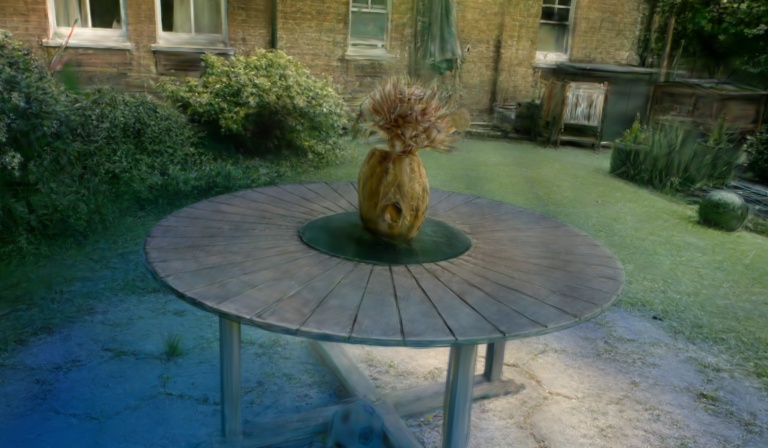} \includegraphics[width=\imgw]{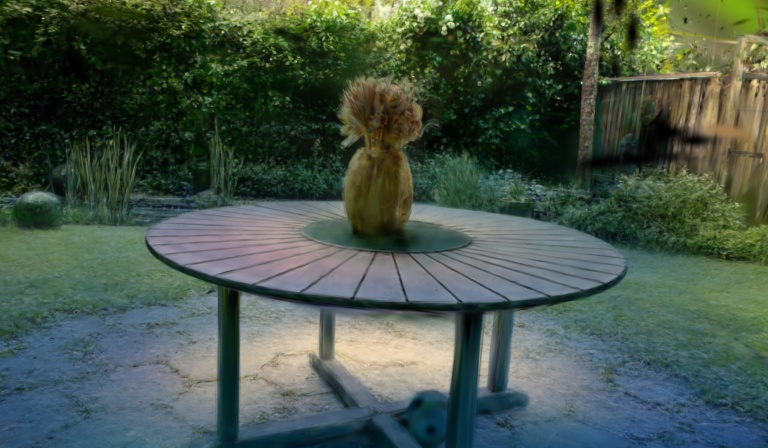}   &
        \includegraphics[width=\imgw]{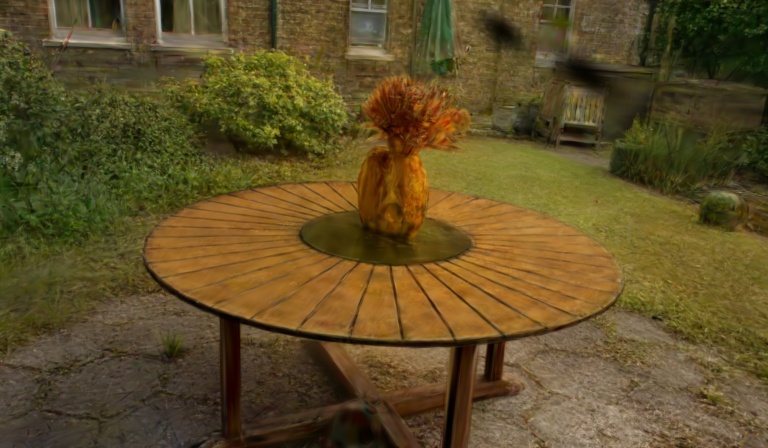} \includegraphics[width=\imgw]{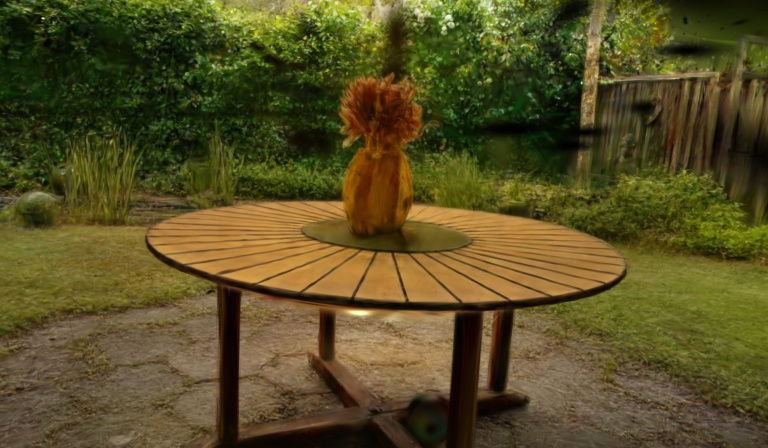} &
        \includegraphics[width=\imgw]{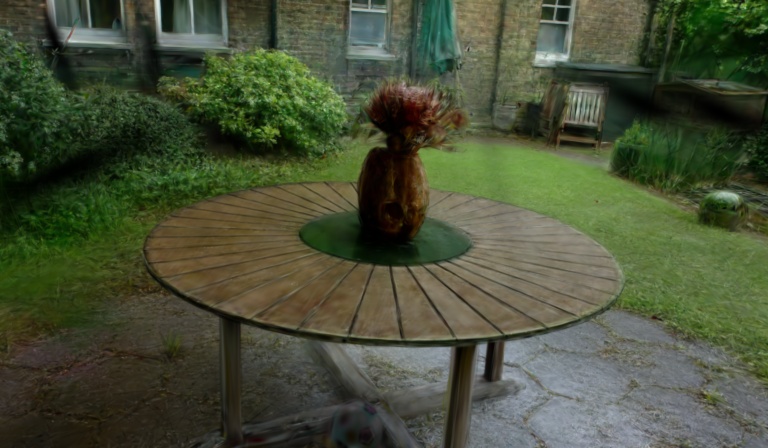} \includegraphics[width=\imgw]{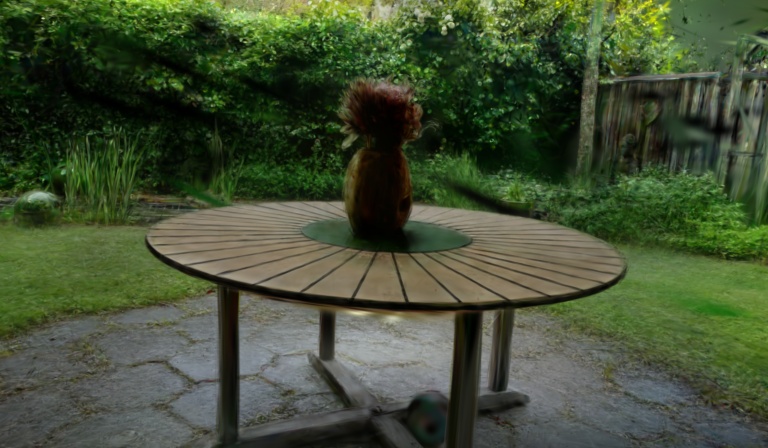}                                                                 \\
        \bottomrule
    \end{tabular}
    \caption{Qualitative results of global 3D scene edits performed with our method showing that our method succeeds in globally changing the appearance of the scene.}
    \label{fig:global_edits}
    \vspace{-0.1cm}
\end{figure}

\begin{table}[t]
    \caption{Quantitative comparison of our local scene editing results against state-of-the-art 3D scene editing methods (CLIP Directional Similarity).}
    \centering
    \def\colw{2.25cm}
    \begin{tabular}{l>{\raggedleft\arraybackslash}p{\colw}>{\raggedleft\arraybackslash}p{\colw}>{\raggedleft\arraybackslash}p{\colw}}
        \toprule
                                               &                           & \multicolumn{2}{c}{User Study}                     \\
        \cmidrule(lr){3-4}
                                               & \shortstack{CLIP \\ dir. sim.} \shortstack{\vspace{0.35em}$\uparrow$} &  \shortstack{\vspace{0.35em}Geometry $\uparrow$} &  \shortstack{\vspace{0.35em}Appearance $\uparrow$} \\
        \midrule
        DreamCatalyst~\cite{Kim2025a}                 & 0.0985               & -                    & -                    \\
        EditSplat~\cite{Lee_2025_CVPR}                & 0.1492               & -                    & -                    \\
        GaussianEditor~\cite{YCZCCZFWXYYWZCLYHLGL2023} & 0.1179               & 16.7\%               & 8.8\%                \\
        DGE~\cite{Chen2025d}                          & 0.1247               & 19.1\%               & 18.7\%               \\
        Ours                                          & \textbf{0.1542}      & \textbf{64.3\%}      & \textbf{72.5\%}      \\
        \bottomrule
    \end{tabular}
    \label{tab:metrics}
    \vspace{-0.2cm}
\end{table}

To quantify the advantages of our method, in \tabref{tab:metrics} we compute the CLIP dir. sim.~\cite{gal2022stylegan} metric, which measures how well the edit direction aligns with the intent specified by the text prompt. Our method archives the highest score, showing that our edits align much better with the provided text prompt. Additionally, we conduct a user study to assess the user preference of changes in geometry and appearance in the edited scenes. The details of the user study are described in the supplement. The results in \tabref{tab:metrics} demonstrate that the edits performed with our method are preferred by users both in terms of geometry and in terms of appearance.

\figref{fig:global_edits} presents qualitative results for global edits, where the entire scene is modified according to a text prompt. Our method is able to produce high-quality results that reflect the desired changes in the scene, such as seasonal, weather or lighting changes.

\subsection{Trading off Specificity and Abstraction}
\label{sec:trade-off}

\begin{figure}[t]
    \def\imgw{0.19\textwidth}
    \centering
    \begin{subfigure}[b]{\imgw}
        \includegraphics[width=\linewidth]{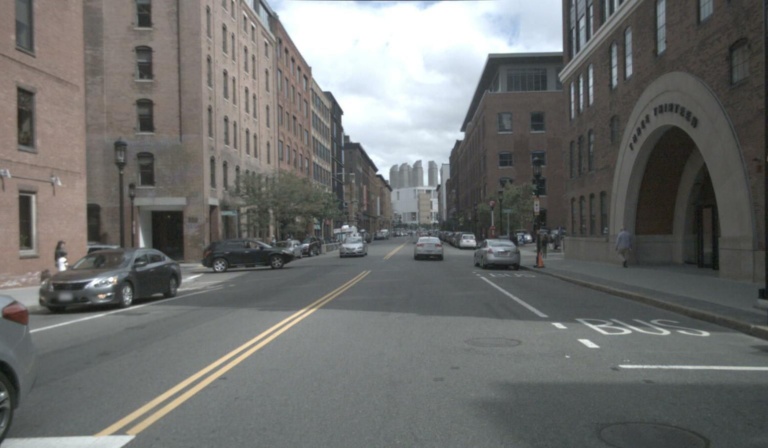}
        \caption{Original}
        \label{fig:edit_o}
    \end{subfigure}
    \begin{subfigure}[b]{\imgw}
        \includegraphics[width=\linewidth]{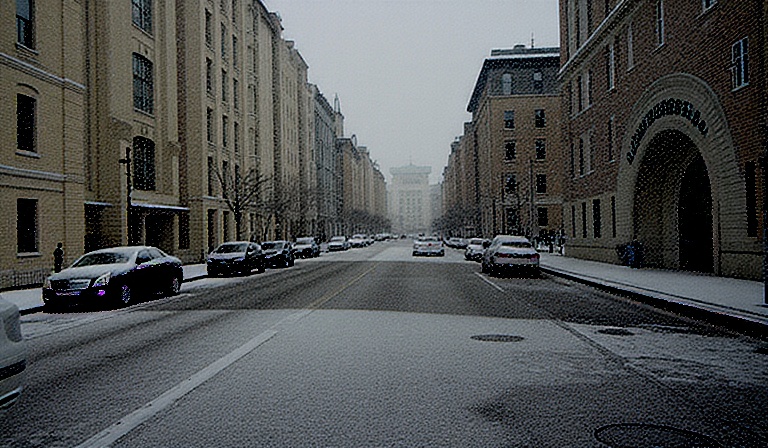}
        \caption{64 dims}
        \label{fig:edit_64}
    \end{subfigure}
    \begin{subfigure}[b]{\imgw}
        \includegraphics[width=\linewidth]{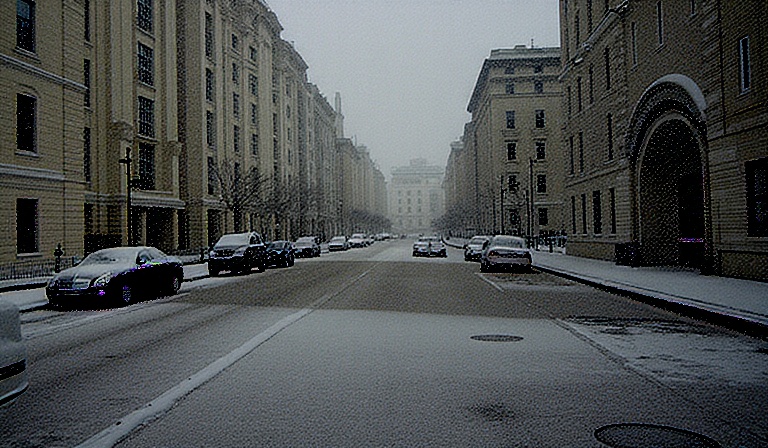}
        \caption{8 dims}
        \label{fig:edit_8}
    \end{subfigure}
    \begin{subfigure}[b]{\imgw}
        \includegraphics[width=\linewidth]{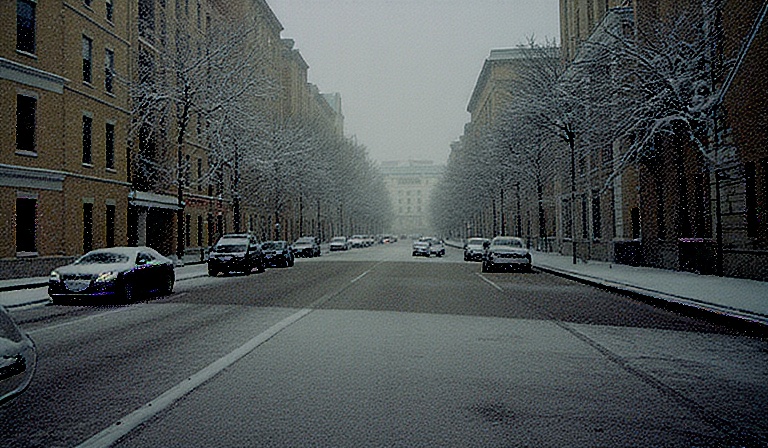}
        \caption{4 dims}
        \label{fig:edit_4}
    \end{subfigure}
    \begin{subfigure}[b]{\imgw}
        \includegraphics[width=\linewidth]{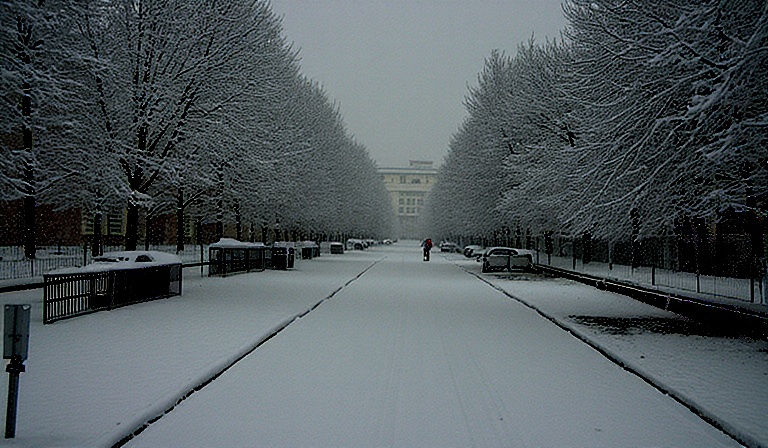}
        \caption{2 dims}
        \label{fig:edit_2}
    \end{subfigure}
    \caption{Generated images with the text prompt: "\textit{in the snow}" with TASE guidance truncated to different levels, showing close alignment to the semantic layout of the original scene for a large number of retained channels and close alignment to the text prompt for a low number of retained channels.}
    \label{fig:tradeoff}
    \vspace{-0.1cm}
\end{figure}

When generating images conditioned on TASE, the number of retained channels controls the balance between prompt alignment and adherence to the scene structure. Using fewer dimensions yields more diverse images that follow the text prompt, while retaining the full embedding mostly preserves the semantic layout of the original image without reproducing it exactly. This behavior is illustrated in Fig.~\ref{fig:tradeoff}: with 2 dimensions, the generated image only loosely follows the original layout but strongly reflects the prompt \emph{“in the snow”}. With 64 dimensions, the result preserves more of the scene structure in the source image, yet introduces little snow. Intermediate truncation levels provide a smooth transition between these extremes. Additional examples are provided in the supplement.  

\begin{table}[t]
    \caption{Ablation study for the method of dimensionality reduction ($\mathcal{M}^\text{TASE}$ vs. PCA), the consistency loss~$\mathcal{L}_{\text{cons}}$ and the finetuning stage (FT). The finetuning only affects the editing.}
    \label{tab:ablation}
    \centering
    \begin{tabular}{lccccc}
        \toprule
        & \multicolumn{3}{c}{Reconstruction} & \multicolumn{1}{c}{Editing} \\
        \cmidrule(lr){2-4} \cmidrule(lr){5-5} 
        & PSNR $\uparrow$ & SSIM $\uparrow$ & $\mathcal{L}_{f}$ $\downarrow$ & CLIP dir. sim. $\uparrow$ \\
        \midrule
        $\text{PCA}$                                               & 22.64           & 0.73           & 0.0879           & 0.1067 \\
        $\text{PCA}+\text{FT}$                                     & ---             & ---            & ---              & 0.1365 \\
        $\mathcal{M}^\text{TASE}$                                  & 23.89           & 0.76           & 0.1016           & 0.0852 \\
        $\mathcal{M}^\text{TASE}+\text{FT}$                        & ---             & ---            & ---              & 0.0905 \\
        $\mathcal{M}^\text{TASE}+\mathcal{L}_\text{cons}$          & $\mathbf{24.64}$ & $\mathbf{0.78}$ & $\mathbf{0.0120}$ & 0.1385 \\
        $\mathcal{M}^\text{TASE}+\mathcal{L}_\text{cons}+\text{FT}$ (Ours) & ---    & ---            & ---              & $\mathbf{0.1542}$ \\
        \bottomrule
    \end{tabular}
    \vspace{-0.2cm}
\end{table}

\subsection{Ablations}
\label{sec:ablations}
To assess the impact of the equivariance loss $\mathcal{L}_\text{eqv}$ described in \secref{subsec:TASE} on editing quality, we compare our full model to a variant omitting $\mathcal{L}_\text{eqv}$ during the training of $\mathcal{M}^\text{TASE}$. We further justify our truncation-aware embedding produced by $\mathcal{M}^\text{TASE}$ under a Matryoshka representation learning scheme (\secref{subsec:TASE}) by comparing it to a PCA of the base features $\mathcal{F}_o$ computed over 50k training samples. Finally, we evaluate all ablations before and after Difix3D+~\cite{Wu2025c} finetuning.

\tabref{tab:ablation} shows the CLIP dir. sim. \cite{gal2022stylegan} for the same local edits used for comparison to the baselines in \secref{sec:edit_exp}, as well as the reconstruction metrics PSNR, SSIM~\cite{Wang2004} and the feature loss $\mathcal{L}_f$ from \eqref{eq:feature_loss} for held out views of the original scene. The much lower $\mathcal{L}_f$ when using $\mathcal{L}_\text{eqv}$ during the training of $\mathcal{M}^\text{TASE}$ indicates that $\mathcal{L}_\text{eqv}$ is important for features that are free from 2D positional bias and therefore multi-view consistent. The large difference in the final editing quality indicates that this property is crucial when using semantic embeddings as a control signal for 3D scene editing. 

\begin{figure}[t]
    \def\imgw{0.18\textwidth}
    \centering
    \begin{subfigure}[b]{\imgw}
        \includegraphics[width=\linewidth, clip, trim=50 0 50 0]{pics/baseline_comparisons/bike/orig_0.jpg}
        \subcaption{}
        \label{fig:abl_orig}
    \end{subfigure}
    \begin{subfigure}[b]{\imgw}
        \includegraphics[width=\linewidth, clip, trim=65 0 35 0]{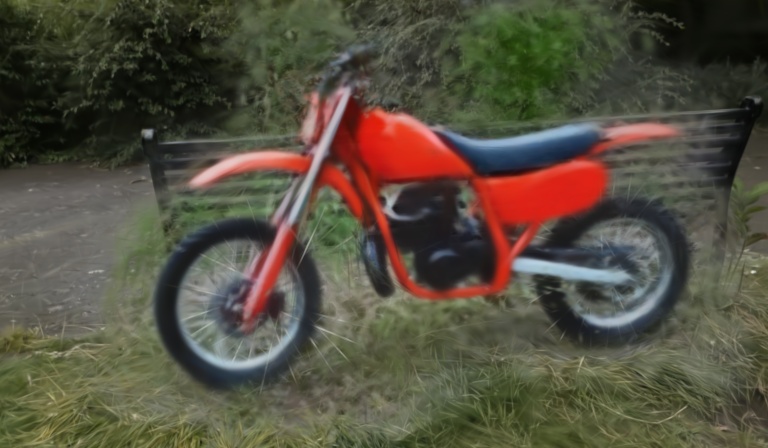}
        \subcaption{}
        \label{fig:edit_full_ft}
    \end{subfigure}
    \begin{subfigure}[b]{\imgw}
        \includegraphics[width=\linewidth, clip, trim=65 0 35 0]{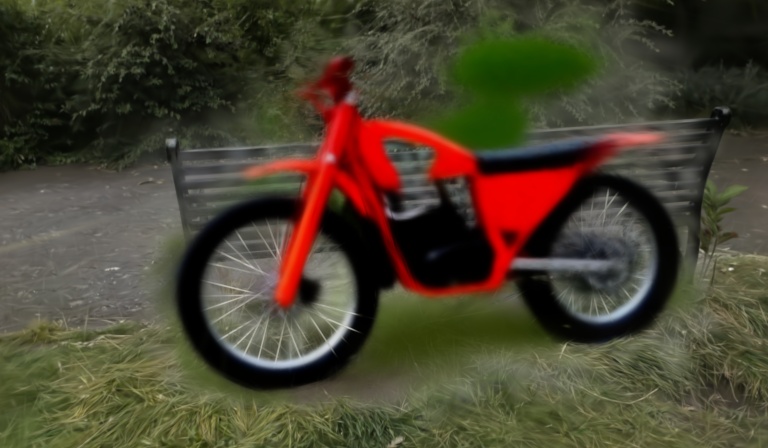}
        \subcaption{}
        \label{fig:edit_full}
    \end{subfigure}

    \vspace{0.05cm}

    \begin{subfigure}[b]{\imgw}
        \includegraphics[width=\linewidth, clip, trim=65 0 35 0]{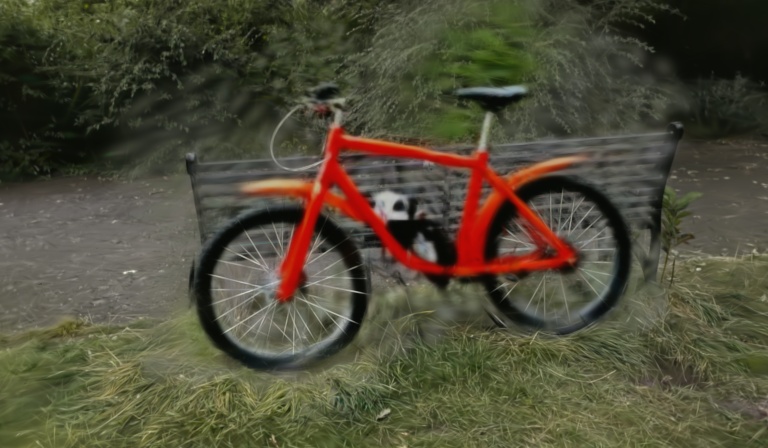}
        \subcaption{}
        \label{fig:edit_no_eqv_ft}
    \end{subfigure}
    \begin{subfigure}[b]{\imgw}
        \includegraphics[width=\linewidth, clip, trim=65 0 35 0]{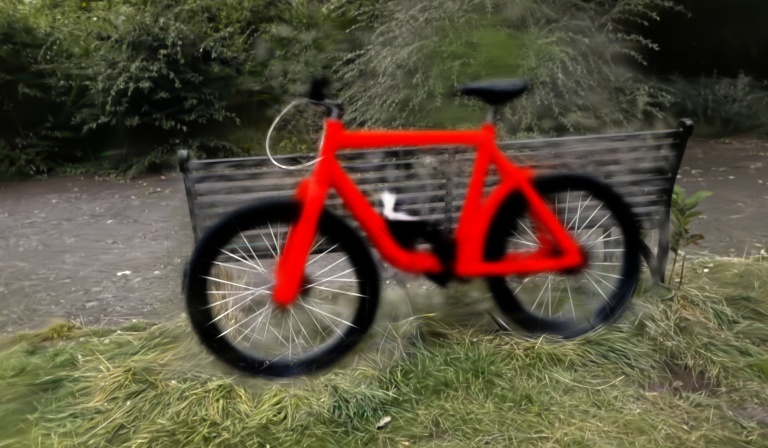}
        \subcaption{}
        \label{fig:edit_no_eqv}
    \end{subfigure}
    \begin{subfigure}[b]{\imgw}
        \includegraphics[width=\linewidth, clip, trim=50 0 50 0]{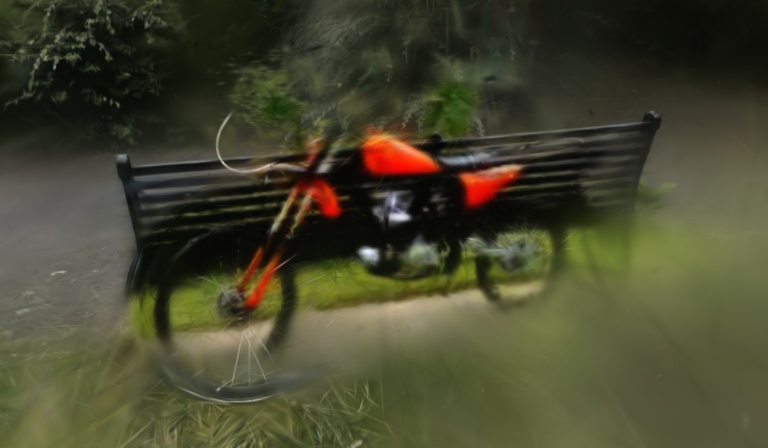}
        \subcaption{}
        \label{fig:edit_pca_ft}
    \end{subfigure}
    \begin{subfigure}[b]{\imgw}
        \includegraphics[width=\linewidth, clip, trim=50 0 50 0]{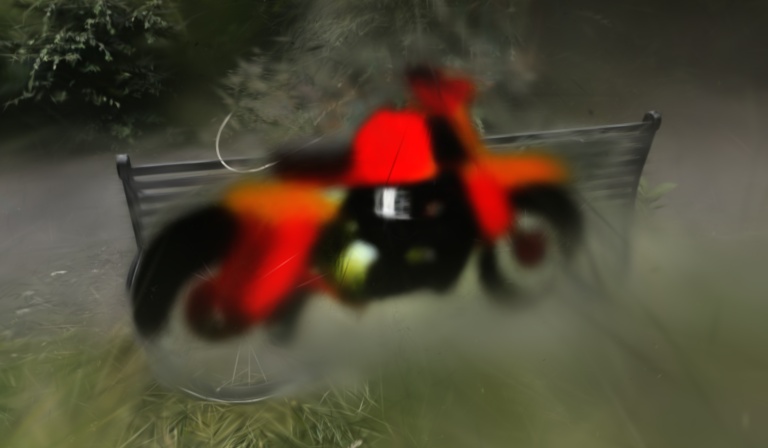}
        \subcaption{}
        \label{fig:edit_pca}
    \end{subfigure}
    \vspace{-0.1cm}
    \caption{
    Qualitative ablation, editing with the prompt: \textit{a bright orange enduro dirt bike with a low black saddle, knobby tires, a big chrome front suspension fork, disk brakes and big orange fenders}: 
    (\ref{fig:abl_orig}): original scene, 
    (\ref{fig:edit_full_ft}): $\mathcal{M}^{\text{TASE}}+\mathcal{L}_{\text{eqv}}+\text{FT}$ (Ours), 
    (\ref{fig:edit_full}): $\mathcal{M}^{\text{TASE}}+\mathcal{L}_{\text{eqv}}$, 
    (\ref{fig:edit_no_eqv_ft}): $\mathcal{M}^{\text{TASE}}+\text{FT}$, 
    (\ref{fig:edit_no_eqv}): $\mathcal{M}^{\text{TASE}}$, 
    (\ref{fig:edit_pca_ft}): $\text{PCA}+\text{FT}$, 
    (\ref{fig:edit_pca}): $\text{PCA}$.
    }
    \label{fig:ablation_qual}
    \vspace{-0.4cm}
\end{figure}

\figref{fig:ablation_qual} shows an editing example from this ablation study, more examples are shown in the supplement. When using PCA as the dimensionality reduction method (\figref{fig:edit_pca},\subref{fig:edit_pca_ft}), the edited scene often displays some of the visual elements requested in the prompt, resulting a a relatively high CLIP dir. sim. (\tabref{tab:ablation}), but does not converge to a sensible geometry. When $\mathcal{M}^{\text{TASE}}$ is trained without $\mathcal{L}_{\text{eqv}}$, the geometry occasionally converges to a largely meaningful result, as illustrated in \figref{fig:edit_no_eqv}, \subref{fig:edit_no_eqv_ft}. However, this behavior is not consistent. In many cases, the optimization fails entirely, as shown in the additional examples in the supplement. These frequent failures lead to the low average CLIP dir. sim. reported in \tabref{tab:ablation}. The Difix3D+ finetuning stage (\figref{fig:edit_full_ft},\subref{fig:edit_no_eqv_ft},\subref{fig:edit_pca_ft}) can be observed to help the model to recover finer detail, independent on the used semantic embedding.

\section{Conclusion}
\label{sec:conclusion}

In this paper, we introduced TASE, a truncation-aware semantic embedding that adapts pretrained 2D features for controllable, text-driven 3D scene editing. By explicitly structuring the embedding such that channel truncation yields increasingly abstract semantics, TASE provides direct control over edit strength and adherence to the original scene content. We further improved cross-view feature consistency using a scale- and translation-equivariance loss, enabling robust fusion in 3D, and proposed a finetuning stage for the editing diffusion model to reduce artifacts introduced by large geometric modifications. Experiments show that our pipeline supports substantial geometry changes and achieves stronger editing quality than prior methods on challenging edits involving major geometric variations. Ablation studies validate the impact of each component, confirming the benefits of the equivariance loss, the diffusion finetuning stage, and our choice of an autoencoder-based truncation-aware embedding over PCA. 

\textbf{Limitations.} Due to the diverse requirements of scene editing tasks, some applications may benefit from hyperparameter tuning, particularly for the 3DGS densification. As with other diffusion-based methods, stochastic sampling can introduce localized artifacts. Moreover, because semantics and appearance are not explicitly disentangled, color adjustments may not always match the prompt and can appear overly saturated, an aspect we view as a promising direction for future work.

\bibliographystyle{splncs04}
\bibliography{main}

\appendix
\clearpage
\setcounter{page}{1}
\maketitlesupplementary

\renewcommand{\thefigure}{S\arabic{figure}}
\renewcommand{\thetable}{S\arabic{table}}
\renewcommand{\thesection}{\Alph{section}}

In this supplementary material we present further material expanding on the implementation details and the experimental evaluation for our paper. In \secref{sec:implementation} we provide more information concerning our implementation. \secref{sec:sup_editing} contains a more in depth description of our 3D scene editing method. In \secref{sec:quant_results} we describe in detail how we arrived at the quantitative evaluation of our method. \secref{sec:segmentation} provides more detail on the segmentation process using our truncation-aware semantic embeddings. Additional qualitative results for local 3D scene editing comparing against state-of-the-art baselines are provided in \secref{sec:results} while qualitative comparisons of our ablations are given in \secref{sec:supp_ablations}.

\section{Further Implementation Details}
\label{sec:implementation}

We use gsplat~\cite{ye2025gsplat} as the backbone for 3DGS rendering and optimization. For densification and pruning we used gsplat's default strategy with the hyperparameters given in \tabref{tab:hyperparams}. Per-scene optimization relies on the Adam optimizer~\cite{kingma2014adam}, the learning rates for all parameters are also given in \tabref{tab:hyperparams}, as well as the weights of the splatting loss $\mathcal{L}_\text{splat}$. To reduce computational cost, $\mathcal{L}_s$ and $\mathcal{L}_\text{maha}$ are evaluated only every $10^\text{th}$ iteration.

\begin{table}[th]
    \centering
    \caption{Hyperparameters for 3DGS during reconstruction and editing.}
    \begin{tabular}{lcc}
        \toprule
                                          & \textbf{Reconstruction} & \textbf{Editing} \\
        \midrule
        \multicolumn{3}{l}{\textbf{Densification}}                                     \\
        \midrule
        prune\_opa                        & 0.005                   & 0.01             \\
        grow\_grad2d                      & 0.001                   & 0.0002           \\
        grow\_scale3d                     & 0.05                    & 0.05             \\
        grow\_scale2d                     & 0.001                   & 0.001            \\
        prune\_scale2d                    & 0.15                    & 0.05             \\
        refine\_every                     & 100                     & 100              \\
        \midrule
        \multicolumn{3}{l}{\textbf{Learning Rates}}                                    \\
        \midrule
        $\eta_{\boldsymbol{\mu}}$         & 0.00016                 & 0.001            \\
        $\eta_\mathbf{s}$                 & 0.005                   & 0.0005           \\
        $\eta_\alpha$                     & 0.05                    & 0.01             \\
        $\eta_\mathbf{q}$                 & 0.001                   & 0.0005           \\
        $\eta_{\mathbf{\hat{c}}_{0}}$     & 0.0025                  & 0.0025           \\
        $\eta_{\mathbf{\hat{c}}_{1...M}}$ & 0.000125                & 0.000125         \\
        $\eta_f$                          & 0.001                   & 0.0001           \\
        \midrule
        \multicolumn{3}{l}{\textbf{Loss Weights}}                                      \\
        \midrule
        $\lambda_\text{l1}$               & \multicolumn{2}{c}{0.3}                    \\
        $\lambda_\text{SSIM}$             & \multicolumn{2}{c}{0.1}                    \\
        $\lambda_f$                       & \multicolumn{2}{c}{0.3}                    \\
        $\lambda_s$                       & \multicolumn{2}{c}{50}                     \\
        $\lambda_\text{maha}$             & \multicolumn{2}{c}{0.1}                    \\
        \bottomrule
    \end{tabular}
    \label{tab:hyperparams}
\end{table}

Our ControlNet implementation is based on the diffusers library~\cite{von-platen-etal-2022-diffusers} and is trained with the Adafactor optimizer~\cite{shazeer2018adafactor}. For the semantic feature extractor, we use DINOv3 ViT-B/16~\cite{Simeoni2025} as the backbone $\mathcal{M}^\text{TASE}_\text{base}$, and train the autoencoder $\mathcal{M}^\text{TASE}_\text{enc}, \mathcal{M}^\text{TASE}_\text{dec}$ using the Adam optimizer~\cite{kingma2014adam} and $\lambda_\text{eqv} = 10^3$ as the weight for the equivariance loss $\mathcal{L}_\text{eqv}$. Learning rates for all components are reported in the main paper.

\def\imgw3{0.125\textwidth}

\begin{figure}[t]
    \centering
    \setlength{\tabcolsep}{2pt}  %
    \begin{tabular}{{>{\centering\arraybackslash}m{\imgw3}>{\centering\arraybackslash}m{\imgw3}>{\centering\arraybackslash}m{\imgw3}>{\centering\arraybackslash}m{\imgw3}>{\centering\arraybackslash}m{\imgw3}>{\centering\arraybackslash}m{\imgw3}>{\centering\arraybackslash}m{\imgw3}}}
        \toprule
        Original                                                                          & 6 iterations                                                                      & 12 iterations                                                                     & 18 iterations                                                                     & 24 iterations                                                                     & 30 iterations                                                                     & Final Result                                                                      \\
        \midrule
        \includegraphics[width=\imgw3]{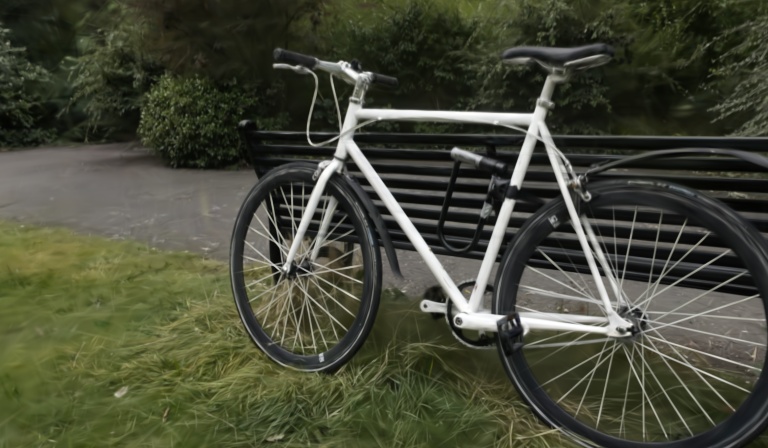}     & \includegraphics[width=\imgw3]{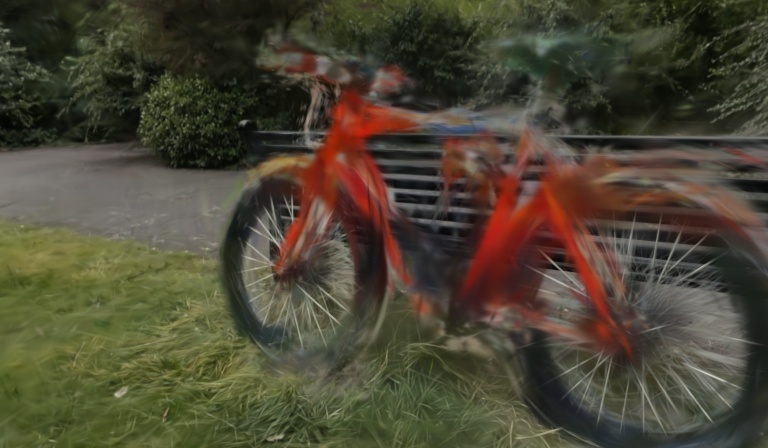}     & \includegraphics[width=\imgw3]{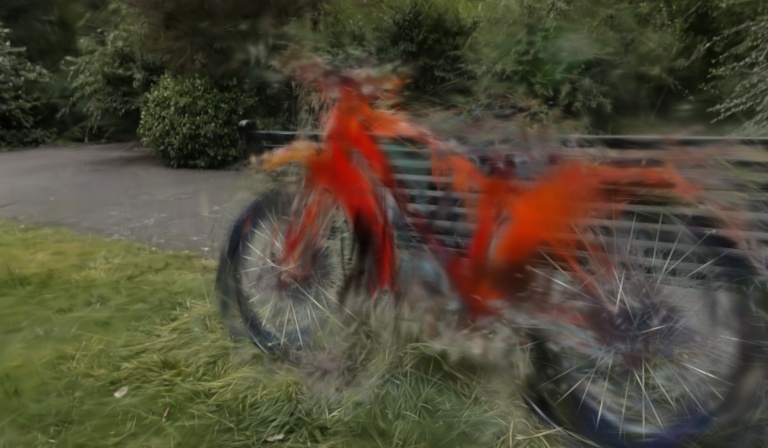}     & \includegraphics[width=\imgw3]{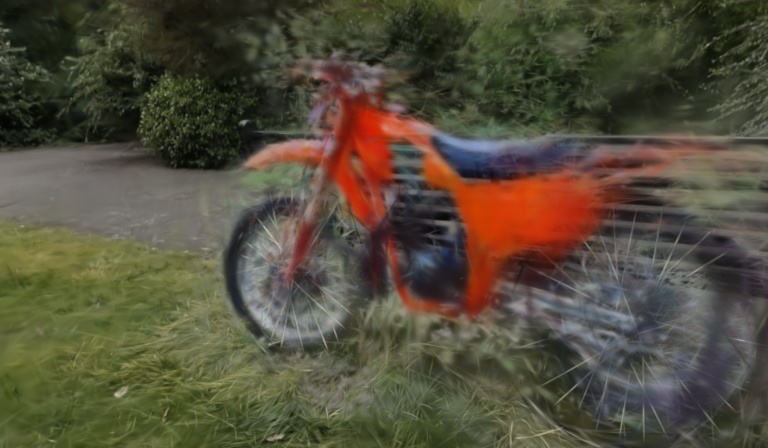}     & \includegraphics[width=\imgw3]{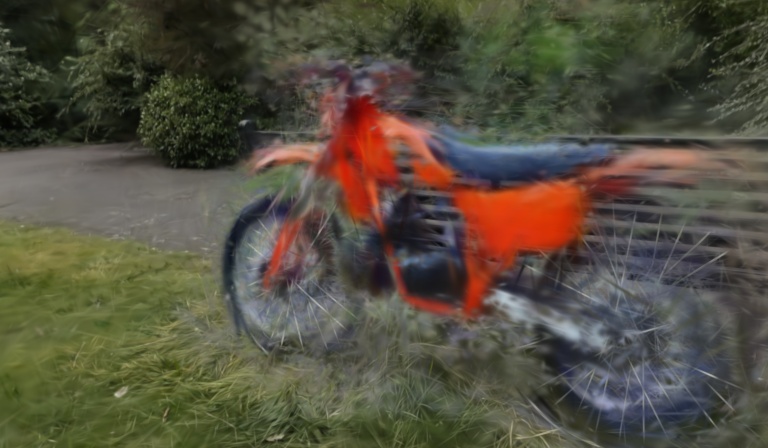}     & \includegraphics[width=\imgw3]{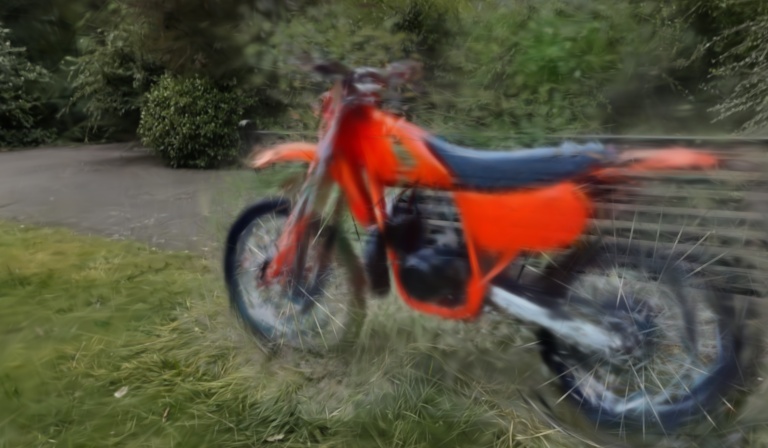}     & \includegraphics[width=\imgw3]{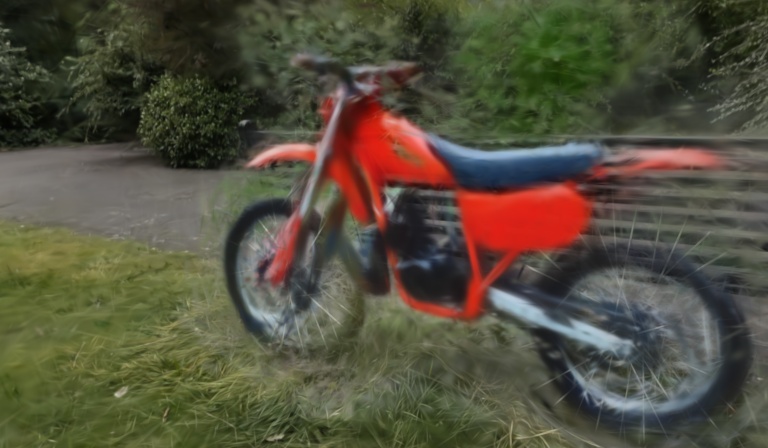}     \\
        \includegraphics[width=\imgw3]{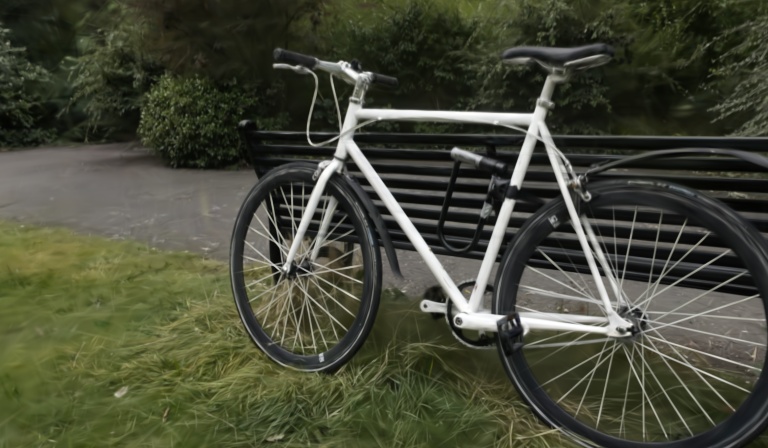} & \includegraphics[width=\imgw3]{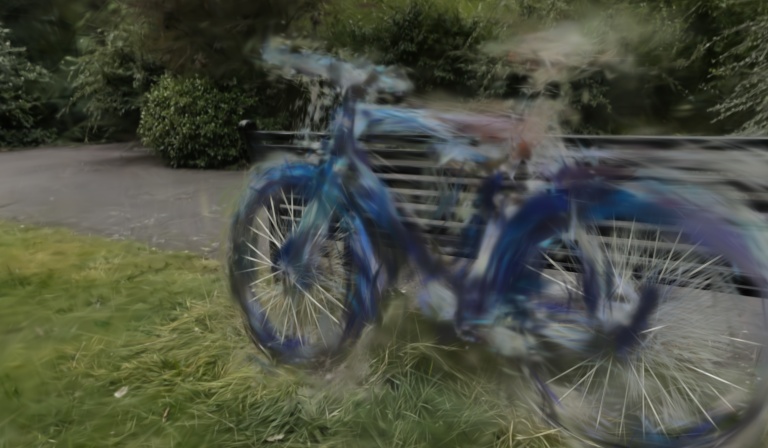} & \includegraphics[width=\imgw3]{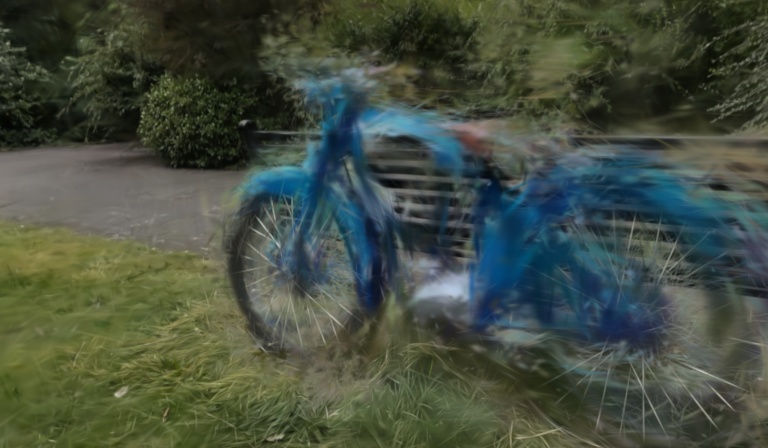} & \includegraphics[width=\imgw3]{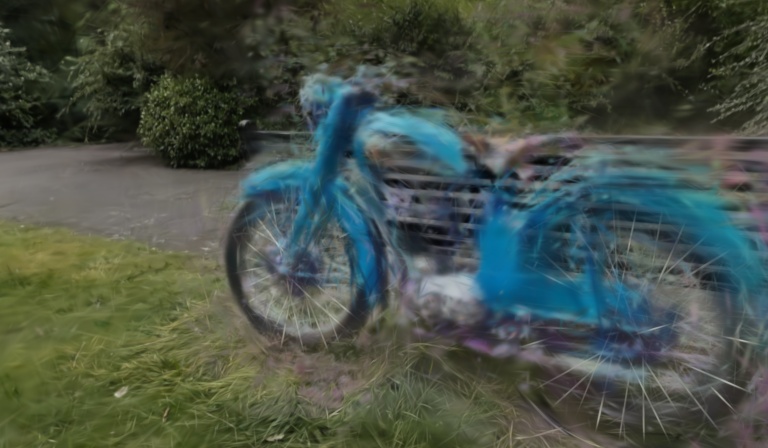} & \includegraphics[width=\imgw3]{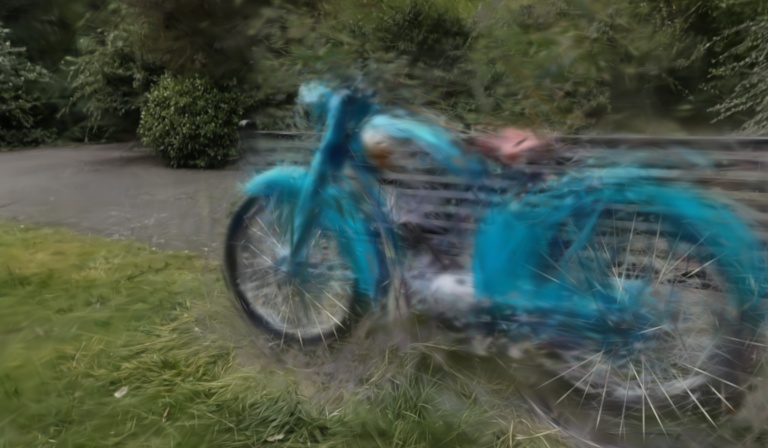} & \includegraphics[width=\imgw3]{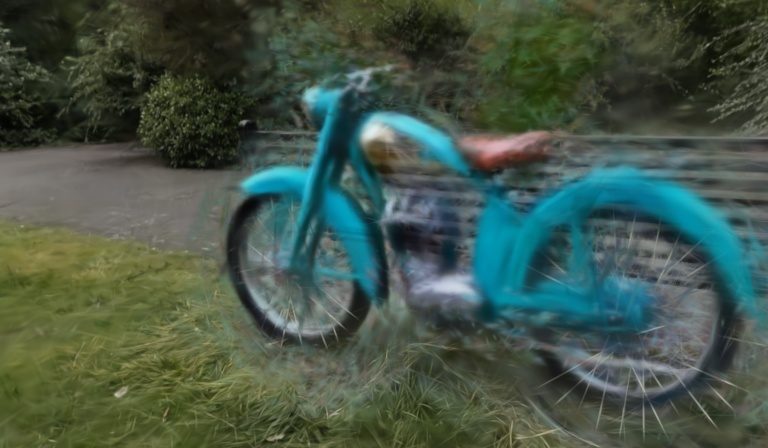} & \includegraphics[width=\imgw3]{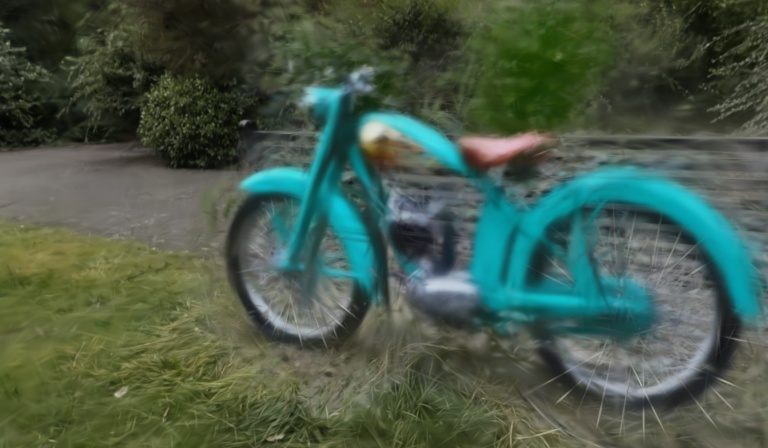} \\
        \midrule
        Original                                                                          & 5 iterations                                                                      & 10 iterations                                                                     & 15 iterations                                                                     & 20 iterations                                                                     & 25 iterations                                                                     & Final Result                                                                      \\
        \midrule
        \includegraphics[width=\imgw3]{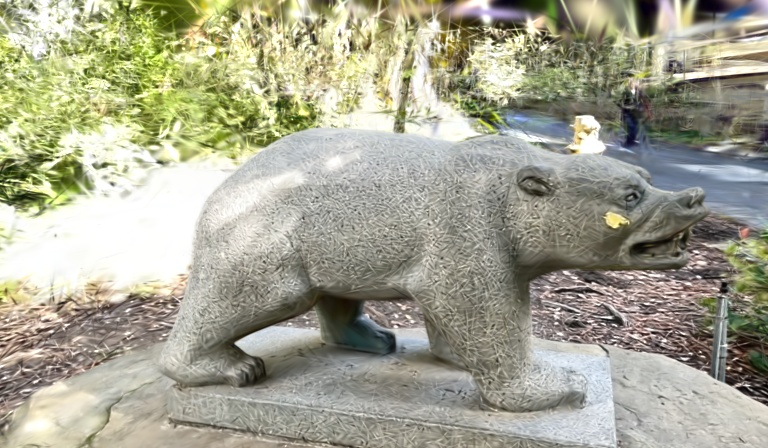}  & \includegraphics[width=\imgw3]{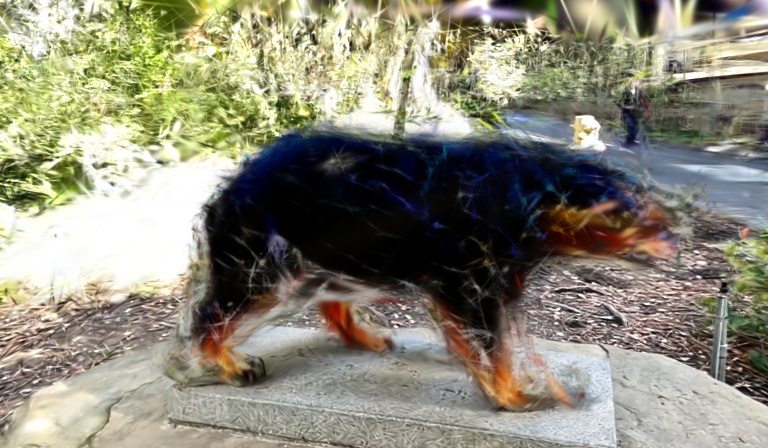}  & \includegraphics[width=\imgw3]{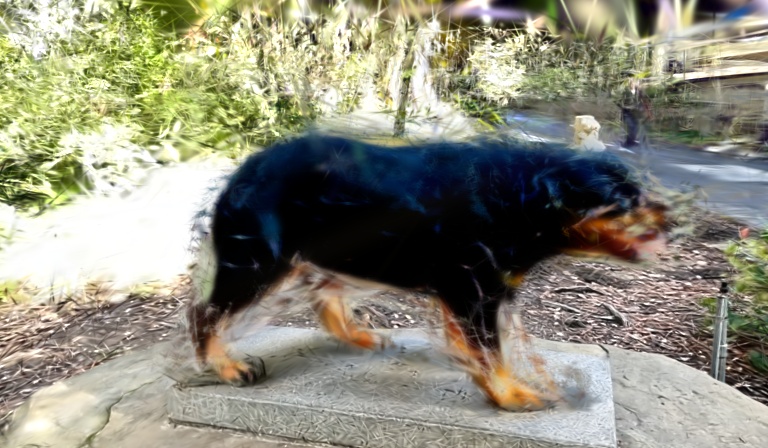}  & \includegraphics[width=\imgw3]{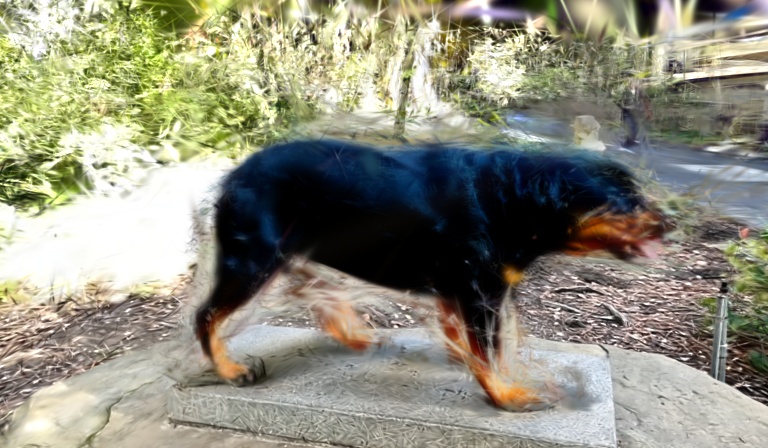}  & \includegraphics[width=\imgw3]{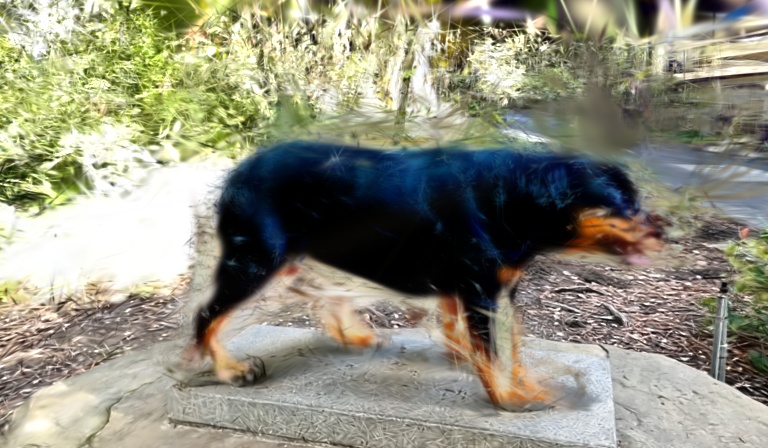}  & \includegraphics[width=\imgw3]{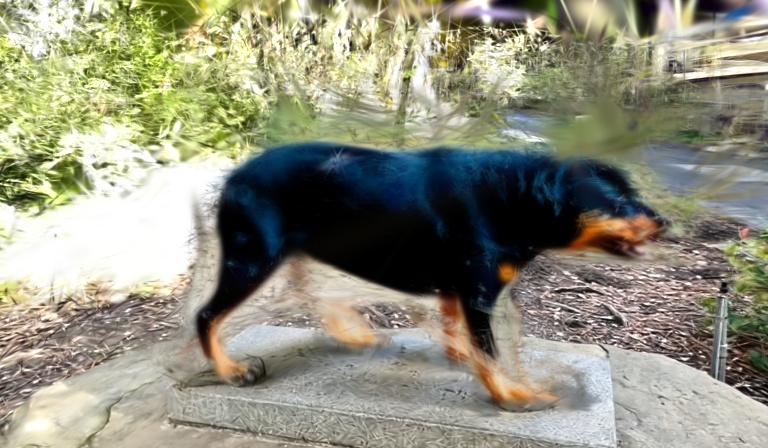}  & \includegraphics[width=\imgw3]{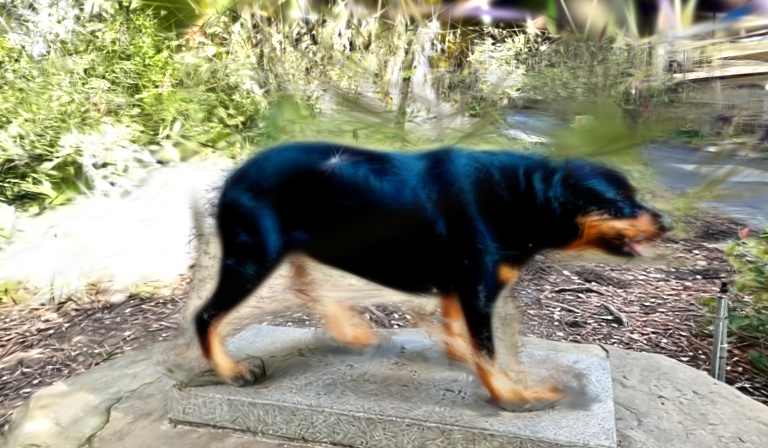}  \\
        \includegraphics[width=\imgw3]{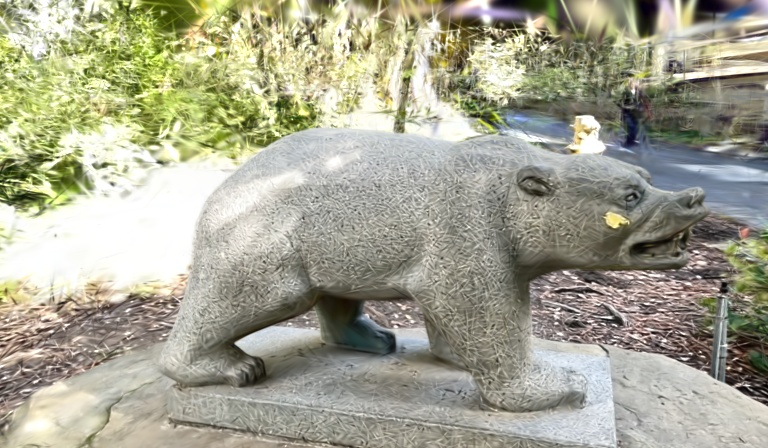}      & \includegraphics[width=\imgw3]{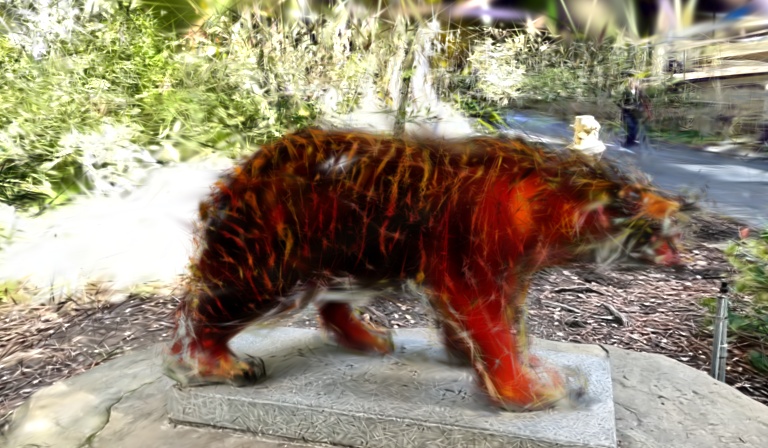}      & \includegraphics[width=\imgw3]{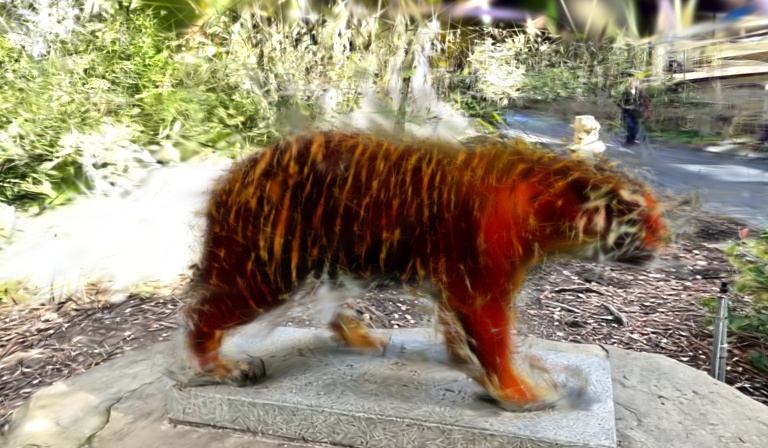}      & \includegraphics[width=\imgw3]{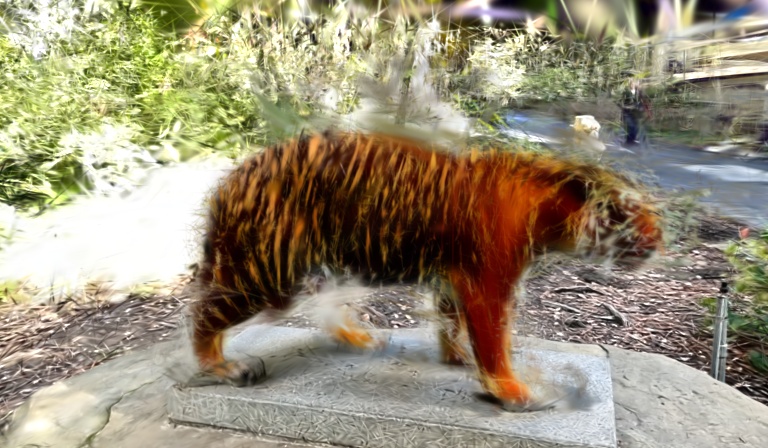}      & \includegraphics[width=\imgw3]{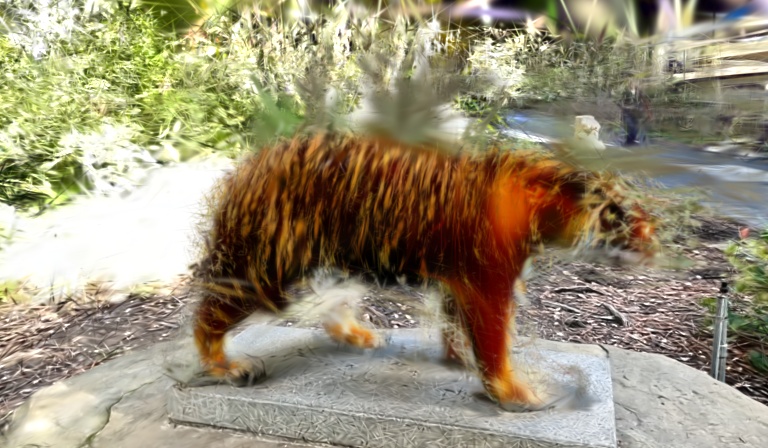}      & \includegraphics[width=\imgw3]{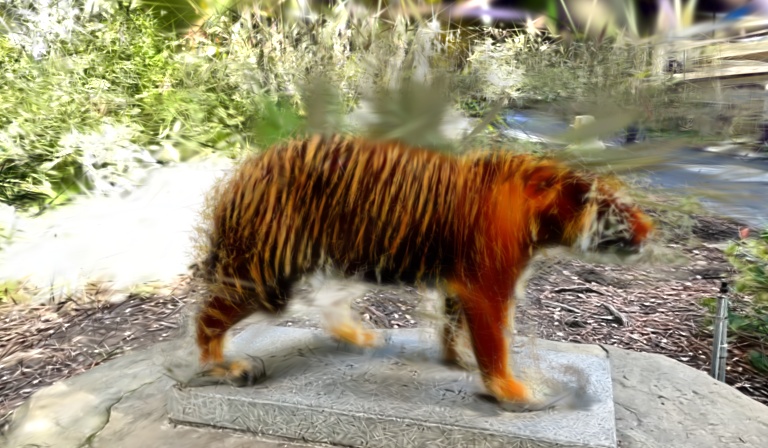}      & \includegraphics[width=\imgw3]{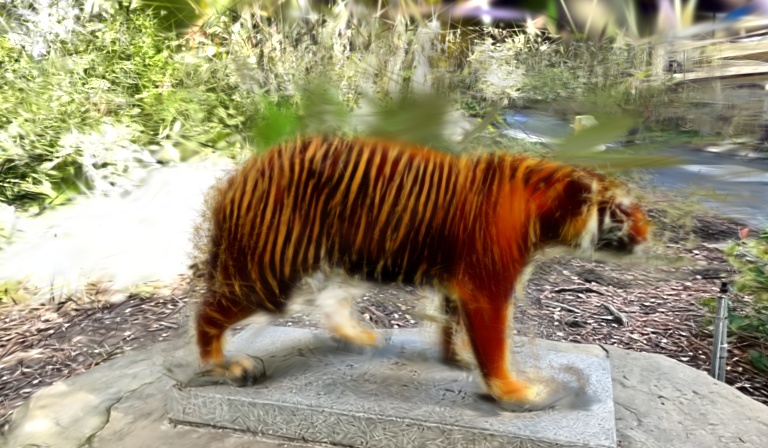}      \\
        \bottomrule
    \end{tabular}
    \caption{Changing of the scene during the editing process using our method. The Final Result is after 36 iterations for the \textit{bicylce} edits and after 30 iterations for the \textit{bear} edits.}
    \label{fig:transformations}
\end{figure}

\section{Further Editing Details}
\label{sec:sup_editing}

We separated the editing sequence into an \textit{editing} phase, in which the method freely adjusts the scene content to align with the text prompt and a \textit{consolidation} phase, in which the method focuses on consolidating the changes made in the editing phase for a multi-view consistent final scene with few artifacts. The number of iterations for editing $T_e$ and consolidation $T_c$ can be adjusted independently. The learning rates for the geometry
$\eta_{\boldsymbol{\mu}}, \eta_{\mathbf{s}}, \eta_{\mathbf{q}}$ are scaled by a factor $s_\eta(t)$, which increases from $s_\text{min}$ to $s_\text{max}$ during editing and decreases back to $s_\text{min}$ during consolidation:
\begin{equation}
    s_\eta(t)
    = s_\text{min}
    + (s_\text{max} - s_\text{min})
    \sin\!\left(\frac{\pi}{2}\,\tau(t)\right),
\end{equation}
with
\begin{equation}
    \tau(t) =
    \begin{cases}
        \dfrac{t}{T_e}, & 0 \le t < T_e,         \\[8pt]
        1 + \dfrac{t - T_e}{T_c},
                        & T_e \le t < T_e + T_c.
    \end{cases}
\end{equation}
where $t$ is the current iteration. The low learning rate at the beginning forces the method to first adjust the appearance. Once the appearance is aligned with the prompt, the higher learning rate allows for larger geometry changes. During consolidation, the decreasing learning rate helps the edit to converge to a multi-view consistent end result. \figref{fig:transformations} shows the way that the 3D scene changes during the editing process. The learning rates for the semantic embeddings and the SH coefficients were kept constant during both phases. The exact hyperparameters used for each edit can be found in \tabref{tab:edit_hyperparams} with $n_\text{images}$ denoting the number of images generated for each iteration of the editing loop and $n_\text{steps}$ being the number of 3DGS optimization steps taken for each iteration. The regularization weights for opacity and color changes as well as their decay rates were adjusted per scene and are given in \tabref{tab:edit_hyperparams_scene}.

\begin{table}[t]
    \centering
    \caption{Hyperparameters for different types of edits.}
    \begin{tabular}{lrrrr}
        \toprule
                          &        & \multicolumn{3}{c}{Local}                  \\
        \cmidrule(lr){3-5}
                          & Global & Small                     & Medium & Large \\
        \midrule
        $T_e$             & 2      & 4                         & 8      & 12    \\
        $T_c$             & 6      & 16                        & 16     & 24    \\
        $n_\text{images}$ & 16     & 6                         & 6      & 6     \\
        $n_\text{steps}$  & 500    & 200                       & 300    & 300   \\
        $s_\text{min}$    & 0.0    & 0.2                       & 0.5    & 0.5   \\
        $s_\text{max}$    & 0.1    & 0.5                       & 1.0    & 3.0   \\
        \bottomrule
    \end{tabular}
    \label{tab:edit_hyperparams}
\end{table}

\begin{table}[t]
    \centering
    \caption{Hyperparameters for different scenes.}
    \begin{tabular}{lrrrr}
        \toprule
                               & Bicycle & Bear  & Face \\
        \midrule
        Opacity Regularization & 0.05    & 0.2   & 0.0  \\
        Opacity Decay          & 0.025   & 0.1   & 0.0  \\
        Color Regularization   & 0.005   & 0.005 & 0.02 \\
        Color Decay            & 0.005   & 0.005 & 0.02 \\
        \bottomrule
    \end{tabular}
    \label{tab:edit_hyperparams_scene}
\end{table}

\section{Quantitative Evaluation}
\label{sec:quant_results}

\begin{figure}
    \centering
    \includegraphics[width=0.47\textwidth, trim = 550 40 550 40, clip]{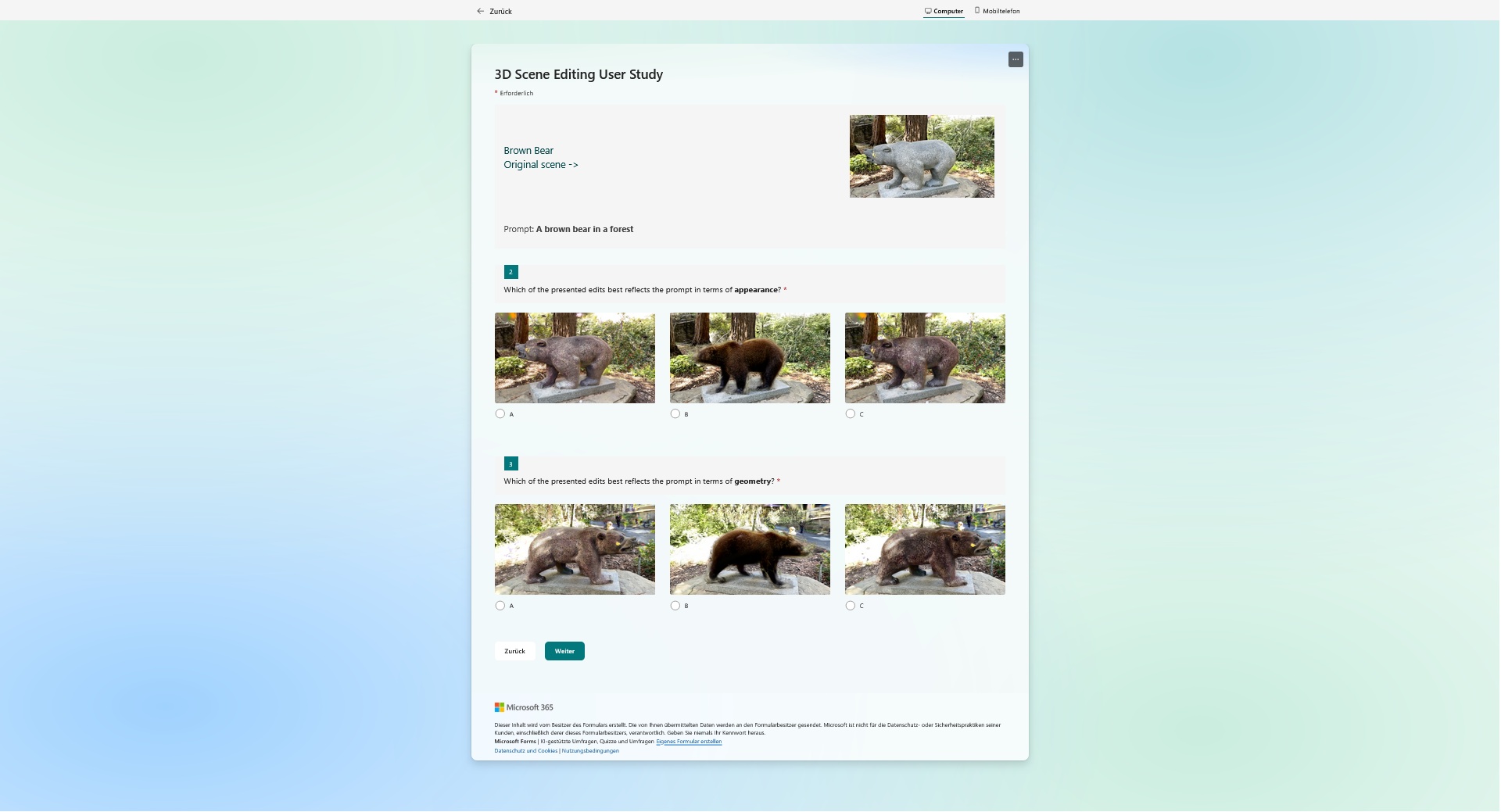}
    \caption{Screenshot from the user study survey.}
    \label{fig:survey}
    \vspace{-0.5cm}
\end{figure}

We used an online survey to conduct a user study to quantitatively evaluate the editing results. We showed 65 participants the original and edited renderings from the two different views shown in in \figref{fig:bike_editing} \figref{fig:bear_editing} and \figref{fig:face_editing} for each of the three datasets and all six edits per dataset. For each edit, we first displayed an image of the original scene and the text prompt describing the desired changes. We then showed the results of our method, Direct~Gaussian~Editing~(DGE)~\cite{Chen2025d} and GaussianEditor~(GE)~\cite{YCZCCZFWXYYWZCLYHLGL2023} side-by-side, in random order. The participants were asked to select which of the three results they thought matched the prompt the best in terms of geometry and appearance separately. We described the geometry as ``the overall shape and structure'' and the appearance as ``the colors and textures'' of the scene. A screenshot from the survey is depicted in \figref{fig:survey}. The per-dataset and per-experiment results are shown in \tabref{tab:quantitave_results}.

We computed the CLIP dir. sim.~\cite{gal2022stylegan} on a set of renders from the original and edited scenes to quantitatively evaluate how well the edits aligned with the text prompts. For \textit{bicycle} and \textit{bear} we used 24 renders in a grid around the object, for \textit{face} we used 20 renders from the front side of the face. To compute the text direction we used ``a photo of \textit{\{PROMPT\}} next to a bench in a park'', ``a photo of \textit{\{PROMPT\}} standing in a forest'', and ``a portrait of \textit{\{PROMPT\}} in a gray sweater in front of a concrete wall'' as the template for the respective text descriptions. We inserted the prompts detailed in \figref{fig:bike_editing}, \figref{fig:bear_editing}, and \figref{fig:face_editing} in place of \textit{\{PROMPT\}}. For the description of the original scenes we inserted ``a white road bicycle'', ``a gray stone statue of a bear'' and ``a blonde man'' instead. The results are included in \tabref{tab:quantitave_results}.

\def\imgw5{0.23\textwidth}
\newcommand{\dotx}{0.725}   %
\newcommand{\doty}{0.45}   %
\newcommand{\dotsize}{1pt} %

\begin{figure}[t]
    \centering
    \begin{subfigure}[b]{\imgw5}
        \centering
        \imgwithdot{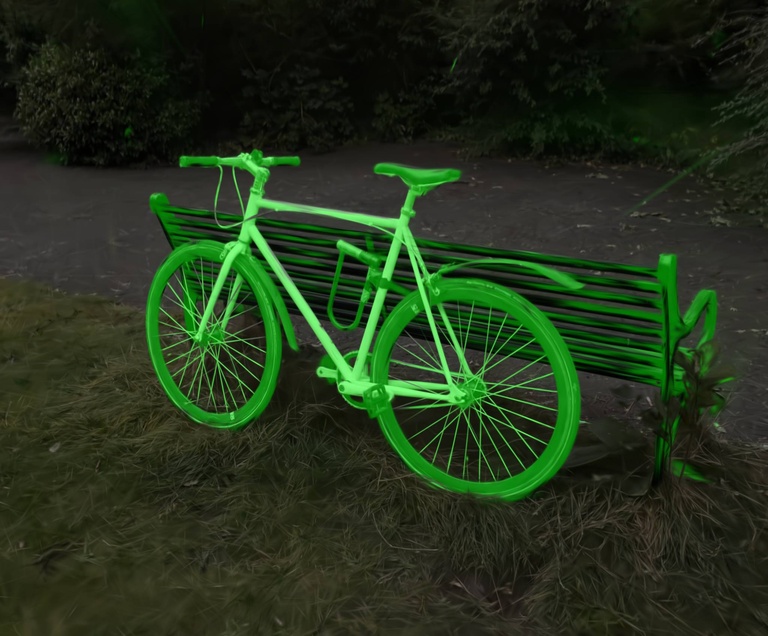}
        \caption{2 Dims}
        \label{fig:seg_trade_2}
    \end{subfigure}
    \begin{subfigure}[b]{\imgw5}
        \centering
        \imgwithdot{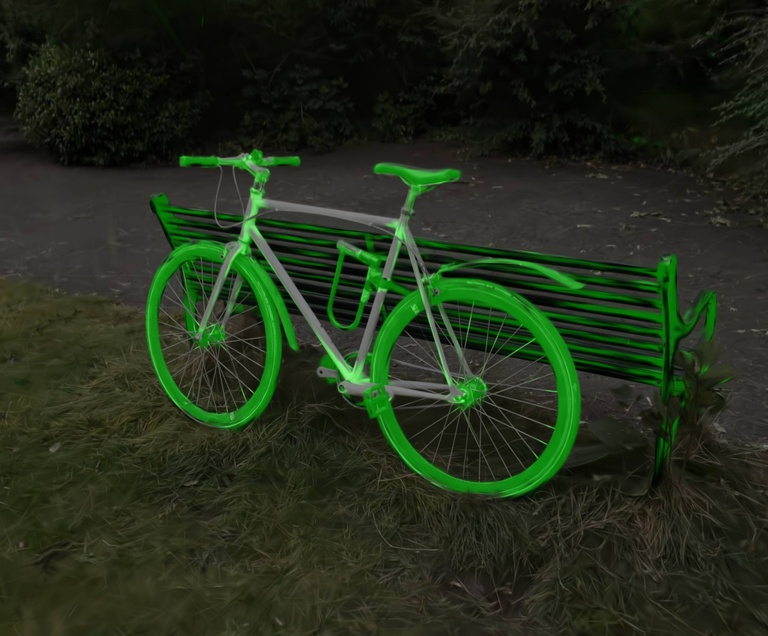}
        \caption{4 Dims}
        \label{fig:seg_trade_4}
    \end{subfigure}
    \begin{subfigure}[b]{\imgw5}
        \centering
        \imgwithdot{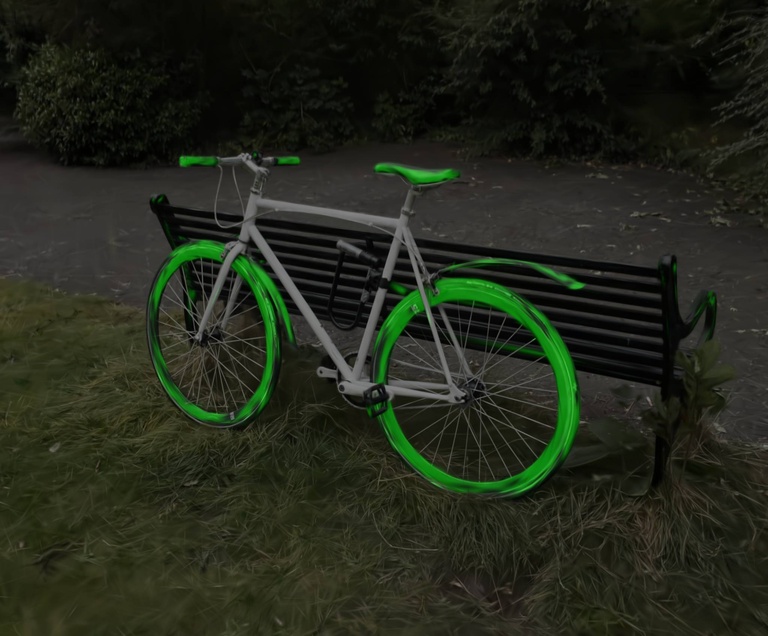}
        \caption{16 Dims}
        \label{fig:seg_trade_16}
    \end{subfigure}
    \begin{subfigure}[b]{\imgw5}
        \centering
        \imgwithdot{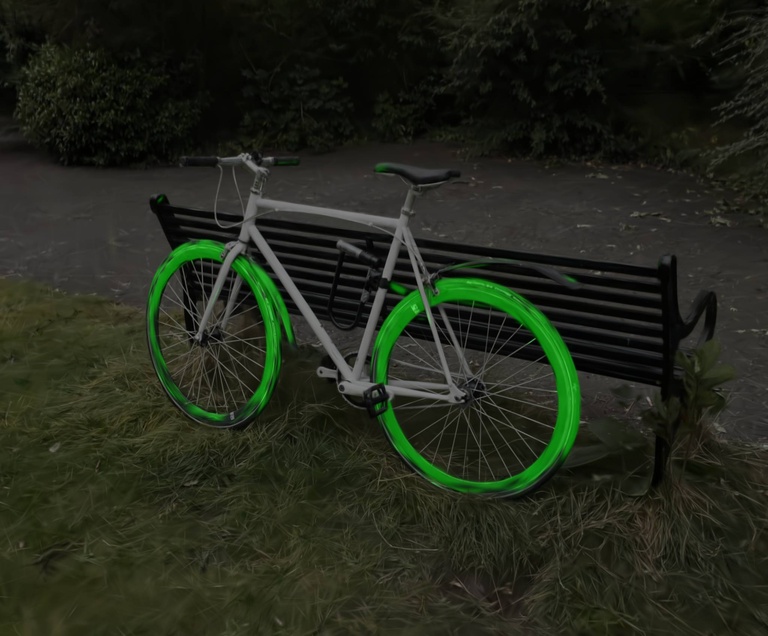}
        \caption{64 Dims}
        \label{fig:seg_trade_64}
    \end{subfigure}
    \caption{Impact of different truncation levels during segmentation for a single anchor on the back wheel (red dot).}
    \label{fig:seg_tradoff}
\end{figure}

\section{Segmentation}
\label{sec:segmentation}

\def\imgw4{0.25\textwidth}

\begin{figure}[t]
    \centering
    \begin{subfigure}[b]{\imgw4}
        \includegraphics[width=\linewidth]{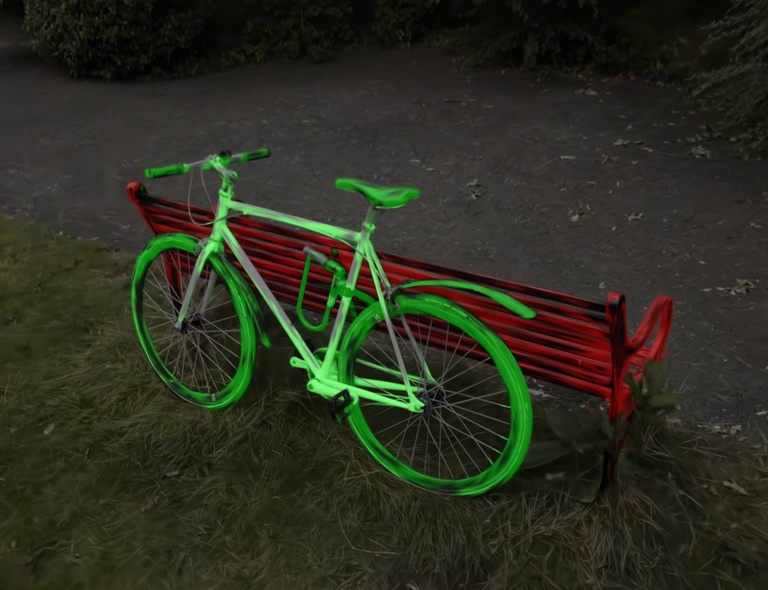}
        \caption{Anchors only}
        \label{fig:proc1}
    \end{subfigure}
    \begin{subfigure}[b]{\imgw4}
        \includegraphics[width=\linewidth]{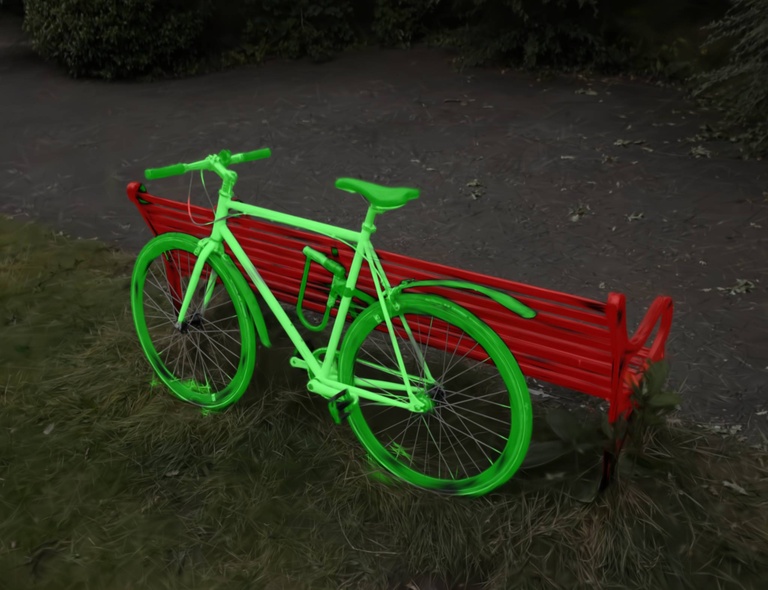}
        \caption{Label Propagation}
        \label{fig:proc2}
    \end{subfigure}
    \begin{subfigure}[b]{\imgw4}
        \includegraphics[width=\linewidth]{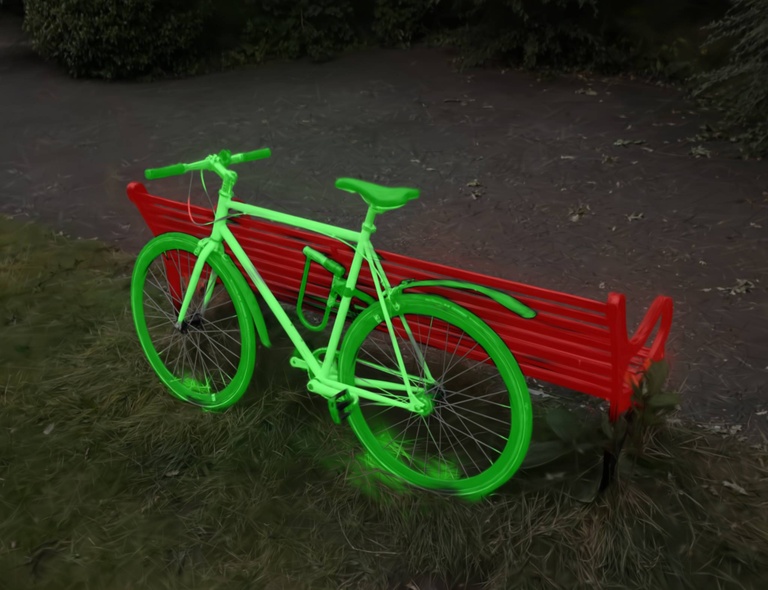}
        \caption{Outlier Removal}
        \label{fig:proc3}
    \end{subfigure}
    \caption{Selected gaussians in green, negative selection in red at different stages of the segmentation.}
    \label{fig:seg_proc}
\end{figure}

In \figref{fig:seg_tradoff} the impact of the truncation level during the segmentation is shown. For illustrative purposes, a single anchor is placed on the back wheel. When only 2 channels of our truncation-aware semantic embedding are used, most of the foreground objects are selected, including all parts of the bike and the bench. For 4 and 16 channels fewer parts of the bench and the bike are included in the selection. At 64 channels, only Gaussians that belong to the wheels of the bike are included. \figref{fig:seg_proc} shows the effects of the different stages during our segmentation method. First the truncation level and the threshold $\tau$ are set to a value such that no false positives are selected. Then anchors are added so that Gaussians in all parts of the scene that should be included are selected. This results in the selection shown in \figref{fig:proc1}. The label propagation leads to a much more complete selection which is shown in \figref{fig:proc2}. Finally, the remaining false positives on the bench and unclassified Gaussians on the wheels are removed through the majority vote which results in the final segmentation shown in \figref{fig:proc3}.

\def\imgw4{0.23\textwidth}

\begin{figure}[t]
    \centering
    \begin{subfigure}[b]{\imgw4}
        \includegraphics[width=\linewidth, trim=20 24 0 20, clip]{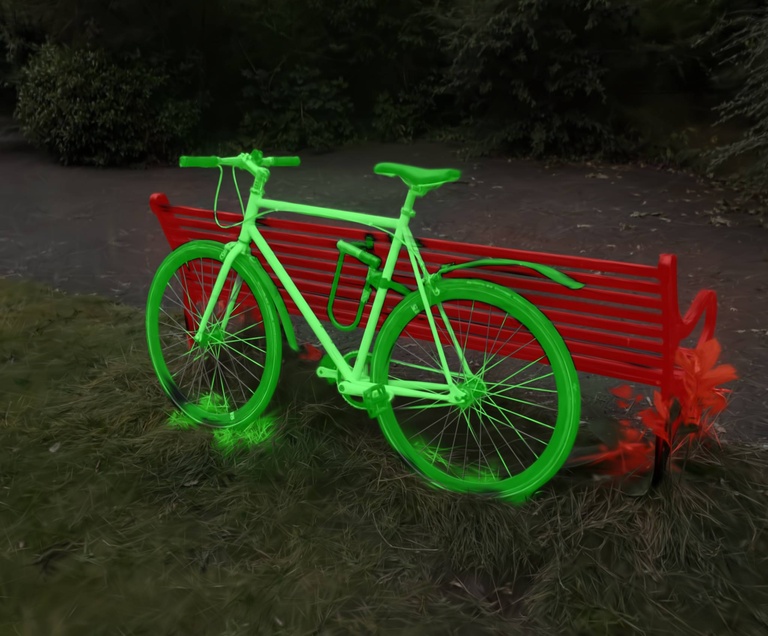}
        \caption{Bicycle}
    \end{subfigure}
    \begin{subfigure}[b]{\imgw4}
        \includegraphics[width=\linewidth]{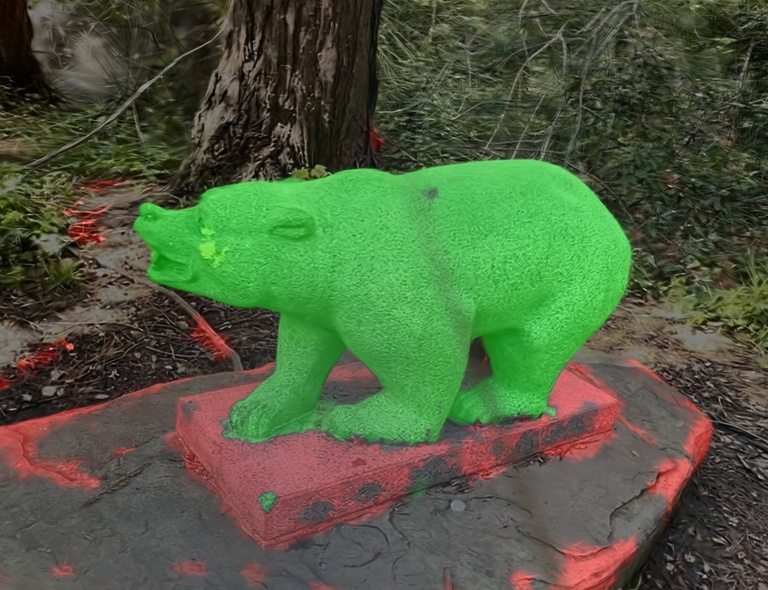}
        \caption{Bear}
    \end{subfigure}
    \begin{subfigure}[b]{\imgw4}
        \includegraphics[width=\linewidth]{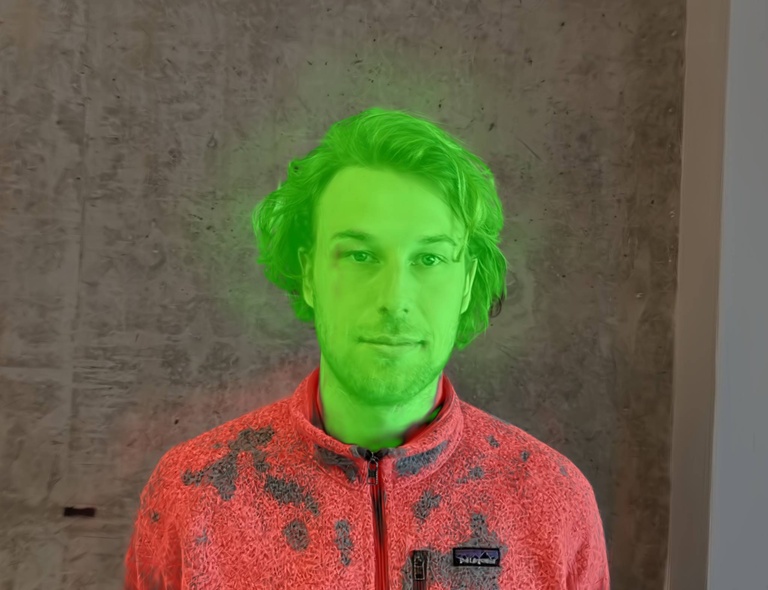}
        \caption{Face}
    \end{subfigure}
    \caption{Segmentation used for local edits. Selected gaussians in green, negative selection in red.}
    \label{fig:seg_edits}
\end{figure}

\begin{figure}[t]
    \centering
    \begin{subfigure}[b]{\imgw4}
        \includegraphics[width=\linewidth, trim=150 68 105 0, clip]{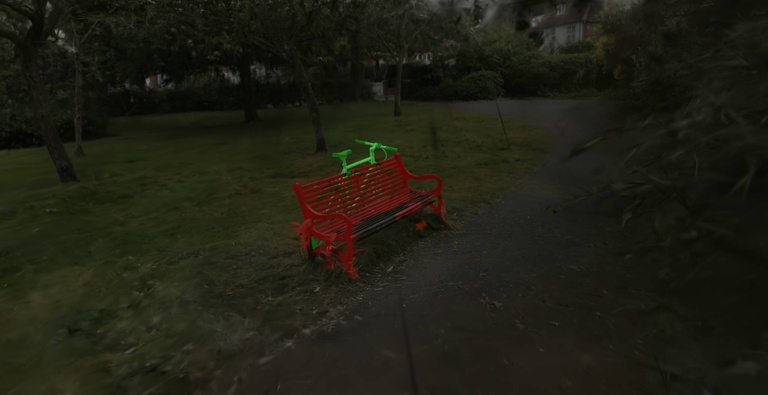}
        \caption{Ours}
        \label{fig:seg_comp_ours}
    \end{subfigure}
    \begin{subfigure}[b]{\imgw4}
        \includegraphics[width=\linewidth, trim=75 0 150 0, clip]{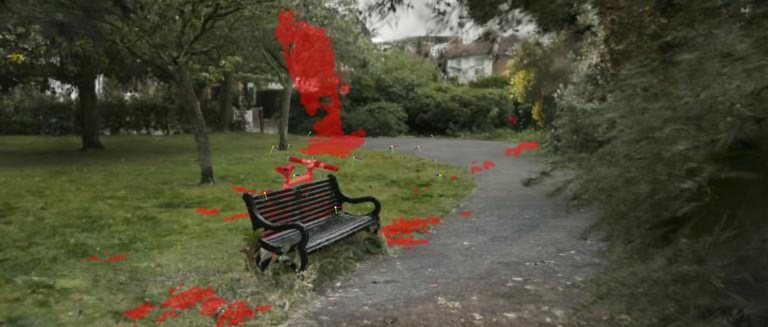}
        \caption{$\tau_\text{conf}=0.9$}
        \label{fig:seg_comp_sam1}
    \end{subfigure}
    \begin{subfigure}[b]{\imgw4}
        \includegraphics[width=\linewidth, trim=75 0 150 0, clip]{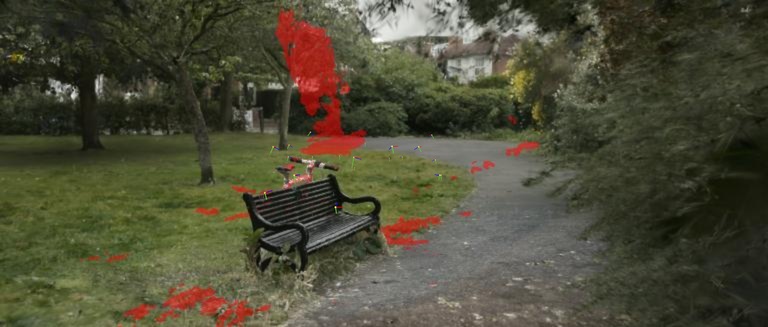}
        \caption{$\tau_\text{conf}=0.99$}
        \label{fig:seg_comp_sam2}
    \end{subfigure}
    \caption{Comparison of segmentation quality to LangSAM. Selected in green for Ours in \figref{fig:seg_comp_ours}, selected in red for LangSAM in \figref{fig:seg_comp_sam1} and \figref{fig:seg_comp_sam2}.}
    \label{fig:seg_comp}
\end{figure}

The exact segmentation masks that were used for the local editing experiments are depicted in \figref{fig:seg_edits}. \figref{fig:seg_comp} illustrates the benefits of using our segmentation method over the approach used both by Direct Gaussian Editing (DGE) and \cite{Chen2025d} GaussianEditor \cite{YCZCCZFWXYYWZCLYHLGL2023}, which employ Language Segment-Anything (LangSAM)~\cite{medeiros-langSAM} to generate segmentation masks for a number of views and project the labels onto the Gaussians. GaussianEditor additionally allows to scale the required confidence for selected Gaussians. \figref{fig:seg_comp_sam1} shows that using LangSAM leads to many outliers on the grass and the trees in the background. These outliers persist even if the confidence threshold $\tau_\text{conf} = 0.99$ is set so high, that not all parts of the bicycle are included in the selection, which is shown in \figref{fig:seg_comp_sam2}. The segmentation from our method shown in \figref{fig:seg_comp_ours} includes no outliers in the background while still capturing the entirety of the target object.

\begin{sidewaystable}[t]
    \caption{Per-experiment user study results and CLIP dir. sim..}
    \label{tab:quantitave_results}
    \centering
    \begin{tabular}{llcccccccccc}
        \toprule
                &                    & \multicolumn{3}{c}{Ours} & \multicolumn{3}{c}{DGE}        & \multicolumn{3}{c}{GaussianEditor}                                                                                                                                           \\
        \cmidrule(lr){3-5} \cmidrule(lr){6-8} \cmidrule(lr){9-11}
                &                    &                          & \multicolumn{2}{c}{User Study} &                                    & \multicolumn{2}{c}{User Study} &                 & \multicolumn{2}{c}{User Study}                                                       \\
        \cmidrule(lr){4-5} \cmidrule(lr){7-8} \cmidrule(lr){10-11}
        Dataset & Experiment         & \shortstack{CLIP                                                                                                                                                                                                                         \\dir. sim.}~\shortstack{\vspace{0.35em}$\uparrow$} & \shortstack{\vspace{0.35em}Geom. $\uparrow$} & \shortstack{\vspace{0.35em}Appear. $\uparrow$} & \shortstack{CLIP\\dir. sim.}~\shortstack{\vspace{0.35em}$\uparrow$} & \shortstack{\vspace{0.35em}Geom. $\uparrow$} & \shortstack{\vspace{0.35em}Appear. $\uparrow$} & \shortstack{CLIP\\dir. sim.}~\shortstack{\vspace{0.35em}$\uparrow$} & \shortstack{\vspace{0.35em}Geom. $\uparrow$} & \shortstack{\vspace{0.35em}Appear. $\uparrow$} \\
        \midrule
        bear    & brown bear         & \textbf{0.0783}          & 16.9\%                         & 47.7\%                             & 0.0436                         & 36.9\%          & \textbf{50.8\%}                & 0.0368          & \textbf{46.2\%} & 1.5\%           \\
        bear    & golden retriever   & \textbf{0.1399}          & \textbf{100.0\%}               & \textbf{100.0\%}                   & 0.0235                         & 0.0\%           & 0.0\%                          & 0.0173          & 0.0\%           & 0.0\%           \\
        bear    & leopard            & \textbf{0.0942}          & \textbf{100.0\%}               & \textbf{96.9\%}                    & 0.0131                         & 0.0\%           & 3.1\%                          & 0.0093          & 0.0\%           & 0.0\%           \\
        bear    & polar bear         & \textbf{0.0381}          & \textbf{75.4\%}                & \textbf{43.1\%}                    & 0.0325                         & 13.8\%          & 30.8\%                         & 0.0326          & 10.8\%          & 26.2\%          \\
        bear    & rottweiler         & \textbf{0.1585}          & \textbf{98.5\%}                & \textbf{100.0\%}                   & 0.0155                         & 1.5\%           & 0.0\%                          & 0.0124          & 0.0\%           & 0.0\%           \\
        bear    & tiger              & \textbf{0.1672}          & \textbf{100.0\%}               & \textbf{100.0\%}                   & 0.0064                         & 0.0\%           & 0.0\%                          & 0.0074          & 0.0\%           & 0.0\%           \\
        bicycle & blue bike          & 0.0303                   & 33.8\%                         & \textbf{92.3\%}                    & 0.0050                         & 18.5\%          & 0.0\%                          & \textbf{0.0449} & \textbf{47.7\%} & 7.7\%          \\
        bicycle & enduro             & \textbf{0.1234}          & \textbf{84.6\%}                & \textbf{84.6\%}                    & 0.1151                         & 4.6\%           & 12.3\%                         & 0.0721          & 10.8\%          & 3.1\%           \\
        bicycle & motorcycle race    & \textbf{0.1058}          & \textbf{84.6\%}                & \textbf{92.3\%}                    & 0.0079                         & 6.2\%           & 3.1\%                          & 0.0076          & 9.2\%           & 4.6\%           \\
        bicycle & motorcycle classic & \textbf{0.1114}          & \textbf{92.3\%}                & \textbf{96.9\%}                    & 0.0079                         & 1.5\%           & 0.0\%                          & 0.0397          & 6.2\%           & 3.1\%           \\
        bicycle & mountain bike      & 0.0579                   & \textbf{75.4\%}                & \textbf{86.2\%}                    & \textbf{0.0692}                & 12.3\%          & 12.3\%                         & 0.0314          & 12.3\%          & 1.5\%           \\
        bicycle & scooter            & \textbf{0.1398}          & \textbf{93.8\%}                & \textbf{100.0\%}                   & 0.0134                         & 3.1\%           & 0.0\%                          & 0.0171          & 3.1\%           & 0.0\%           \\
        face    & clown              & 0.1196                   & \textbf{63.1\%}                & \textbf{87.7\%}                    & 0.1507                         & 7.7\%           & 1.5\%                          & \textbf{0.1762} & 29.2\%          & 10.8\%          \\
        face    & einstein           & 0.1083                   & \textbf{49.2\%}                & \textbf{53.8\%}                    & \textbf{0.1394}                & 24.6\%          & 29.2\%                         & 0.1379          & 26.2\%          & 16.9\%          \\
        face    & elf                & \textbf{0.0896}          & \textbf{64.6\%}                & \textbf{78.5\%}                    & 0.0784                         & 32.3\%          & 18.5\%                         & 0.0320          & 3.1\%           & 3.1\%           \\
        face    & old man            & \textbf{0.1027}          & 15.4\%                         & 27.7\%                             & 0.0889                         & 40.0\%          & 27.7\%                         & 0.0889          & \textbf{44.6\%} & \textbf{44.6\%} \\
        face    & orc                & 0.1056                   & 3.1\%                          & 12.3\%                             & \textbf{0.1248}                & \textbf{87.7\%} & \textbf{87.7\%}                & 0.0548          & 9.2\%           & 0.0\%           \\
        face    & vampire            & 0.0534                   & 6.2\%                          & 4.6\%                              & \textbf{0.1261}                & \textbf{52.3\%} & \textbf{60.0\%}                & 0.0899          & 41.5\%          & 35.4\%          \\
        \midrule
        bear    & average            & \textbf{0.1127}          & \textbf{81.8\%}                & \textbf{81.3\%}                    & 0.0224                         & 8.7\%           & 14.1\%                         & 0.0193          & 9.5\%           & 4.6\%           \\
        bicycle & average            & \textbf{0.0948}          & \textbf{77.4\%}                & \textbf{92.1\%}                    & 0.0364                         & 7.7\%           & 4.6\%                          & 0.0354          & 14.9\%          & 3.3\%           \\
        face    & average            & 0.0965                   & 33.6\%                         & \textbf{44.1\%}                    & \textbf{0.1181}                & \textbf{40.8\%} & 37.4\%                         & 0.0966          & 25.6\%          & 18.5\%          \\
        \midrule
        all     & average            & \textbf{0.1013}          & \textbf{64.3\%}                & \textbf{72.5\%}                    & 0.0590                         & 19.1\%          & 18.7\%                         & 0.0505          & 16.7\%          & 8.8\%           \\
        \bottomrule
    \end{tabular}
\end{sidewaystable}

\section{Editing Results}
\label{sec:results}

We executed our editing experiments on three different datasets, commonly used to showcase 3D scene editing: \textit{bicycle} from mip-NeRF 360~\cite{Barron2022}, and \textit{bear} and \textit{face} from Instruct-NeRF2NeRF~\cite{Haque2023}. For each of the datasets we created six prompts for editing that require different amounts of appearance and geometry change. A qualitative comparison of all of the results of our editing pipeline to those of Direct Gaussian Editing (DGE) and \cite{Chen2025d} GaussianEditor \cite{YCZCCZFWXYYWZCLYHLGL2023} is given in \figref{fig:bike_editing}, \figref{fig:bear_editing} and \figref{fig:face_editing} for the respective datasets. Video sequences of the edited scenes can be found alongside this PDF. \figref{fig:editing_2d} shows additional examples of images generated with our method at different truncation levels of $\mathcal{F}_h$, demonstrating the smooth trade-off between semantic richness and abstraction. \figref{fig:finetuning_2d} provides additional examples from the validation split of our splat artifact removal finetuning dataset, highlighting the effectiveness of our approach in mitigating splat artifacts while recovering fine detail.

\section{Ablations}
\label{sec:supp_ablations}

To evaluate the impact of the individual components within our editing pipeline, we performed some ablation studies. We compared our AE-based method of dimensionality reduction $\mathcal{M}^\text{TASE}$ against a PCA of the features produced by the feature backbone, which was computed across $50000$ samples from \mbox{ImageNet-1k}~\cite{Deng2009}. To quantify the influence of the equivariance loss $\mathcal{L}_\text{eqv}$ that we used during AE training to reduce 2D positional bias, we compared against a version of our pipeline without $\mathcal{L}_\text{eqv}$. We also evaluated each of these ablations before and after Difix3D+~\cite{Wu2025c} finetuning. We perform the same local edits as in \secref{sec:results}, the results of which are shown in \figref{fig:bike_ablation}, \figref{fig:bear_ablation}, and \figref{fig:face_ablation} for the datasets \emph{bicycle}, \emph{bear}, and \emph{face} respectively. The CLIP directional similarities for each of the edits are shown in \tabref{tab:supl_ablations}.

To qualitatively demonstrate the controllable abstraction that TASE enables during 2D image generation, we provide some image generation examples using TASE, truncated to different levels, as guidance in \figref{fig:editing_2d}. The original images are taken from the validation set of \mbox{ImageNet-1k}~\cite{Deng2009}. We use a text prompt that is not aligned with the image content (``a hand-drawn cartoon'') to demonstrate that this capability can be used for image editing. In the examples it can be observed that for a low number of retained channels, the generated image more closely follows the text prompt while for a high number of retained channels, the generated image is more closely aligned with the semantic layout of the original image. We also provide some examples from the validation set of our Difix3D+~\cite{Wu2025c} artifact fixing finetuning dataset in \figref{fig:finetuning_2d}, with images generated with our full method before and after the finetuning stage. In these examples it can be clearly seen, that the finetuning reduces blurryness in the generated images and allows the model to reintroduce more finegrained detail.

\begin{sidewaystable}[t]
    \caption{Per-experiment CLIP dir. sim. for each of the ablations.}
    \label{tab:supl_ablations}
    \centering
\begin{tabular}{llcccccc}
        \toprule
        Dataset & Experiment &
        \shortstack{$\mathcal{M}^\text{TASE}$ \\ $+\ \mathcal{L}_\text{eqv}$ \\ $+\ \text{FT}$} &
        \shortstack{$\mathcal{M}^\text{TASE}$ \\ $+\ \mathcal{L}_\text{eqv}$} &
        \shortstack{$\mathcal{M}^\text{TASE}$ \\ $+\ \text{FT}$} &
        $\mathcal{M}^\text{TASE}$ &
        \shortstack{$\text{PCA}$ \\ $+\ \text{FT}$} &
        PCA \\
        \midrule
        bike & blue bike & 0.0303 & $\mathbf{0.0360}$ & 0.0318 & 0.0292 & 0.0204 & 0.0186 \\
        bike & enduro & 0.1234 & 0.0955 & 0.1080 & 0.0794 & $\mathbf{0.1257}$ & 0.1040 \\
        bike & motorcycle & $\mathbf{0.1114}$ & 0.0850 & 0.0646 & 0.0651 & 0.0780 & 0.0987 \\
        bike & motorcycle race & $\mathbf{0.1058}$ & 0.0956 & 0.0575 & 0.0612 & 0.0808 & 0.0847 \\
        bike & mtb & 0.0579 & $\mathbf{0.0686}$ & 0.0496 & 0.0516 & 0.0585 & 0.0675 \\
        bike & scooter & 0.1398 & 0.1127 & 0.1039 & 0.1141 & $\mathbf{0.1488}$ & 0.1345 \\
        bear & brown bear & $\mathbf{0.0783}$ & 0.0734 & 0.0669 & 0.0670 & 0.0715 & 0.0653 \\
\bottomrule
    \end{tabular}
\end{sidewaystable}

\newcommand{\promptcell}[1]{%
    \linespread{0.75}\selectfont
    \textit{\scriptsize #1}%
}%

\def\imgw{0.16\textwidth}

\begin{figure}[t]
    \centering

    \begin{tabular}{{>{\centering\arraybackslash}m{\imgw}>{\centering\arraybackslash}m{0.2\textwidth}>{\centering\arraybackslash}m{\imgw}>{\centering\arraybackslash}m{\imgw}>{\centering\arraybackslash}m{\imgw}}}
        \toprule
        Original                                                                                                                                          & Prompt                                                                                                                                                  & Ours                                                                                     & DGE & GE \\
        \midrule
        \includegraphics[width=\imgw]{pics/baseline_comparisons/bike/orig_0.jpg} \includegraphics[width=\imgw]{pics/baseline_comparisons/bike/orig_1.jpg} & \promptcell{a light blue classic road bike with a brown and chrome handlebar, a brown saddle and beige tires}                                               & \includegraphics[width=\imgw]{pics/baseline_comparisons/bike/blue_bike_Ours_0.jpg}
        \includegraphics[width=\imgw]{pics/baseline_comparisons/bike/blue_bike_Ours_1.jpg}                                                                & \includegraphics[width=\imgw]{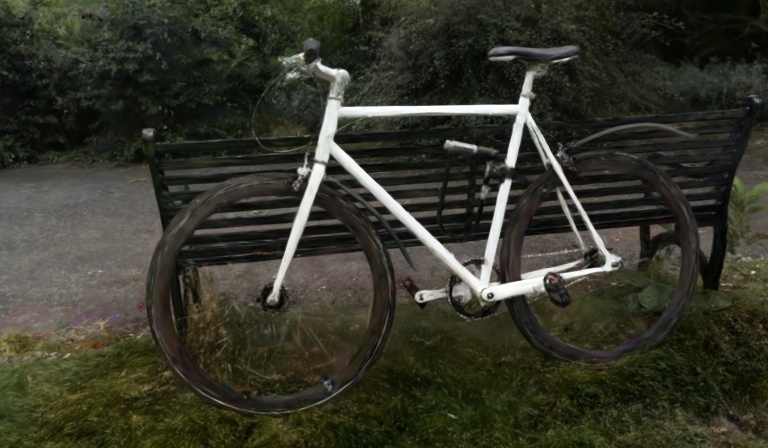}
        \includegraphics[width=\imgw]{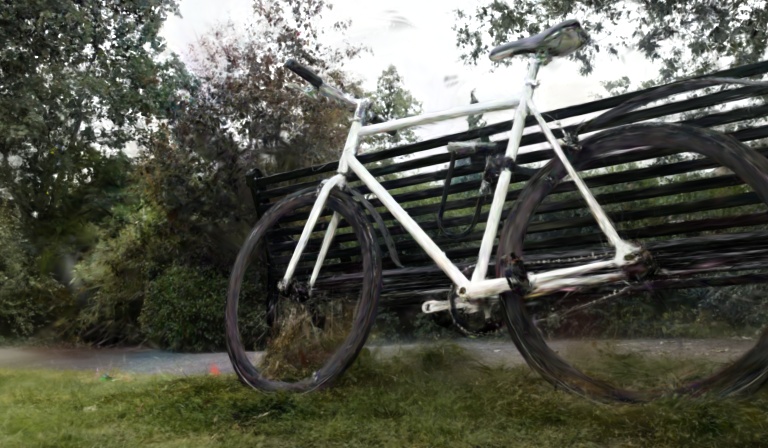}                                                                 & \includegraphics[width=\imgw]{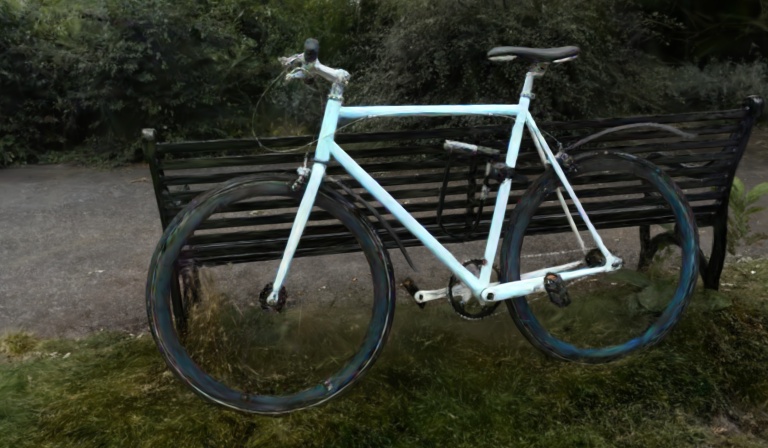}
        \includegraphics[width=\imgw]{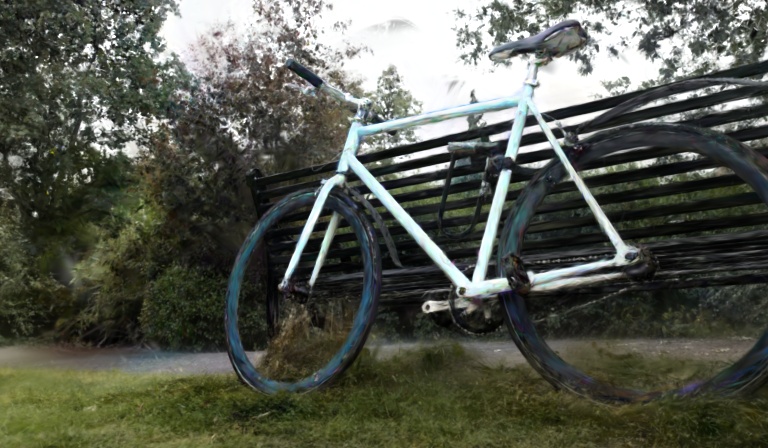}                                                                                                                                                                                                                                                                                                                                  \\
\bottomrule
    \end{tabular}
    \caption{Qualitative comparison of our editing results against the state-of-the-art 3D scene editing methods Direct Gaussian Editing~(DGE)~\cite{Chen2025d} and GaussianEditor~(GE)~\cite{YCZCCZFWXYYWZCLYHLGL2023} for local 3D scene editing on the dataset \emph{bicycle}.}
    \label{fig:bike_editing}
\end{figure}

\begin{figure}[t]
    \centering
    \begin{tabular}{{>{\centering\arraybackslash}m{\imgw}>{\centering\arraybackslash}m{0.2\textwidth}>{\centering\arraybackslash}m{\imgw}>{\centering\arraybackslash}m{\imgw}>{\centering\arraybackslash}m{\imgw}}}
        \toprule
        Original                                                                                                                                          & Prompt                                                                                              & Ours                                                                                      & DGE & GE \\
        \midrule
        \includegraphics[width=\imgw]{pics/baseline_comparisons/bear/orig_0.jpg} \includegraphics[width=\imgw]{pics/baseline_comparisons/bear/orig_1.jpg} & \promptcell{a brown bear in a forest}                                                                   & \includegraphics[width=\imgw]{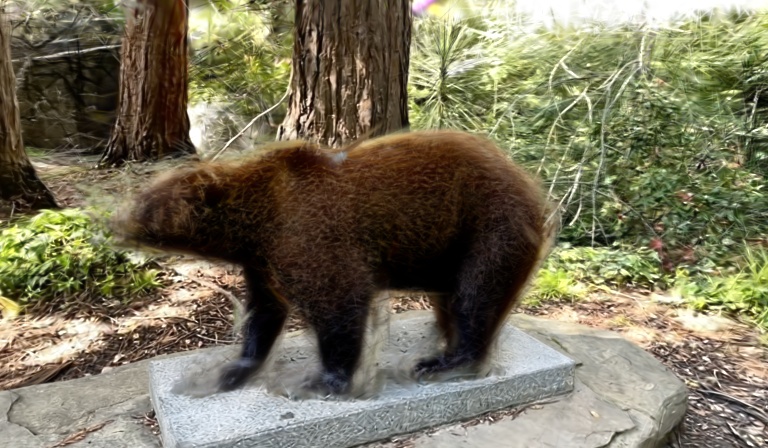}
        \includegraphics[width=\imgw]{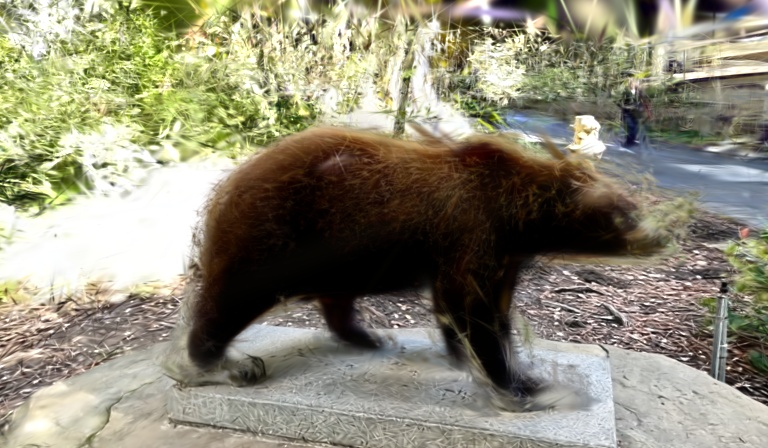}                                                               & \includegraphics[width=\imgw]{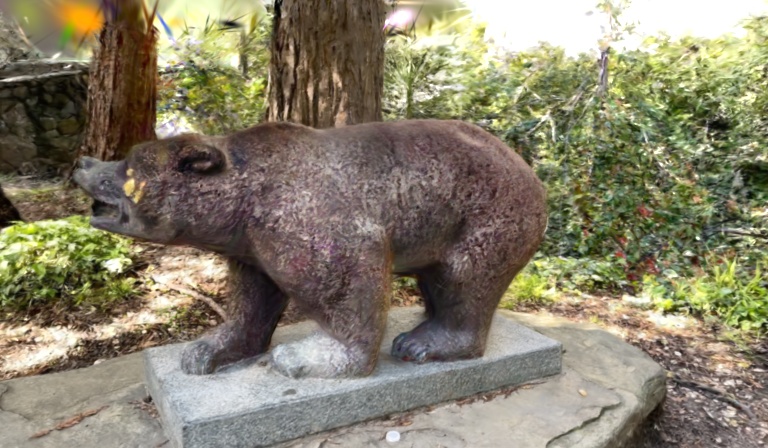}
        \includegraphics[width=\imgw]{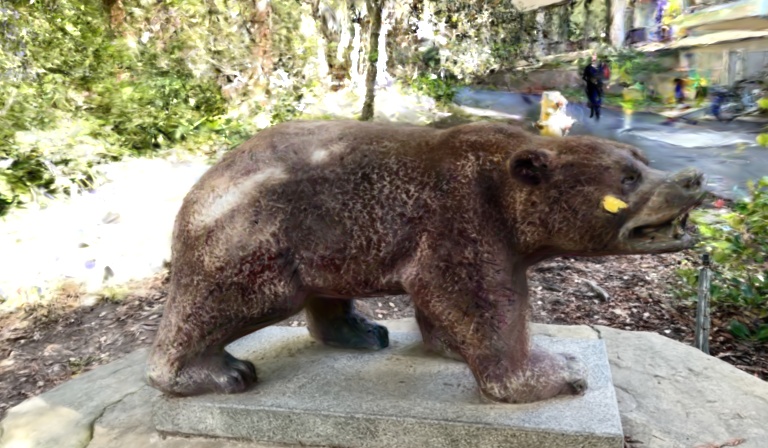}                                                                & \includegraphics[width=\imgw]{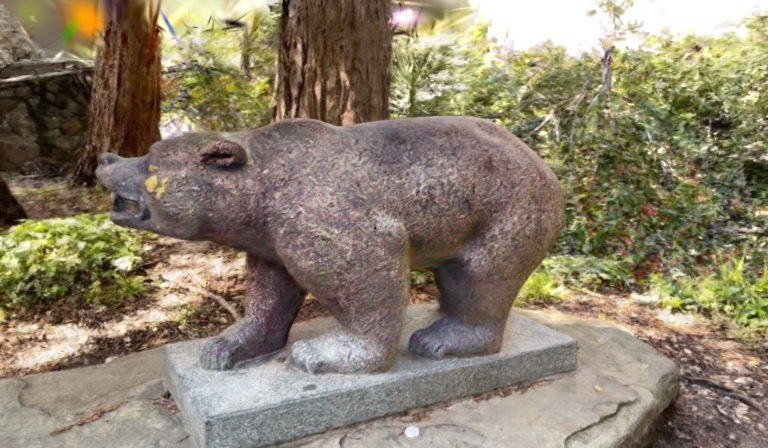}
        \includegraphics[width=\imgw]{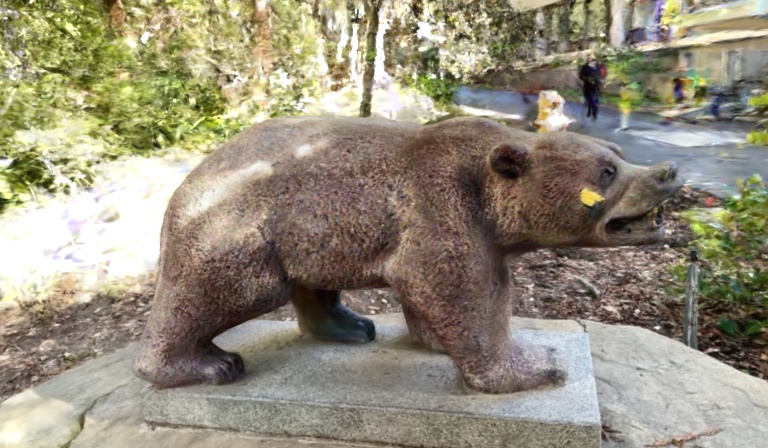}                                                                                                                                                                                                                                                                              \\
\bottomrule
    \end{tabular}
    \caption{Qualitative comparison of our editing results against the state-of-the-art 3D scene editing methods Direct Gaussian Editing~(DGE)~\cite{Chen2025d} and GaussianEditor~(GE)~\cite{YCZCCZFWXYYWZCLYHLGL2023} for local 3D scene editing on the dataset \emph{bear}.}
    \label{fig:bear_editing}
\end{figure}

\begin{figure}[t]
    \centering
    \begin{tabular}{{>{\centering\arraybackslash}m{\imgw}>{\centering\arraybackslash}m{0.2\textwidth}>{\centering\arraybackslash}m{\imgw}>{\centering\arraybackslash}m{\imgw}>{\centering\arraybackslash}m{\imgw}}}
        \toprule
        Original                                                                                                                                          & Prompt                                                                                                             & Ours                                                                              & DGE & GE \\
        \midrule
        \includegraphics[width=\imgw]{pics/baseline_comparisons/face/orig_0.jpg} \includegraphics[width=\imgw]{pics/baseline_comparisons/face/orig_1.jpg} & \promptcell{a circus clown with white makeup, a red nose, red cheeks and red hair}                                     & \includegraphics[width=\imgw]{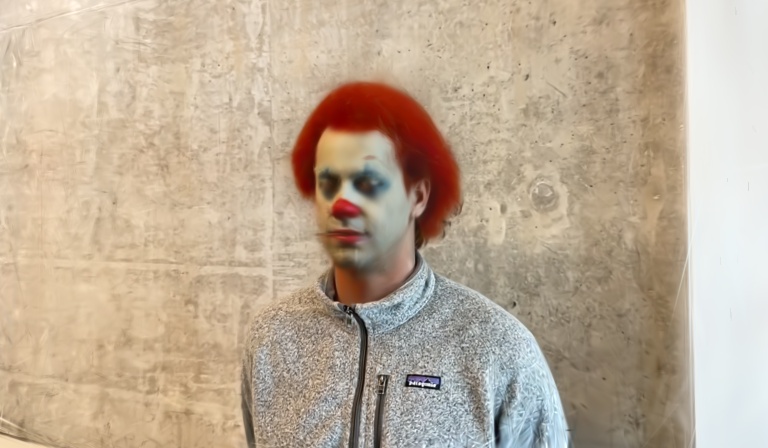}
        \includegraphics[width=\imgw]{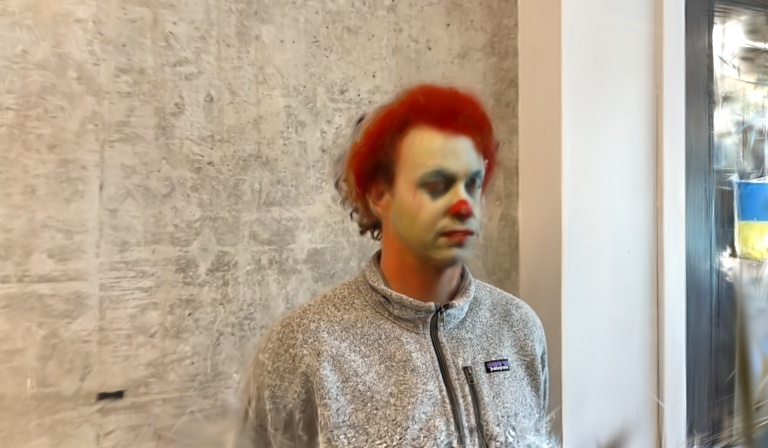}                                                                    & \includegraphics[width=\imgw]{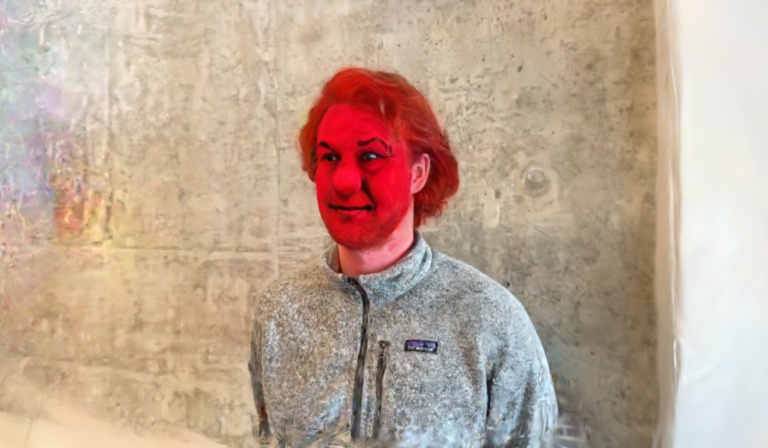}
        \includegraphics[width=\imgw]{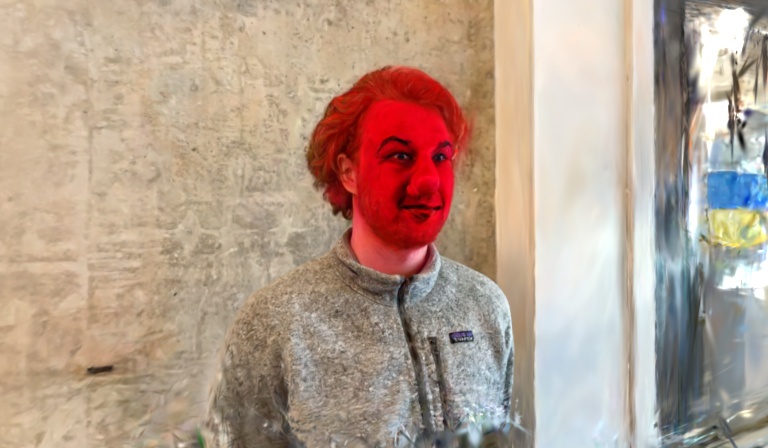}                                                                     & \includegraphics[width=\imgw]{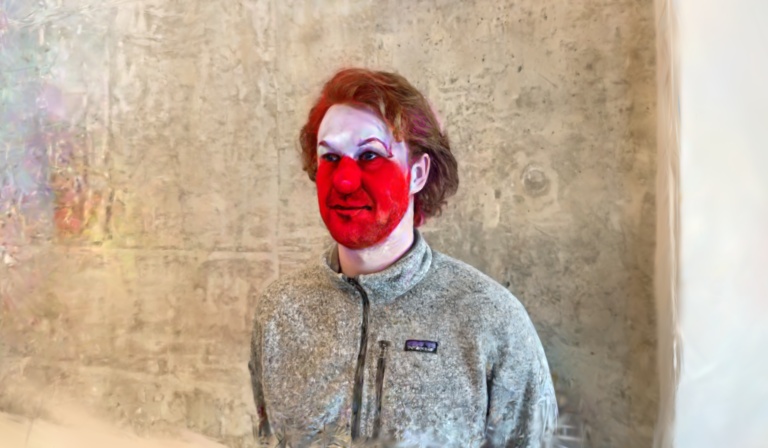}
        \includegraphics[width=\imgw]{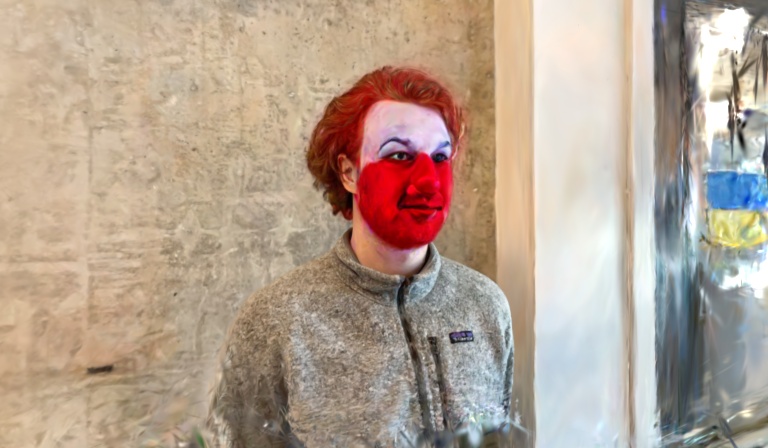}                                                                                                                                                                                                                                                                                          \\
\bottomrule
    \end{tabular}
    \caption{Qualitative comparison of our editing results against the state-of-the-art 3D scene editing methods Direct Gaussian Editing~(DGE)~\cite{Chen2025d} and GaussianEditor~(GE)~\cite{YCZCCZFWXYYWZCLYHLGL2023} for local 3D scene editing on the dataset \emph{face}.}
    \label{fig:face_editing}
\end{figure}

\def\imgw{0.16\textwidth}
\def\txtw{0.2\textwidth}

\newcommand{\cellcheck}[1]{%
 \cmark \hspace{2cm} \phantom{.} \hspace{2cm} \phantom{.} \hspace{2cm} \xmark
}%

\begin{figure}[t]
    \centering
    \begin{tabular}{{>{\centering\arraybackslash}m{\imgw}>{\centering\arraybackslash}m{\txtw}>{\centering\arraybackslash}m{0.05\textwidth}>{\centering\arraybackslash}m{\imgw}>{\centering\arraybackslash}m{\imgw}>{\centering\arraybackslash}m{\imgw}}}
    \toprule
    Original    & Prompt & FT   & $\mathcal{M}^\text{TASE} + \mathcal{L}_\text{eqv}$ &  $\mathcal{M}^\text{TASE}$  & PCA \\                                                                             
    \midrule
    \includegraphics[width=\imgw]{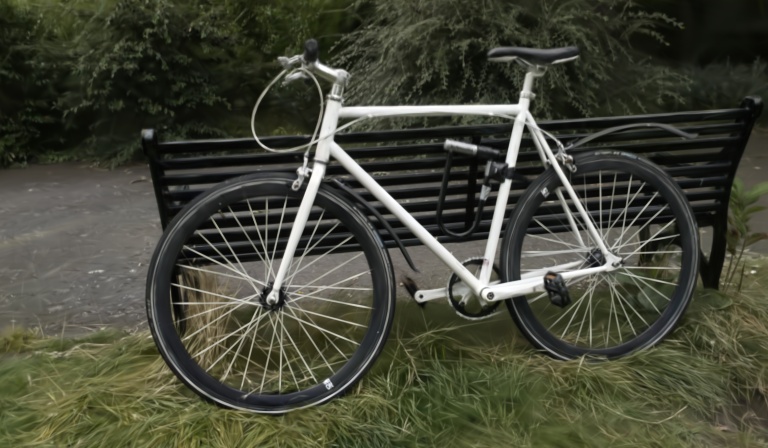} \includegraphics[width=\imgw]{pics/ablations/bike/orig_0.jpg} & \promptcell{a light blue classic road bike with a brown and chrome handlebar, a brown saddle and beige tires} & \cellcheck && \includegraphics[width=\imgw]{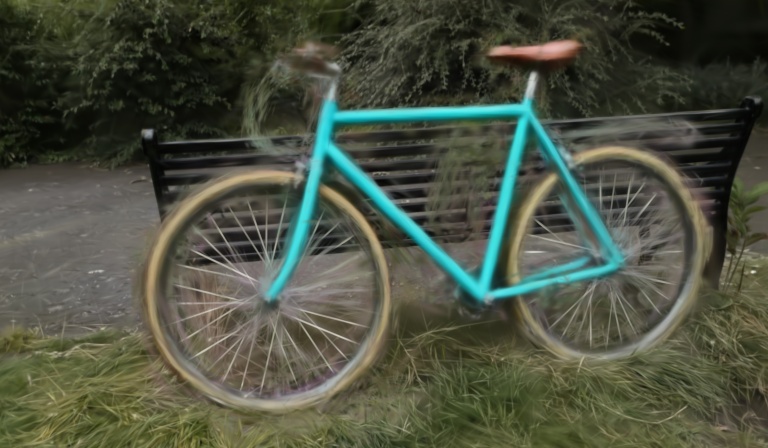} \includegraphics[width=\imgw]{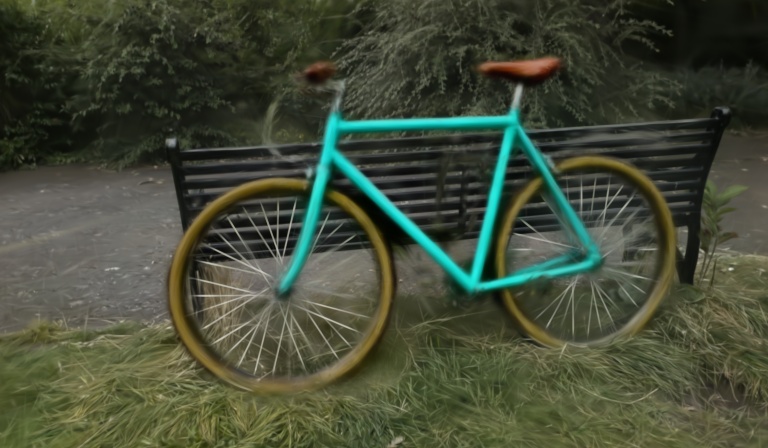} & \includegraphics[width=\imgw]{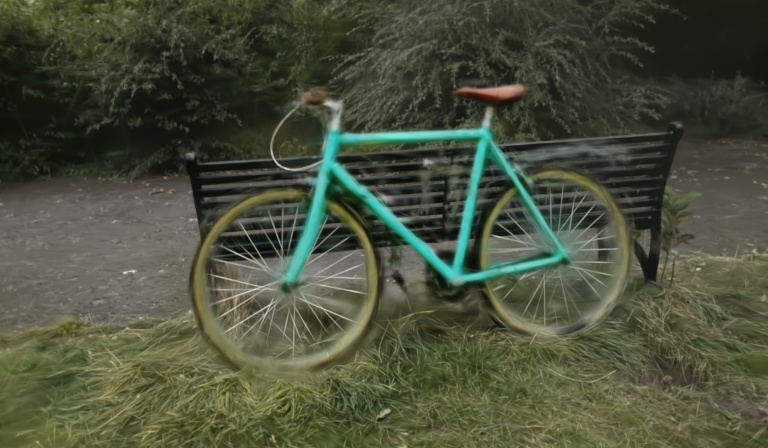} \includegraphics[width=\imgw]{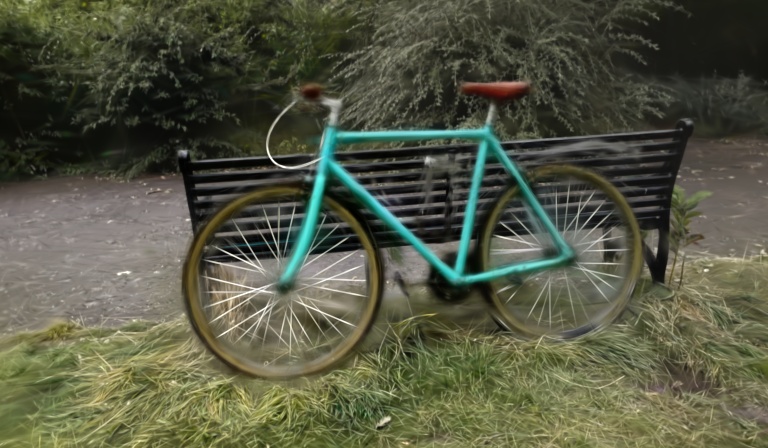} & \includegraphics[width=\imgw]{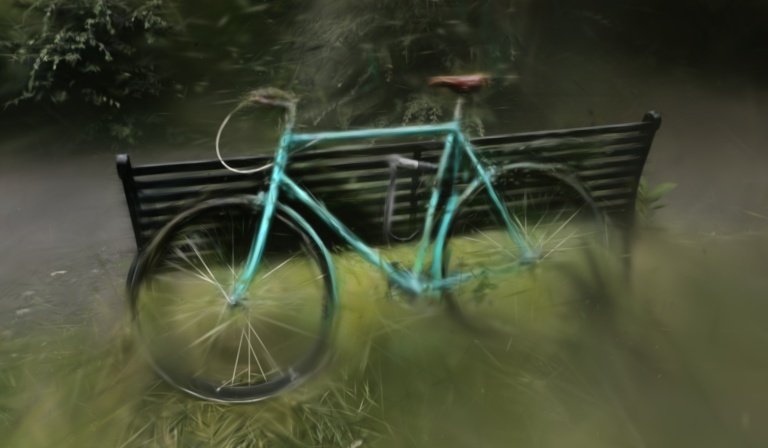} \includegraphics[width=\imgw]{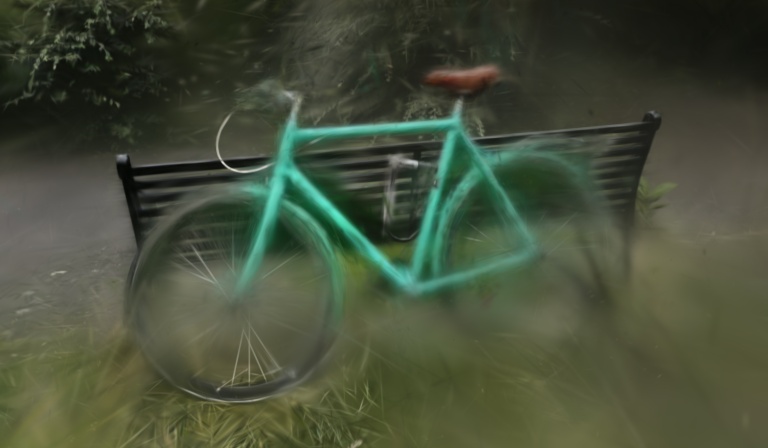} \\
    \midrule
    \includegraphics[width=\imgw]{pics/ablations/bike/orig_0.jpg} \includegraphics[width=\imgw]{pics/ablations/bike/orig_0.jpg} & \promptcell{a bright orange enduro dirt bike with a low black saddle, knobby tires, a big chrome front suspension fork, disk brakes and big orange fenders} & \cellcheck && \includegraphics[width=\imgw]{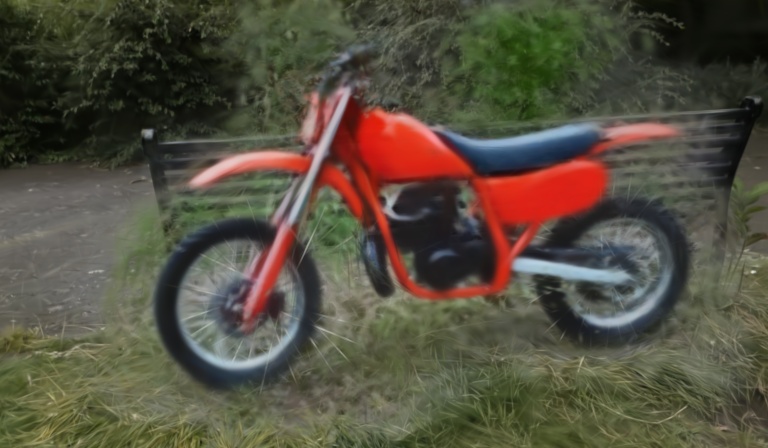}  \includegraphics[width=\imgw]{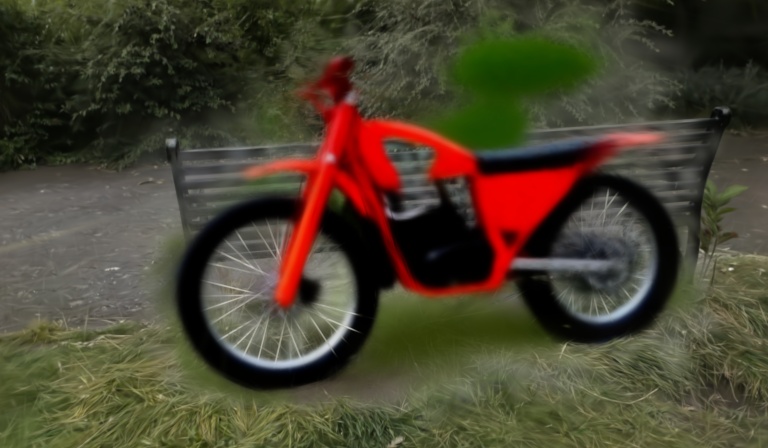} & \includegraphics[width=\imgw]{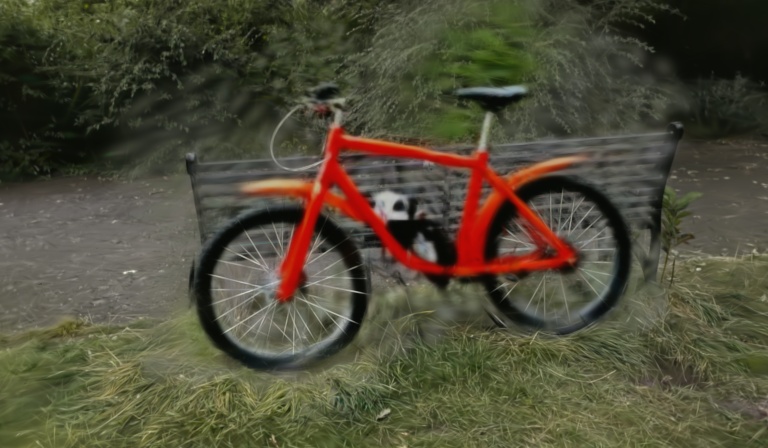} \includegraphics[width=\imgw]{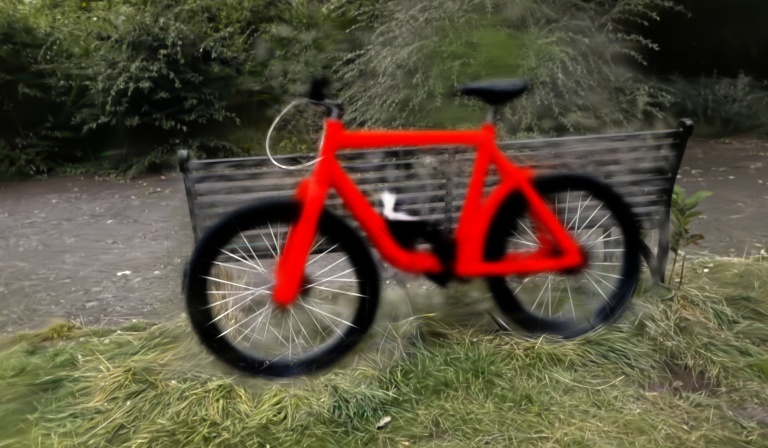} & \includegraphics[width=\imgw]{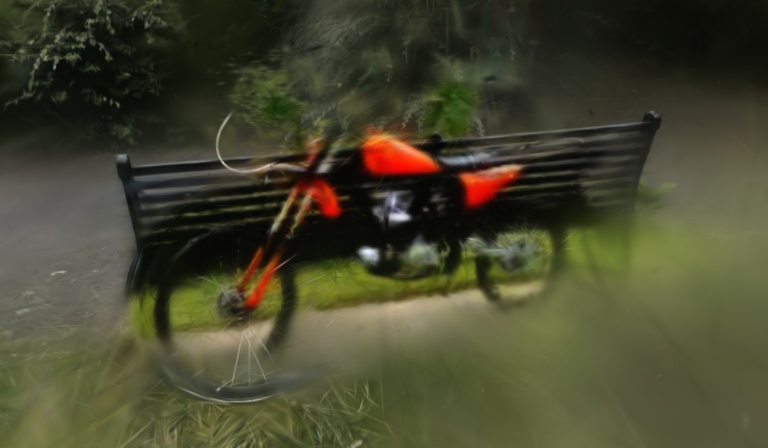} \includegraphics[width=\imgw]{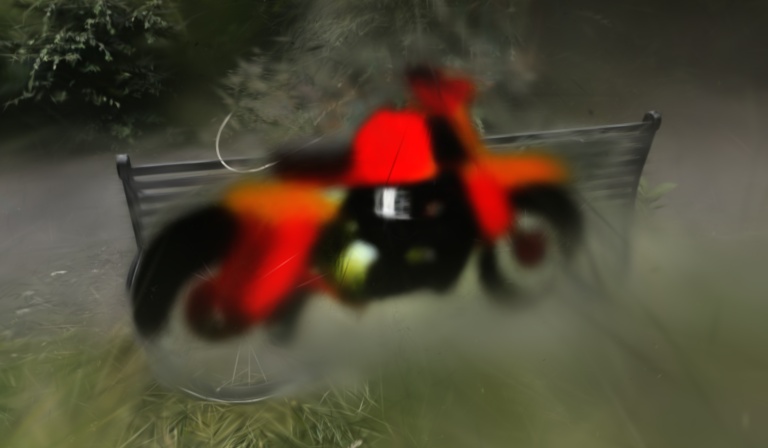} \\
\bottomrule
    \end{tabular}
    \caption{Qualitative ablation of: the method of dimensionality reduction ($\mathcal{M}^{\text{TASE}}$ vs. PCA), the equivariance loss $\mathcal{L}_{\text{eqv}}$, and the Difix3D+~\cite{Wu2025c} finetuning strategy (FT), on the dataset \emph{bicycle}. Our full method is $\mathcal{M}^{\text{TASE}} + \mathcal{L}_\text{eqv} + \text{FT}$.}
    \label{fig:bike_ablation}
\end{figure}

\begin{figure}[t]
    \centering
    \begin{tabular}{{>{\centering\arraybackslash}m{\imgw}>{\centering\arraybackslash}m{\txtw}>{\centering\arraybackslash}m{0.05\textwidth}>{\centering\arraybackslash}m{\imgw}>{\centering\arraybackslash}m{\imgw}>{\centering\arraybackslash}m{\imgw}}}
    \toprule
    Original    & Prompt & FT   & $\mathcal{M}^\text{TASE} + \mathcal{L}_\text{eqv}$ &  $\mathcal{M}^\text{TASE}$  & PCA \\                                                                             
    \midrule
    \includegraphics[width=\imgw]{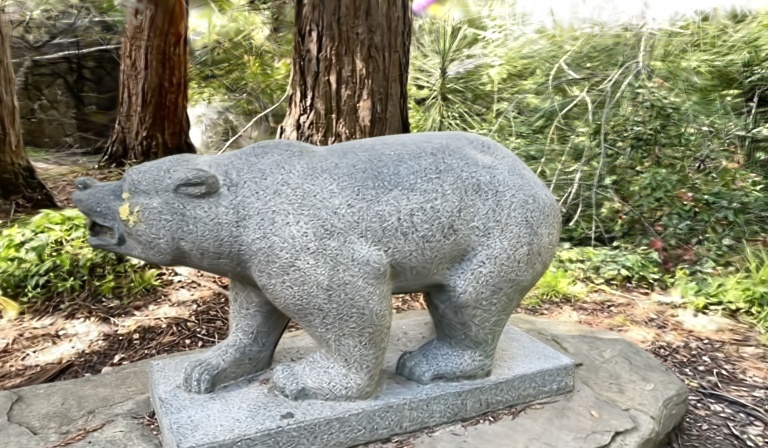} \includegraphics[width=\imgw]{pics/ablations/bear/orig_0.jpg}
    & \promptcell{A brown bear in a forest}
    & \cellcheck
    && \includegraphics[width=\imgw]{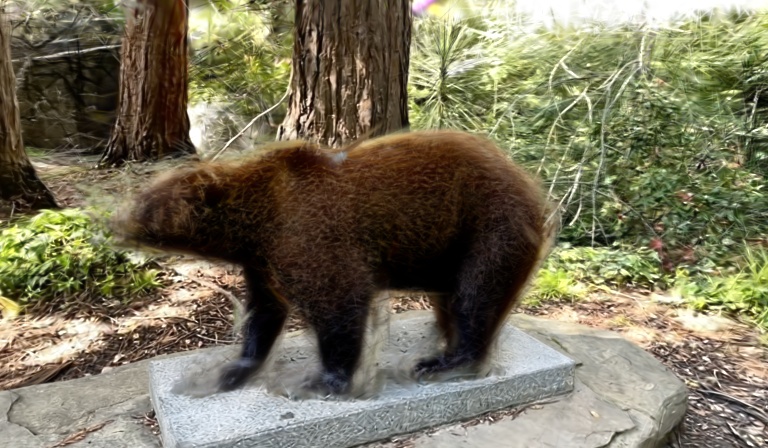} \includegraphics[width=\imgw]{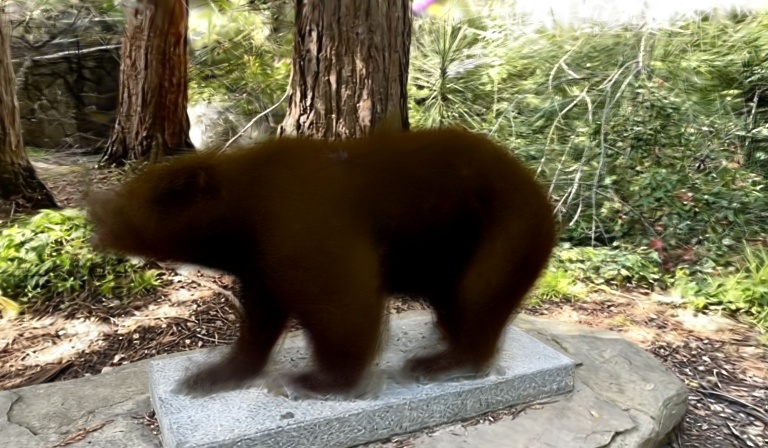}
    & \includegraphics[width=\imgw]{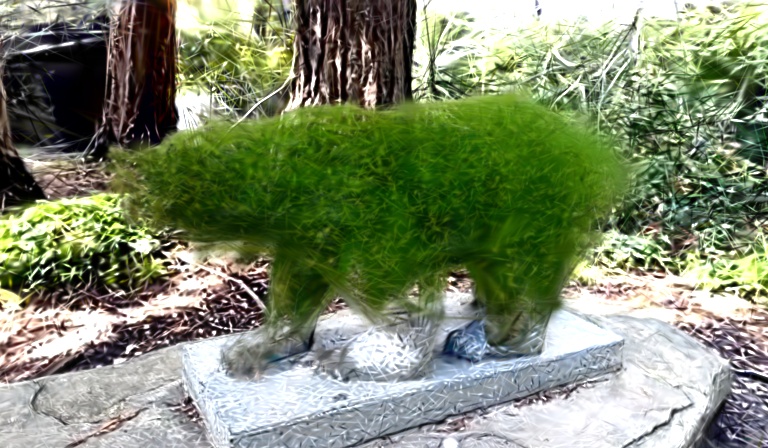} \includegraphics[width=\imgw]{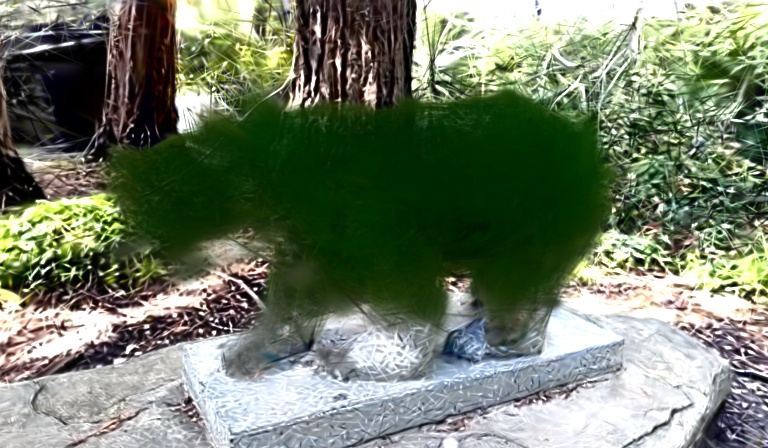}
    & \includegraphics[width=\imgw]{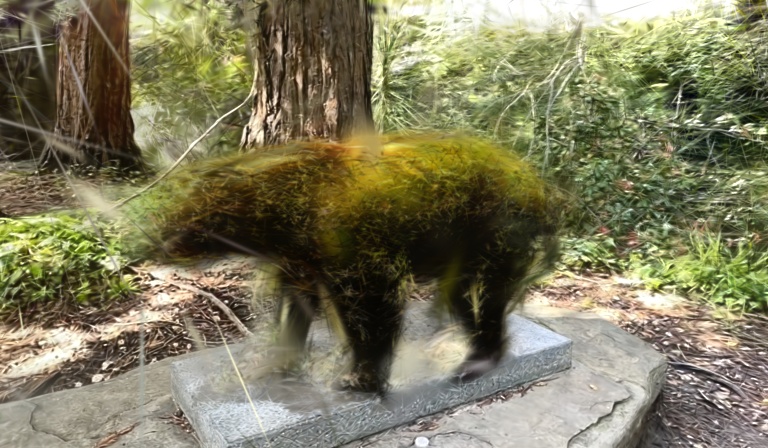} \includegraphics[width=\imgw]{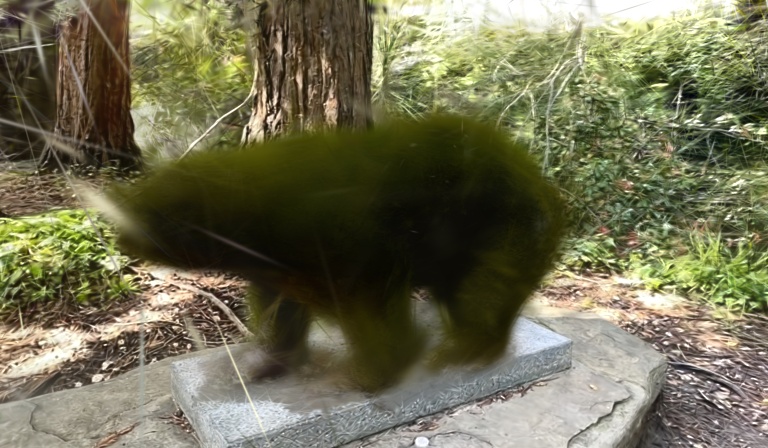} \\
    \midrule
    \includegraphics[width=\imgw]{pics/ablations/bear/orig_0.jpg} \includegraphics[width=\imgw]{pics/ablations/bear/orig_0.jpg}
    & \promptcell{a golden retriever dog with a shiny golden coat and floppy ears}
    & \cellcheck
    && \includegraphics[width=\imgw]{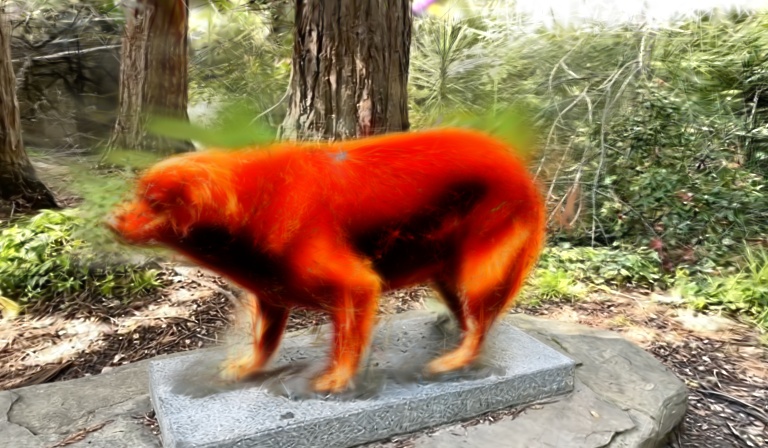} \includegraphics[width=\imgw]{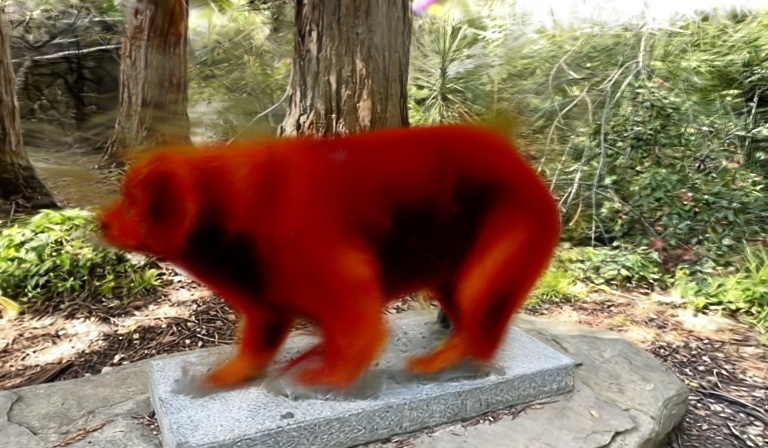}
    & \includegraphics[width=\imgw]{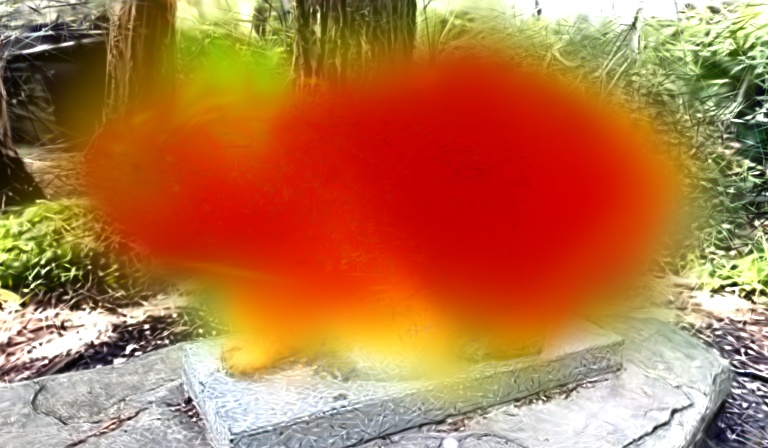} \includegraphics[width=\imgw]{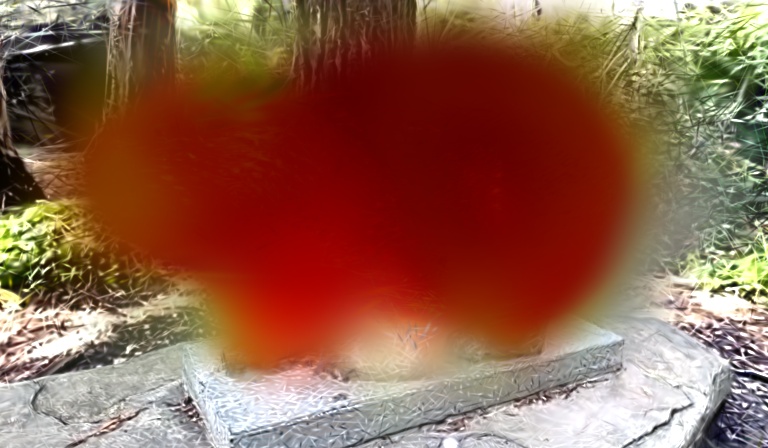}
    & \includegraphics[width=\imgw]{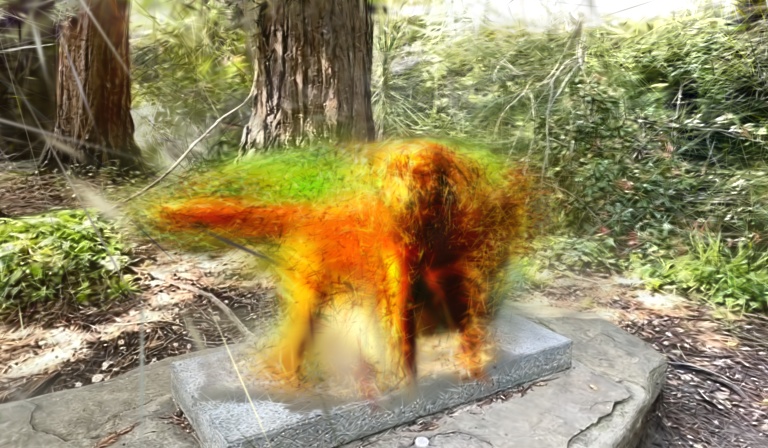} \includegraphics[width=\imgw]{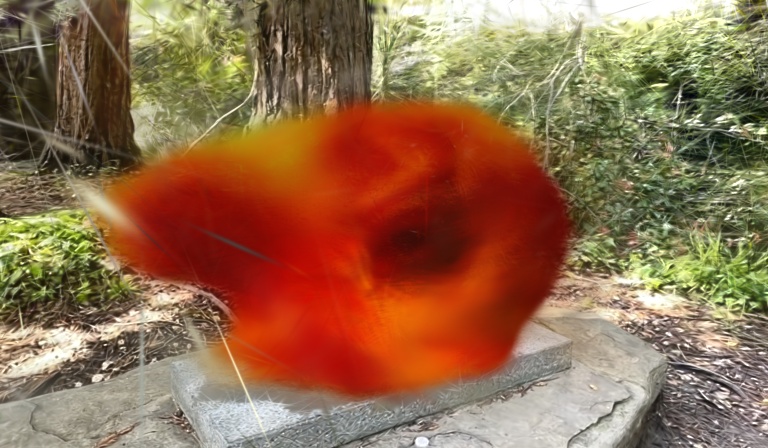} \\
\bottomrule
    \end{tabular}
    \caption{Qualitative ablation of: the method of dimensionality reduction ($\mathcal{M}^{\text{TASE}}$ vs. PCA), the equivariance loss $\mathcal{L}_{\text{eqv}}$, and the Difix3D+~\cite{Wu2025c} finetuning strategy (FT), on the dataset \emph{bear}. Our full method is $\mathcal{M}^{\text{TASE}} + \mathcal{L}_\text{eqv} + \text{FT}$.}
    \label{fig:bear_ablation}
\end{figure}

\begin{figure}[t]
    \centering
    \begin{tabular}{{>{\centering\arraybackslash}m{\imgw}>{\centering\arraybackslash}m{\txtw}>{\centering\arraybackslash}m{0.05\textwidth}>{\centering\arraybackslash}m{\imgw}>{\centering\arraybackslash}m{\imgw}>{\centering\arraybackslash}m{\imgw}}}
    \toprule
    Original    & Prompt & FT   & $\mathcal{M}^\text{TASE} + \mathcal{L}_\text{eqv}$ &  $\mathcal{M}^\text{TASE}$  & PCA \\                                                                             
    \midrule
    \includegraphics[width=\imgw]{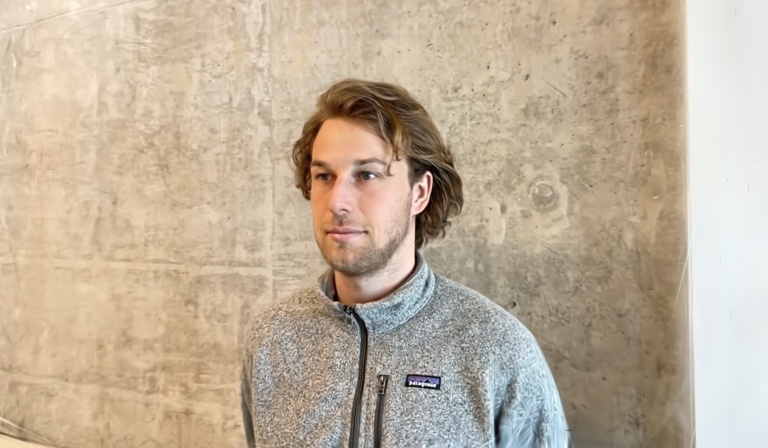} \includegraphics[width=\imgw]{pics/ablations/face/orig_0.jpg}
    & \promptcell{a circus clown with white makeup, a red nose, red cheeks and red hair}
    & \cellcheck
    && \includegraphics[width=\imgw]{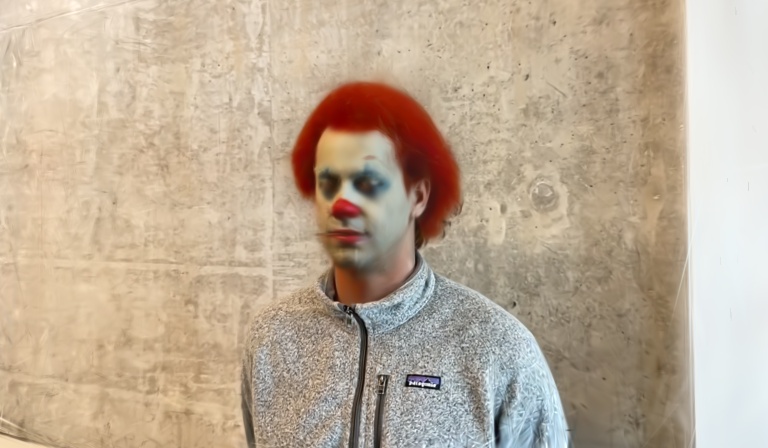} \includegraphics[width=\imgw]{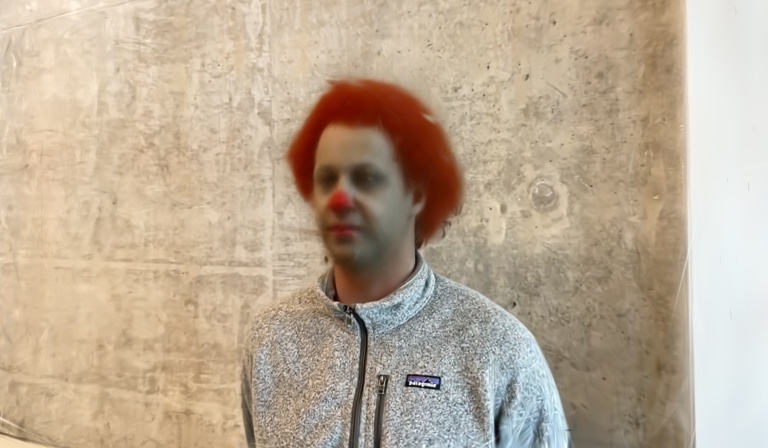}
    & \includegraphics[width=\imgw]{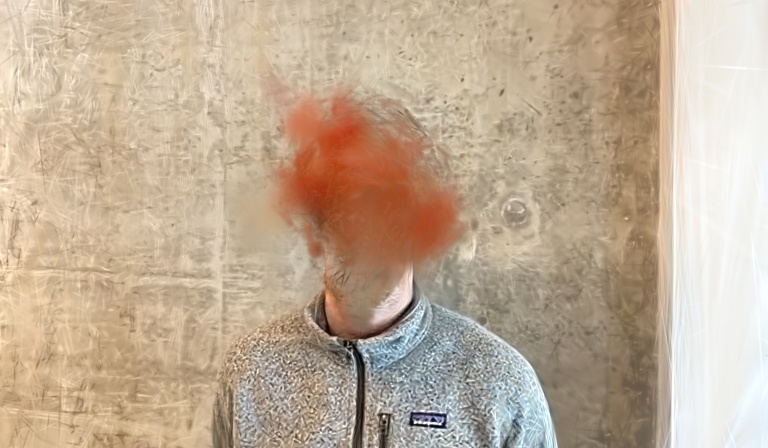} \includegraphics[width=\imgw]{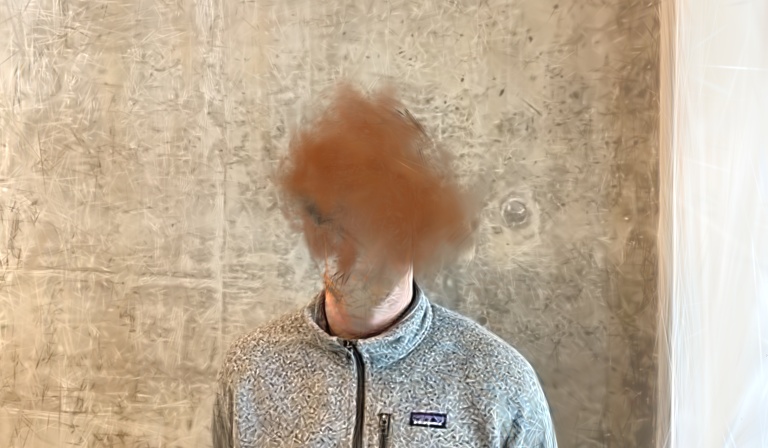}
    & \includegraphics[width=\imgw]{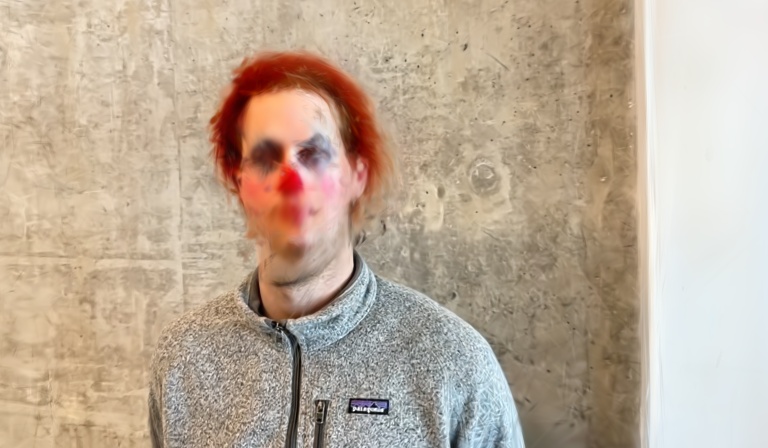} \includegraphics[width=\imgw]{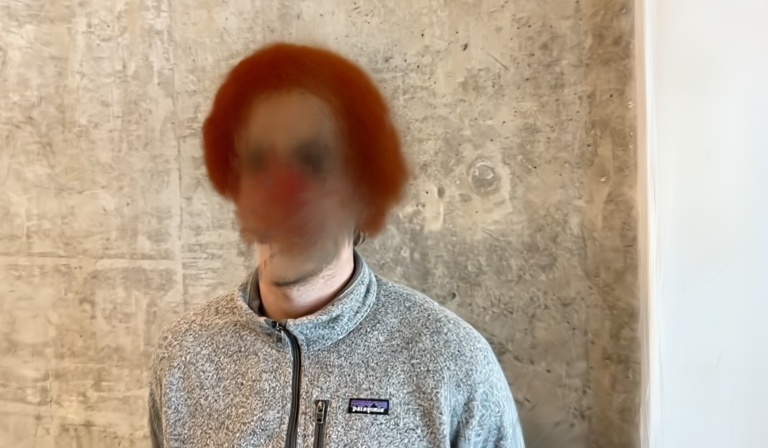} \\
    \midrule
    \includegraphics[width=\imgw]{pics/ablations/face/orig_0.jpg} \includegraphics[width=\imgw]{pics/ablations/face/orig_0.jpg}
    & \promptcell{albert einstein with wild gray hair and mustache}
    & \cellcheck
    && \includegraphics[width=\imgw]{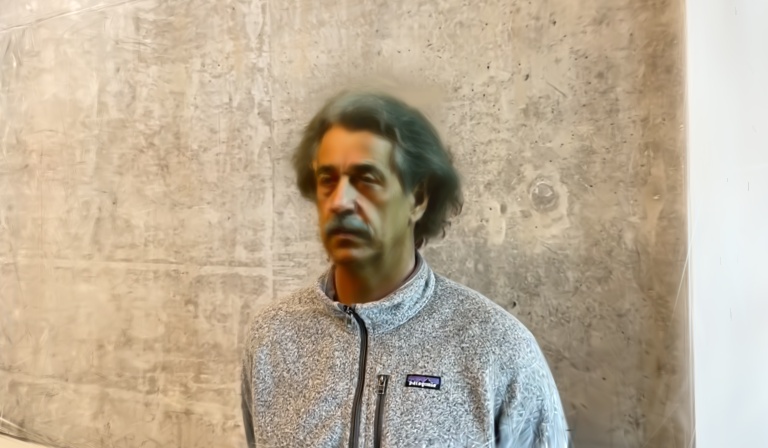} \includegraphics[width=\imgw]{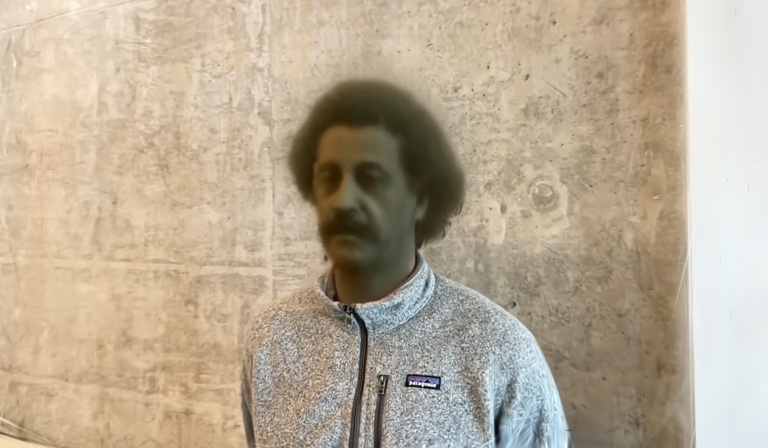}
    & \includegraphics[width=\imgw]{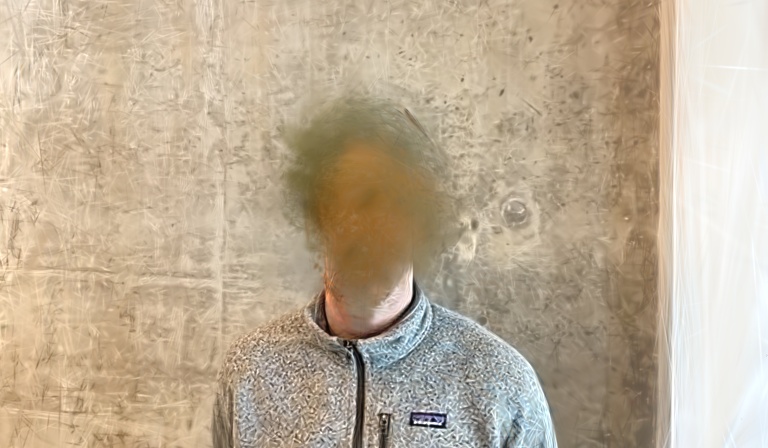} \includegraphics[width=\imgw]{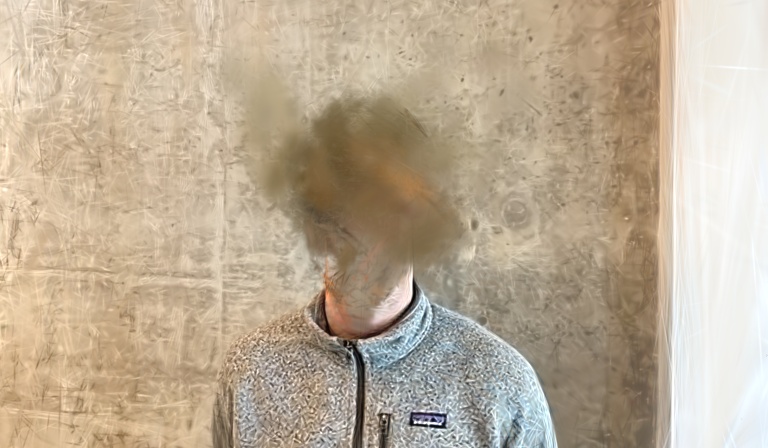}
    & \includegraphics[width=\imgw]{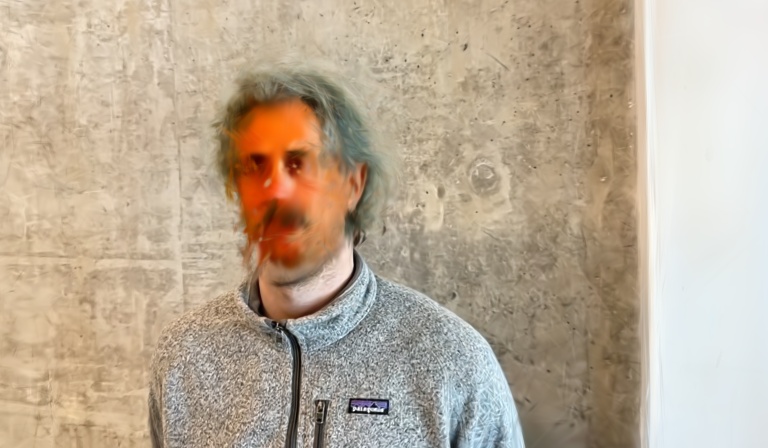} \includegraphics[width=\imgw]{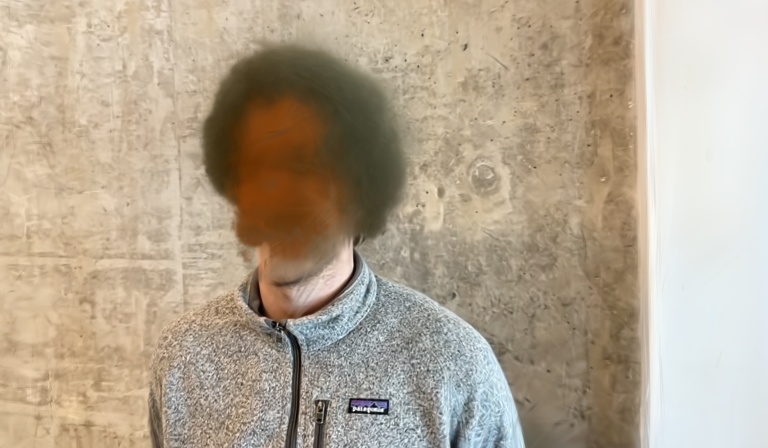} \\
\bottomrule
    \end{tabular}
    \caption{Qualitative ablation of: the method of dimensionality reduction ($\mathcal{M}^{\text{TASE}}$ vs. PCA), the equivariance loss $\mathcal{L}_{\text{eqv}}$, and the Difix3D+~\cite{Wu2025c} finetuning strategy (FT), on the dataset \emph{face}.  Our full method is $\mathcal{M}^{\text{TASE}} + \mathcal{L}_\text{eqv} + \text{FT}$.}
    \label{fig:face_ablation}
\end{figure}

\def\imgw2{0.125\textwidth}

\begin{figure}[t]
    \centering
    \begin{tabular}{>{\centering\arraybackslash}m{\imgw2}>{\centering\arraybackslash}m{\imgw2}>{\centering\arraybackslash}m{\imgw2}>{\centering\arraybackslash}m{\imgw2}>{\centering\arraybackslash}m{\imgw2}>{\centering\arraybackslash}m{\imgw2}>{\centering\arraybackslash}m{\imgw2}}
        \toprule
        $\mathcal{F}_h$                                                                           & Original                                                                           & 64 dims                                                                            & 16 dims                                                                            & 8 dims                                                                            & 4 dims                                                                            & 2 dims                                                                            \\
        \midrule
        \includegraphics[width=\imgw2]{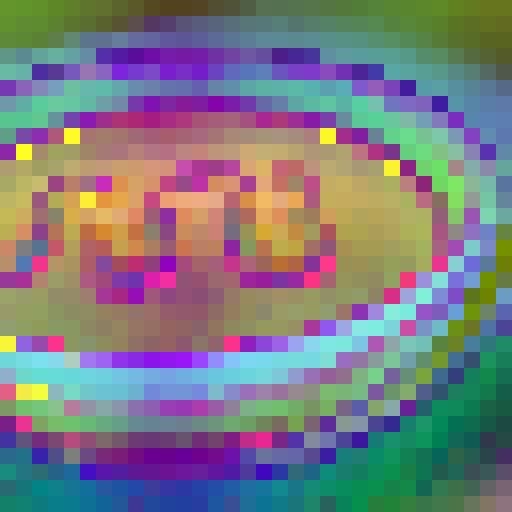}  & \includegraphics[width=\imgw2]{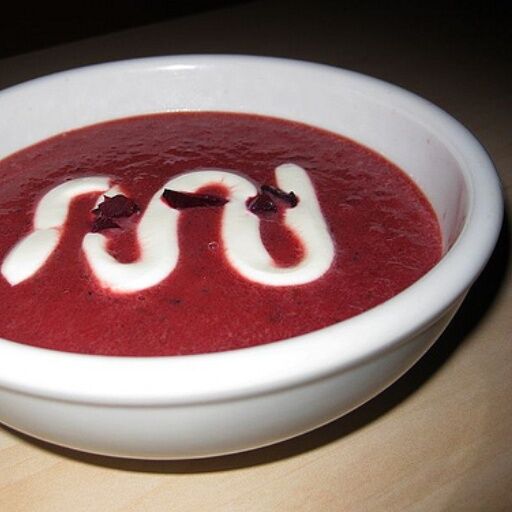}  & \includegraphics[width=\imgw2]{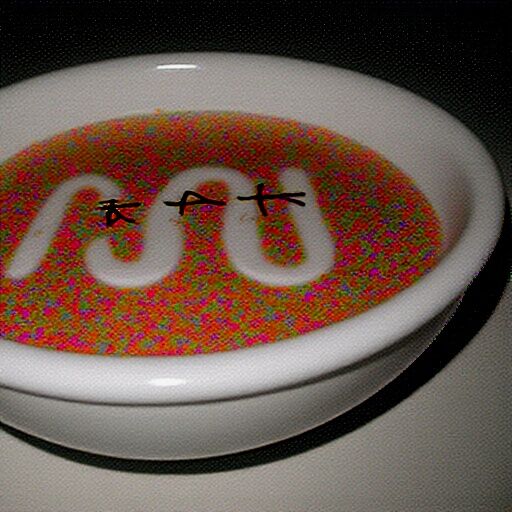}  & \includegraphics[width=\imgw2]{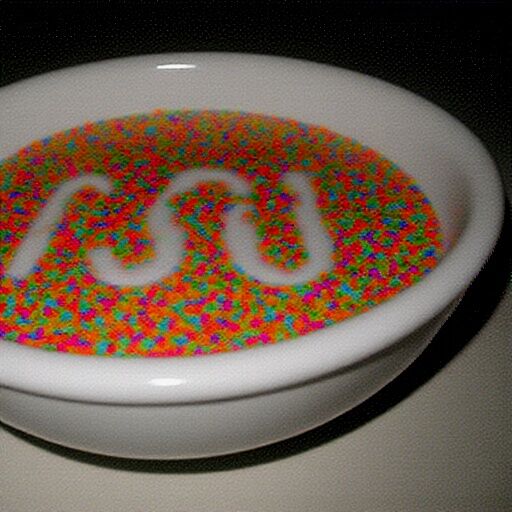}  & \includegraphics[width=\imgw2]{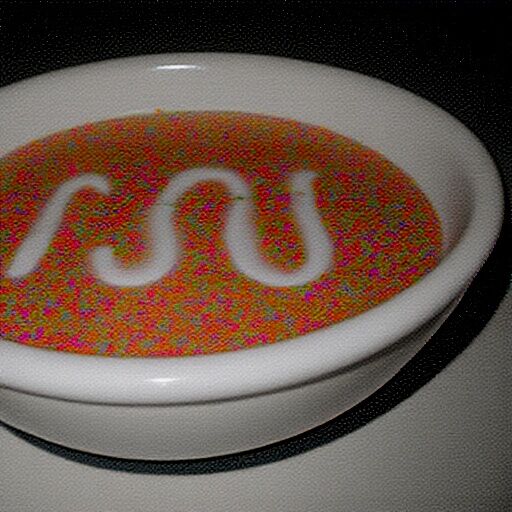}  & \includegraphics[width=\imgw2]{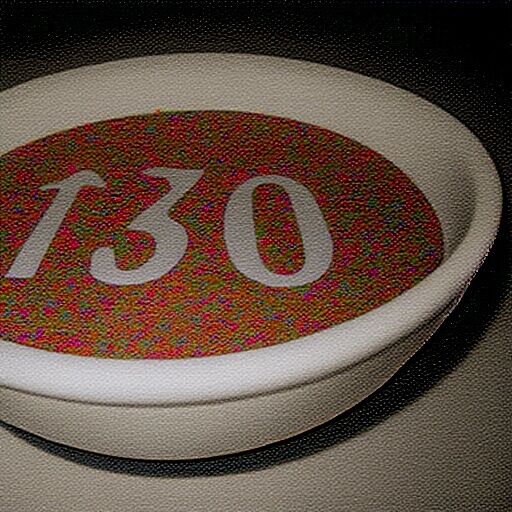}  & \includegraphics[width=\imgw2]{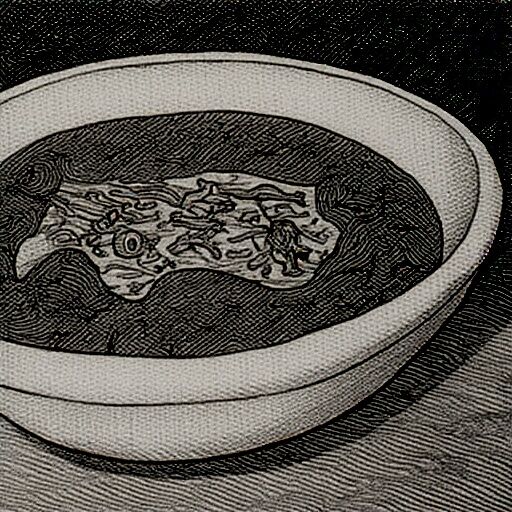}  \\
\bottomrule
    \end{tabular}
    \caption{Qualitative 2D image generation results on samples from the ImageNet-1k validation set with the prompt ``\textit{a hand-drawn cartoon}''.}
    \label{fig:editing_2d}
\end{figure}

\begin{figure}[t]
    \centering
    \begin{tabular}{>{\centering\arraybackslash}m{\imgw2}>{\centering\arraybackslash}m{\imgw2}>{\centering\arraybackslash}m{\imgw2}>{\centering\arraybackslash}m{\imgw2}>{\centering\arraybackslash}m{\imgw2}}
        \toprule
        $\mathcal{F}_h$                                                                     & Corrupted                                                                  &             Target                                                      & With FT                                                                  &       No FT                                                       \\
        \midrule
        \includegraphics[width=\imgw2]{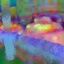}  & \includegraphics[width=\imgw2]{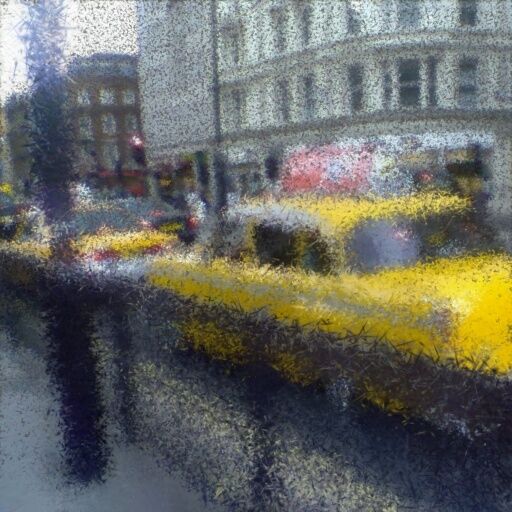}  & \includegraphics[width=\imgw2]{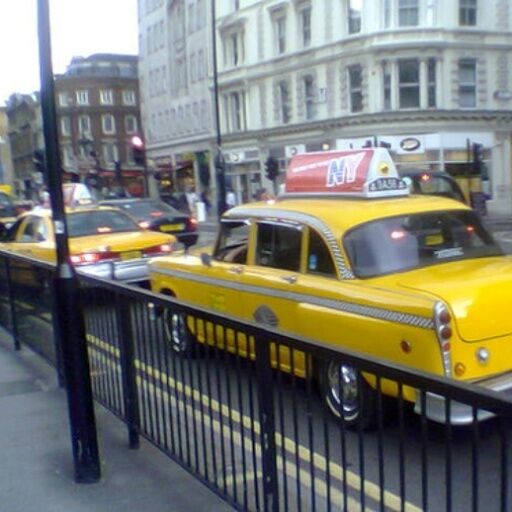}  & \includegraphics[width=\imgw2]{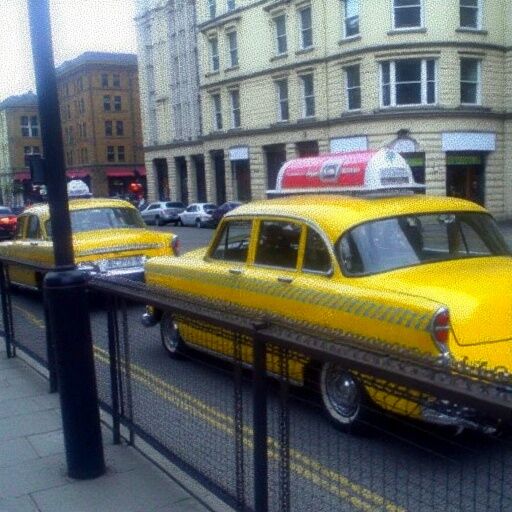}  & \includegraphics[width=\imgw2]{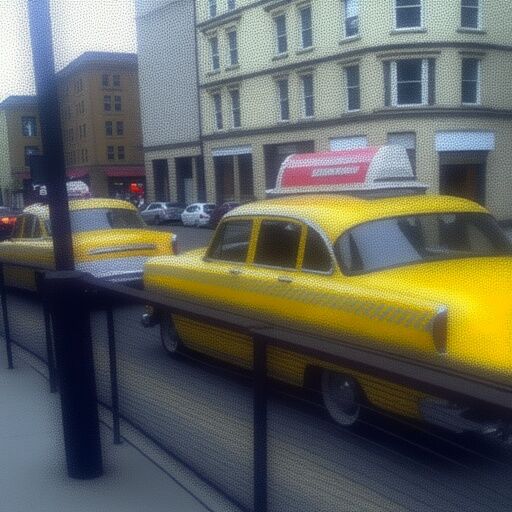}  \\
        \includegraphics[width=\imgw2]{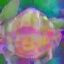} & \includegraphics[width=\imgw2]{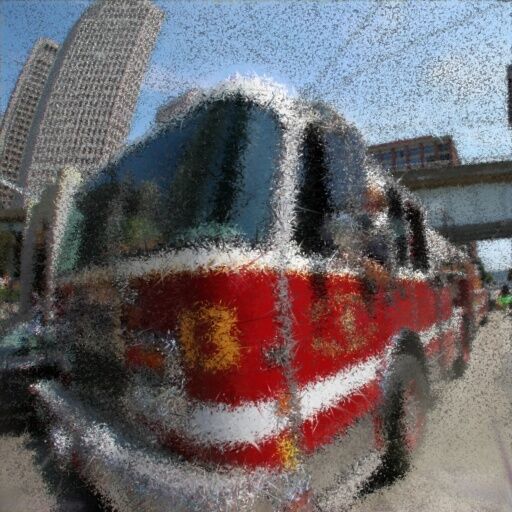} & \includegraphics[width=\imgw2]{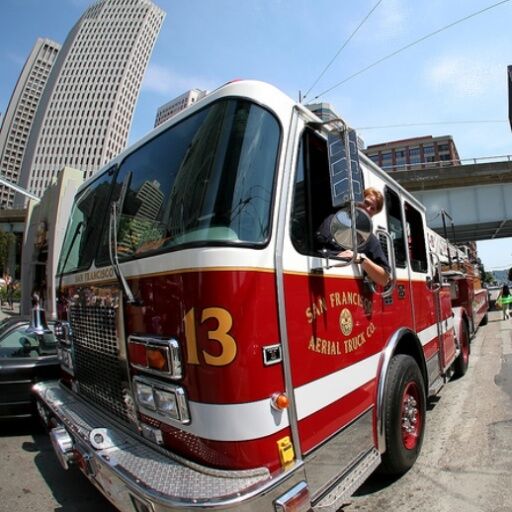} & \includegraphics[width=\imgw2]{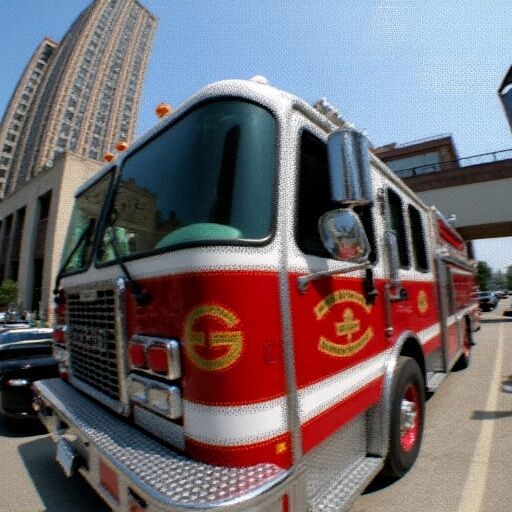} & \includegraphics[width=\imgw2]{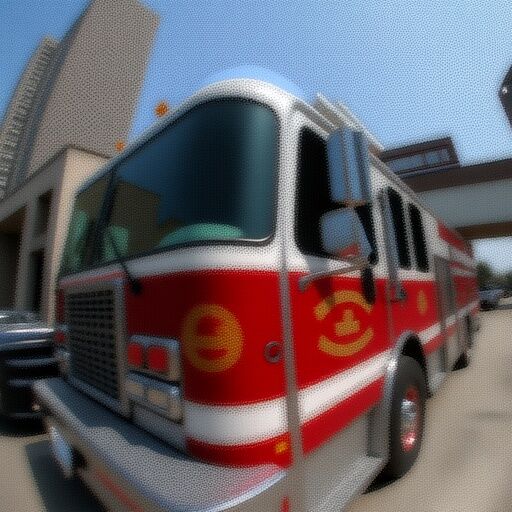} \\
        \includegraphics[width=\imgw2]{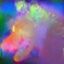} & \includegraphics[width=\imgw2]{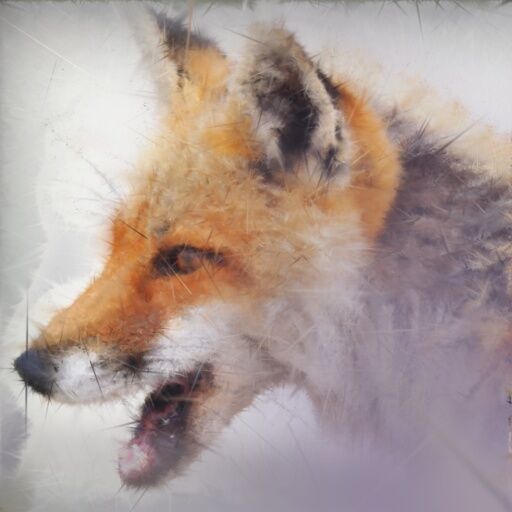} & \includegraphics[width=\imgw2]{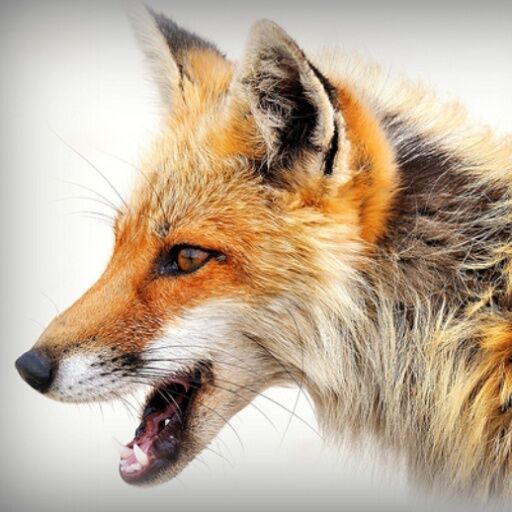} & \includegraphics[width=\imgw2]{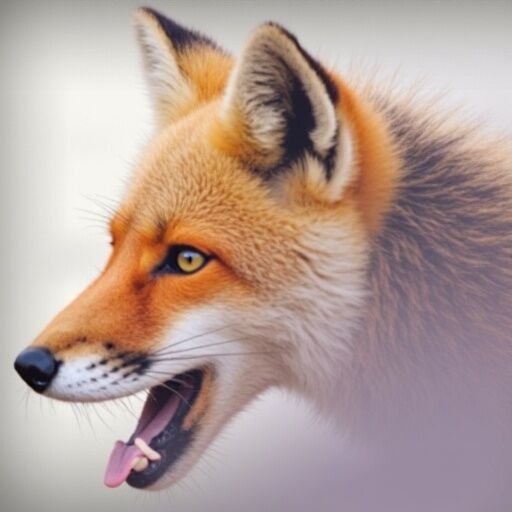} & \includegraphics[width=\imgw2]{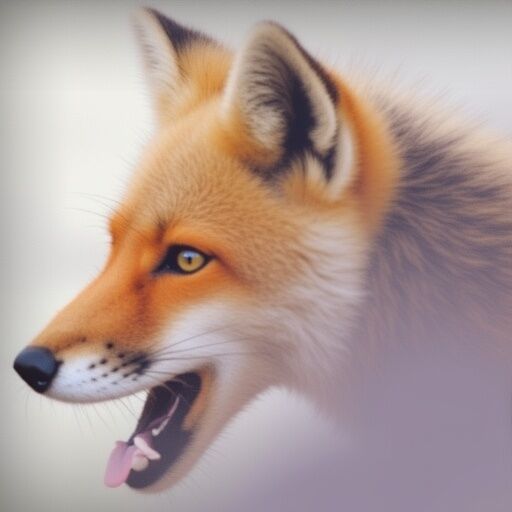} \\
\bottomrule
    \end{tabular}
    \caption{Qualitative examples of 2D image generation with and without Difix3D+~\cite{Wu2025c} finetuning (FT) for splat artifact removal.}
    \label{fig:finetuning_2d}
\end{figure}

\end{document}